\documentclass{article}

\usepackage{arxiv}

\usepackage{amsmath,amsfonts}
\usepackage{algorithmic}
\usepackage{algorithm}
\usepackage{array}
\usepackage{textcomp}
\usepackage{stfloats}
\usepackage{url}
\usepackage{verbatim}
\usepackage{graphicx}
\usepackage[colorlinks,linkcolor=red,anchorcolor=blue,citecolor=green]{hyperref}

\usepackage{stmaryrd}
\usepackage{multirow}
\usepackage{subfigure}
\usepackage[numbers]{natbib}
\DeclareMathOperator*{\argmax}{argmax}

\title{Universal Adversarial Defense in Remote Sensing Based on Pre-trained Denoising Diffusion Models}

\author{Weikang Yu, Pedram Ghamisi\\
        Helmholtz Institute Freiberg for Resource Technology\\
	Helmholtz-Zentrum Dresden-Rossendorf (HZDR)\\
	Freiberg, 09599 Germany\\
	\texttt{\{w.yu, p.ghamisi\}@hzdr.de} \\
	\And
        Yonghao Xu \thanks{Corresponding author} \\
	Computer Vision Laboratory\\
	Linköping University\\
	Linköping, 58183 Sweden \\
	\texttt{yonghao.xu@liu.se} \\
}
\renewcommand{\shorttitle}{UAD-RS}
\begin{document}
\maketitle
\begin{abstract}
Deep neural networks (DNNs) have risen to prominence as key solutions in numerous AI applications for earth observation (AI4EO). However, their susceptibility to adversarial examples poses a critical challenge, compromising the reliability of AI4EO algorithms. This paper presents a novel Universal Adversarial Defense approach in Remote Sensing Imagery (UAD-RS), leveraging pre-trained diffusion models to protect DNNs against universal adversarial examples exhibiting heterogeneous patterns. Specifically, a universal adversarial purification framework is developed utilizing pre-trained diffusion models to mitigate adversarial perturbations through the introduction of Gaussian noise and subsequent purification of the perturbations from adversarial examples. Additionally, an Adaptive Noise Level Selection (ANLS) mechanism is introduced to determine the optimal noise level for the purification framework with a task-guided Frechet Inception Distance (FID) ranking strategy, thereby enhancing purification performance. Consequently, only a single pre-trained diffusion model is required for purifying universal adversarial samples with heterogeneous patterns across each dataset, significantly reducing training efforts for multiple attack settings while maintaining high performance without prior knowledge of adversarial perturbations. Experimental results on four heterogeneous RS datasets, focusing on scene classification and semantic segmentation, demonstrate that UAD-RS outperforms state-of-the-art adversarial purification approaches, providing universal defense against seven commonly encountered adversarial perturbations.

\end{abstract}
	
\section{Introduction}
With the advancement of artificial intelligence (AI) algorithms over recent decades, deep learning-based approaches have become the leading solutions for many earth observation applications, such as change detection \cite{zhang2022multilevel} and disaster monitoring \cite{zhang2023cross}. Despite the promising results achieved in these areas, deep learning methods remain vulnerable to adversarial examples. By introducing imperceptible perturbations to remotely sensed data, adversarial examples can be generated to fool state-of-the-art deep learning models into making wrong predictions with high confidence. These adversarial examples have been identified in a wide range of remote sensing data, including optical \cite{chen2021empirical}, SAR \cite{xia2022sar}, and hyperspectral imagery \cite{shi2021hyperspectral}, affecting various applications such as scene classification and semantic segmentation. Consequently, adversarial examples pose a significant threat to current AI for Earth Observation (AI4EO) frameworks, compromising the reliability and responsibility of these algorithms, particularly in high-risk applications \cite{xu2023ai}. To address the hazards of adversarial examples, various adversarial defense methods have been developed with the goal of improving the robustness of the deep learning algorithms against adversarial perturbations. 
	% \begin{figure}
	% 	\centering
	% 	\subfigure[Scene Classification]{\includegraphics[page=1,scale=0.25]{figures/pdf/figure1.pdf}}
	% 	\subfigure[Semantic Segmentation]{\includegraphics[page=2,scale=0.3]{figures/pdf/figure1.pdf}}
	% 	\caption{Illustration of adversarial attacks on scene classification and semantic segmentation in remote sensing (RS) images. With the addition of minimal adversarial noise, deep neural networks (DNNs) can be deceived into making drastically different predictions for images that appear almost identical.}
	% 	\label{advattack}	
	% \end{figure}

In general, adversarial defense methods can be categorized into adversarial training, randomization, detection, and adversarial purification. Adversarial training methods \cite{zheng2020efficient} aim to enhance a model’s resilience to adversarial examples by incorporating them directly into the training process, thereby enabling the model to counteract adversarial perturbations during testing. However, adversarial training typically addresses only the perturbations seen during training and often fails to generalize to universal adversarial examples. In contrast, randomization-based methods \cite{xie2017mitigating} employ randomness to achieve robustness by blending adversarial perturbations with random noise. Despite this, classification with noisy data mixtures still suffers from degraded feature quality compared to adversarial-free samples. Detection-based methods \cite{chen2021lie} focus on identifying and removing adversarial examples from the dataset using an additional detection network, but this can compromise the dataset's integrity. Consequently, although these methods have shown efficacy in enhancing the robustness of AI4EO applications, they often require specific deep learning algorithm designs or compromise dataset integrity, making them challenging to integrate into existing AI4EO workflows.

Motivated by the straightforward idea of directly removing adversarial perturbations from remote sensing data, adversarial purification methods have been developed. These methods utilize an external denoising network to purify adversarial examples end-to-end before they are fed into deep learning algorithms. Since they are independent of the application networks, adversarial purification methods can be integrated into any AI4EO application. For example, \citet{xu2022task} proposed a denoising convolutional neural network (CNN) with a task-guided loss to remove the perturbations of adversarial samples for remote sensing (RS) scene classification. Similarly, \citet{cheng2021perturbation} developed a perturbation-seeking Generative Neural Network (GAN) structure to purify adversarial examples for robust RS scene classification. However, the practicality of existing adversarial purification methods is still limited due to common issues such as training instability and the limited sampling diversity of GAN models \cite{saxena2021generative}, as well as the inadequate generalization capability of CNNs \cite{swiderski2022random}. Furthermore, existing adversarial defense methods introduced above are typically trained to target specific adversarial attacks, which can render them ineffective in complex real-world scenarios where they must adapt to a wider range of adversarial attacks with varying algorithms and intensities.
 \begin{figure}
		\centering
		\includegraphics[width=\textwidth]{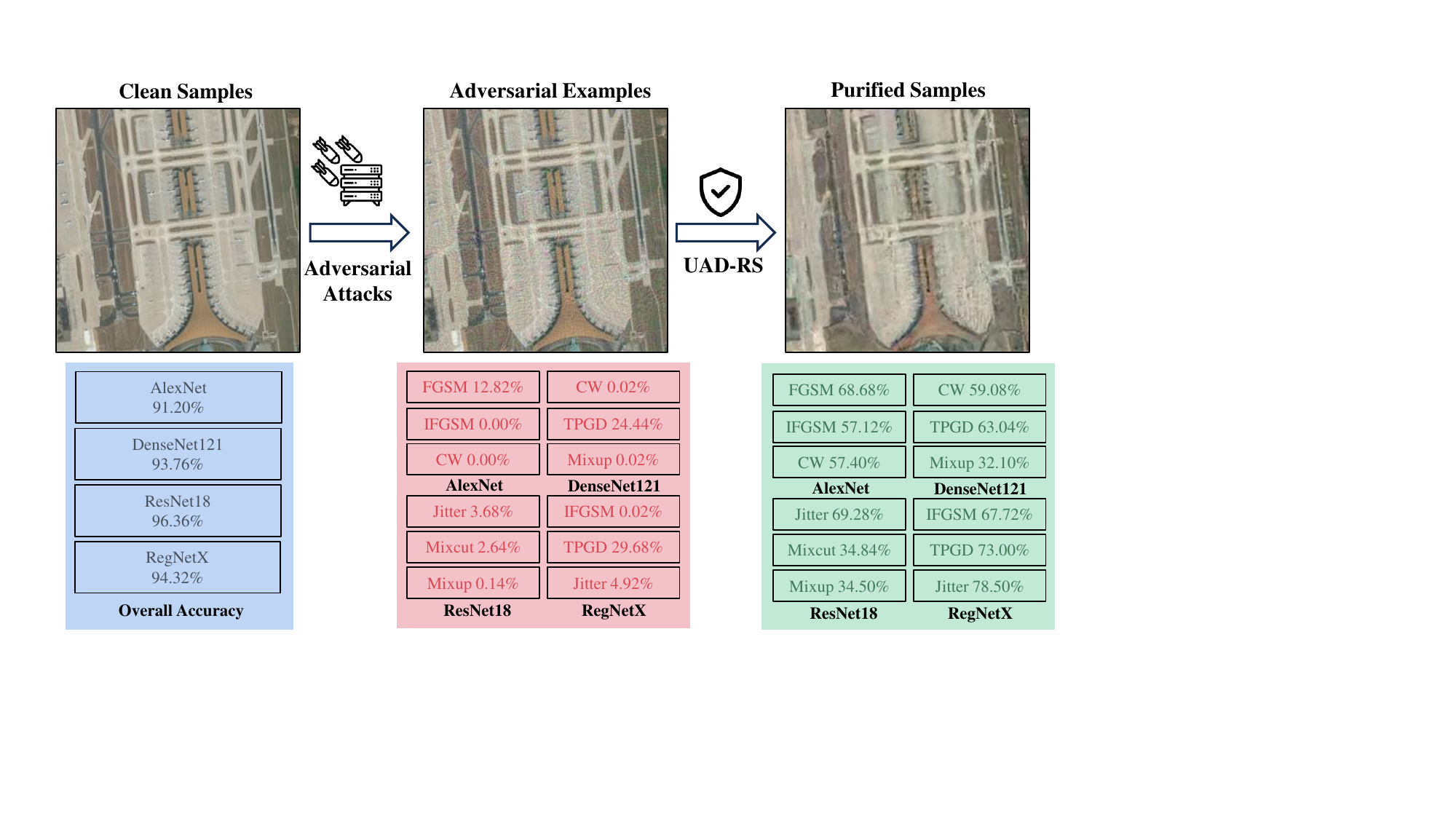}
		\caption{Illustration of universal adversarial defense on scene classification in remote sensing (RS) images. The adversarial examples generated with heterogeneous attack methods significantly affect the performance of different DNNs. The proposed UAD-RS aims to protect the DNNs from the universal adversarial patterns by purifying the adversarial examples in a unified model.}
		\label{advattack}	
	\end{figure}
Recent advances in diffusion models offer new insights into enhancing the generalization capabilities of deep learning models \cite{croitoru2023diffusion}, inspiring us to explore this technique for adversarial purification. 
Diffusion models employ a process of gradually adding Gaussian noise to an image input and then reversing this process to recover a clean image. We integrate this concept, along with the randomness and adversarial purification strategies mentioned above, to develop a universal adversarial defense framework for remote sensing (UAD-RS).
As shown in Fig. \ref{advattack}, this framework aims to protect AI4EO applications from multiple sources of adversarial examples within a unified model. First, we pre-train diffusion models on the target remote sensing dataset to establish generalization capabilities in adversarial-free domains. Then, the pre-trained diffusion models are incorporated into the UAD-RS framework to disrupt the adversarial perturbations with Gaussian noise and purify them into adversarial-free samples in the forward diffusing and reverse denoising processes, respectively. 
Specifically, to maximize the purification efficacy for universal adversarial perturbations, we designed an adaptive noise level selection (ANSL) algorithm to automatically determine the optimal Gaussian noise level for the UAD-RS framework based on a task-guided Fréchet inception distance (FID) ranking strategy.

The main contributions of this study are summarized as follows:
	\begin{enumerate}
		\item We develop a universal adversarial defense framework (UAD-RS) to improve the robustness of AI4EO applications against universal adversarial examples generated by various attack algorithms using a unified pre-trained diffusion model, whichmix random noise with the adversarial examples and purify the mixture into adversarial-free examples through a forward diffusing and a reverse denoising processes, respectively.
        \item An adaptive noise level selection (ANSL) algorithm is developed to dynamically find the optimal noise level for the diffusing and denoising processes in UAD-RS based on the task-guided FID ranking strategy, achieving optimal purification performance for universal adversarial examples with heterogeneous patterns.
        \item Extensive experiments on four RS datasets for both scene classification and semantic segmentation demonstrate that UAD-RS outperforms state-of-the-art methods for universal adversarial defense against seven common adversarial attacks, with only one pre-trained diffusion model required for each RS dataset.
        \end{enumerate}	
	
	%The article is organized as follows. In section II, a brief review is provided regarding adversarial attack, adversarial purification, and diffusion models. Section III describes the proposed adversarial defense methodology based on the pretraining diffusion models, while the performance of generative diffusion models and adversarial purification framework is experimentally evaluated in Section IV. In Section V, the experimental results are thoroughly discussed, with the limitations of the generative diffusion-based adversarial defense method and several potential avenues for future research being identified. Finally, Section VI concludes this article.
The article is organized as follows. Section \ref{sec:related work} provides a brief review of adversarial attacks, adversarial purification, and diffusion models. Section \ref{sec:methodology} describes the proposed adversarial defense methodology based on pre-trained diffusion models and an adaptive noise level selection mechanism. The performance of generative diffusion models and the adversarial purification framework is experimentally evaluated in Section \ref{sec:experimental results}, while the experimental results are thoroughly analyzed. In Section \ref{sec:discussion}, some key concerns regarding this paper are discussed along with the limitations of the generative diffusion-based adversarial defense method, and several potential avenues for future research are identified. Finally, Section \ref{sec:conclusion} concludes this article.

	\section{Related Work}\label{sec:related work}
	\subsection{Adversarial Attacks}\label{attackdescription}
In the last decade, various adversarial attack algorithms have been developed to create more intense adversarial perturbations that can significantly impact a broader range of deep neural networks (DNNs). Contrary to the typical training objective of minimizing the loss function, adversarial perturbations are designed to deceive the model into making incorrect predictions by maximizing the loss function. Depending on the accessibility of the model’s internal parameters, adversarial perturbations can be generated through white-box attacks, such as the Fast Gradient Sign Method (FGSM) \cite{goodfellow2014explaining}, or black-box attacks, such as Mixcut-attack \cite{xu2022universal}. We briefly introduce several widely adopted adversarial attack algorithms employed in our subsequent experiments as follows.
	\subsubsection{Fast Gradient Sign Method (FGSM)}
	One of the most intuitive strategies for adversarial attacks is leveraging how they learn gradients. Based on this idea, a gradient-based attack, namely, Fast Gradient Sign Method (FGSM), was proposed to adjust the input data to maximize the objective function by adjusting the weights based on the back-propagated gradients \cite{goodfellow2014explaining}.
	Given an image $x$ and its true label $y$, the adversarial sample $x^{adv}$ can be generated as:
	\begin{eqnarray}
		x^{adv}=x+\epsilon \cdot sign(\nabla_{x}\mathcal{L}(\theta,x,y)),
	\end{eqnarray}
	where $sign$ denotes the sign function, $\nabla_{x}J(\theta,x,y)$ calculates the gradients of an objective function $\mathcal{L}(\cdot,\cdot)$ with respect to each input $x$ and ground-truth $y$, and $\epsilon$ is a scalar value that restricts the norm of the perturbation. To improve the performance of the adversarial attack, Kurakin et al. \cite{kurakin2016adversarial} proposed an iterative-FGSM (IFGSM), which applies FGSM multiple times with a small step size, as follows:
	\begin{eqnarray}
		x_{t+1}^{adv}=x^{adv}_{t}+\alpha\cdot sign(\nabla_{x^{adv}_{t}}\mathcal{L}(\theta,x_{t}^{adv},y)),
	\end{eqnarray}
	where an adversarial sample $x_{adv}$ is iteratively calculated in $T$ steps of FGSM, and $\alpha=\frac{\epsilon}{T}$ is the step size that reduces the perturbation of each step that sums to a similar intense of attack with the FGSM.
	\subsubsection{Trade-off Projected Gradient Descent (TPGD) Attack}
	Since FGSM attacks aim to maximize a loss function measuring the pixel-wise difference between probability maps of adversarial and clean images, \citet{zhang2019theoretically} proposed the Trade-off Projected Gradient Descent (TPGD) Attack. This method generates adversarial perturbations based on the Kullback–Leibler (KL) divergence between the probability distributions of clean and adversarial predictions, serving as a cross-domain covariance.:
	\begin{equation}
		x_{t+1}^{adv}=x_{t}^{adv}+\nabla_{x_{t}^{adv}} \mathbf{KL}(\theta, x_{t}^{adv}, y).
	\end{equation}
	
	\subsubsection{Carlini and Wagner (CW) Attack}
    Instead of directly utilizing objective functions to measure adversarial samples, Carlini and Wagner (CW) proposed an attack method that encourages the adversarial sample $x_{adv}$ to have a higher probability score for an incorrect class than for the correct class. The CW method directly optimizes the distance between the clean and adversarial samples as follows:
	\begin{equation}
		\mathop{\arg\min}\limits_{x^{adv}} ||x^{adv}-x||_{\infty}-\mu\mathcal{L}(\theta,x^{adv},y),
	\end{equation}
	where $\mu$ is a weighting factor. 
	
	\subsubsection{Mixcut-Attack}
	Since FGSM and its variants are white-box attacks that require complete knowledge of the target model, they are impractical for real-world scenarios where detailed information about the deployed model is usually unavailable, especially in the field of RS \cite{akhtar2021advances}. Consequently, black-box attacks have been proposed to generate adversarial samples without prior knowledge of the victim model \cite{narodytska2017simple}. One of the most advanced black-box attacks for RS imagery is the Mixup-attack, which employs a surrogate model to produce universal adversarial examples capable of deceiving various heterogeneous DNNs with a high success rate \cite{xu2022universal}. Given an input $x$ with ground truth $y$ and a surrogate model $\theta_{s}$, the adversarial samples can be iteratively generated as follows:
	\begin{align}
		g^{t+1}&=g^{t}+\frac{\nabla_{x}\mathcal{L}(\theta_{s},x^{t},y)}{||\theta_{s},x^{t},y)||_{1}},\\
		x^{t+1}&=clip(x^{t}+\alpha\cdot\frac{g^{t+1}}{||g^{t+1}||_{\infty}}),\\
		x^{adv}&=x^{T},
	\end{align}
	where $g_{t}$ denotes the momentum term at the $t$-th iteration, and $clip(\cdot)$ clips the pixel values in the image.

	\subsection{Adversarial Purification}
        The development of Convolutional Neural Networks (CNNs) and Generative Adversarial Networks (GANs) has significantly advanced research on adversarial purification methods. Recent literature highlights two primary approaches to purify perturbations from adversarial samples: denoising-based methods and generative-based methods. Denoising-based methods leverage the progress in denoising models within computer vision to effectively remove perturbations as noise using end-to-end CNNs. For example, \citet{meng2017magnet} introduced a reformer network that defends against adversarial examples by transforming them towards the manifold of normal samples. This reformer network utilizes an auto-encoder (AE) with an encoder to extract high-dimensional features from the input image and a decoder to reconstruct an adversarial-free image. Additionally, \citet{xu2022task} developed a task-guided loss that aligns adversarial and normal images in the perceptual domain, thereby enhancing adversarial purification performance.

        Conversely, generative-based methods employ domain translation to convert adversarial examples into an adversarial-free domain. GANs, known for their robust representation capabilities within specific domain distributions, consider adversarial examples as conditions to generate corresponding clean domain images. For example, \citet{samangouei2018defense} proposed Defense-GAN, which utilizes GANs' expressive capabilities to filter out perturbations for adversarial purification. Defense-GAN trains a CNN-based generator to learn the distribution of unperturbed images and generates outputs resembling the original images without adversarial alterations during inference.
        However, existing adversarial purification methods face challenges in defending against heterogeneous adversarial perturbations due to the limited generalization capabilities of the CNNs and CNN-based generators in GANs. Additionally, the reliability of generative-based models is constrained by unstable training issues, such as vanishing gradients, mode collapse, and failure to converge \cite{weng2019gan,zhang2018convergence}.
        
        In this paper, we integrate the next-generation generative model, the diffusion model, into the Universal Adversarial Defense for Remote Sensing (UAD-RS) to explore universal adversarial purification in remote sensing imagery. Diffusion models exhibit robust generalization capabilities in representing data distributions and can be leveraged to learn adversarial-free representations \cite{wang2022guided}. The random Gaussian noises in the diffusion process can disrupt complex adversarial distributions, which can then be removed in the reverse denoising process. Utilizing these advantages, the UAD-RS framework enables the purification of universal adversarial examples using a unified diffusion model.

	\subsection{Diffusion Models}\label{related work: diffusion model}
	
	% Diffusion models are a class of probabilistic generative models designed for unsupervised modelling, which have shown strong sample quality and diversity in image synthesis \cite{song2021maximum}. In practical cases, generative diffusion models usually define two opposite processes, including a diffusion process and a reverse process, to gradually learn the distribution of training data \cite{ho2020denoising}. Particularly, the diffusion process gradually adds noise to the image input until it becomes a Gaussian noise, which is iteratively denoised by the reverse process to reconstruct a clean image. Let $p_{data}$ be the distribution of all the input images $x$, and $p_{latent}$ be the latent distribution, the diffusion process $q$ with $T$ steps can be defined as follows:
	Diffusion models are a class of probabilistic generative models designed for unsupervised modeling, and they have demonstrated strong sample quality and diversity in image synthesis \cite{song2021maximum}. In practical scenarios, generative diffusion models typically involve two complementary processes: a diffusion process and a reverse process, which iteratively learns the distribution of the training data \cite{ho2020denoising}. Specifically, the diffusion process gradually introduces noise to the input image until it becomes Gaussian noise, while the reverse process denoises the noise iteratively to reconstruct a clean image. Let $p_{data}$ represent the distribution of all input images $x$, and $p_{latent}$ denote the latent distribution. The diffusion process $q$ with $T$ steps can be defined as follows:
	\begin{align}
		\label{forwarddiff}		q(x_{1},...,x_{T}|x_{0})=\prod_{t=1}^{T}q(x_{t}|x_{t-1}). 
	\end{align}
	In contrast, the reverse process iteratively eliminates the noise added to the diffusion process, gradually restoring a clean image. Given the latent variable $x_{T}$, the reverse process $p$ also consists of $T$ steps, resulting in the generation of the clean data $x_{0}$ as follows:
	\begin{align}
		\label{reversediff}
		p_{\theta}(x_{0},...,x_{T-1}|x_{T})=\prod_{t=1}^{T}p_{\theta}(x_{t-1}|x_{t}),
	\end{align}
	where $\theta$ parameterized the reverse diffusion process. In the literature, there are mainly two variants of diffusion models: Denoising diffusion probabilistic model (DDPM) \cite{ho2020denoising} and stochastic differential equation (SDE)-based generative methods \cite{song2020score}. Both methods introduce Gaussian noise to the input data in the forward process but employ different denoising algorithms in the reverse process. 
	On one hand, DDPM incorporates two Markov chains for the forward and reverse processes and aims to match the reverse transitional kernel $p_{\theta}\left(x_{t-1}|x_{t}\right)$ with the forward transitional kernel $q\left(x_{t}|x_{t-1}\right)$ at each time step $t$ by adjusting the parameter $\theta$ in the reverse Markov chain.
	On the other hand, SDE-based generative methods typically utilize a noise-conditioned score network (NCSN) to estimate score functions for all noise distributions. These score functions are sequentially applied to decrease the noise levels and eventually sample a clean image.
	Intuitively, DDPM tackles the Gaussian Markov chain as a discrete SDE using Ancestral Sampling, while SDE-based generative models focus on solving continuous-time SDEs based on Langevin Dynamics.
	
	\section{Methodology}\label{sec:methodology}
 This section provides a detailed description of the proposed UAD-RS adversarial purification framework, as shown in Fig. \ref{fulldiff}. First, a Denoising diffusion probabilistic model (DDPM) model is pre-trained on a target remote sensing dataset to obtain the generalization capability on the specific adversarial-free domain. After that, the pre-trained model is applied in the UAD-RS framework to iteratively diffuse the adversarial examples into noise latent and subsequently denoise them into clean images. Specifically, an adaptive noise level selection (ANSL) module is introduced to find the optimal noise level for the purification process with a Fréchet inception distance (FID) ranking strategy.

\subsection{Pre-training Generative Diffusion Model}\label{method:pretrain}
	\begin{figure*}
		\centering
		\includegraphics[scale=0.6]{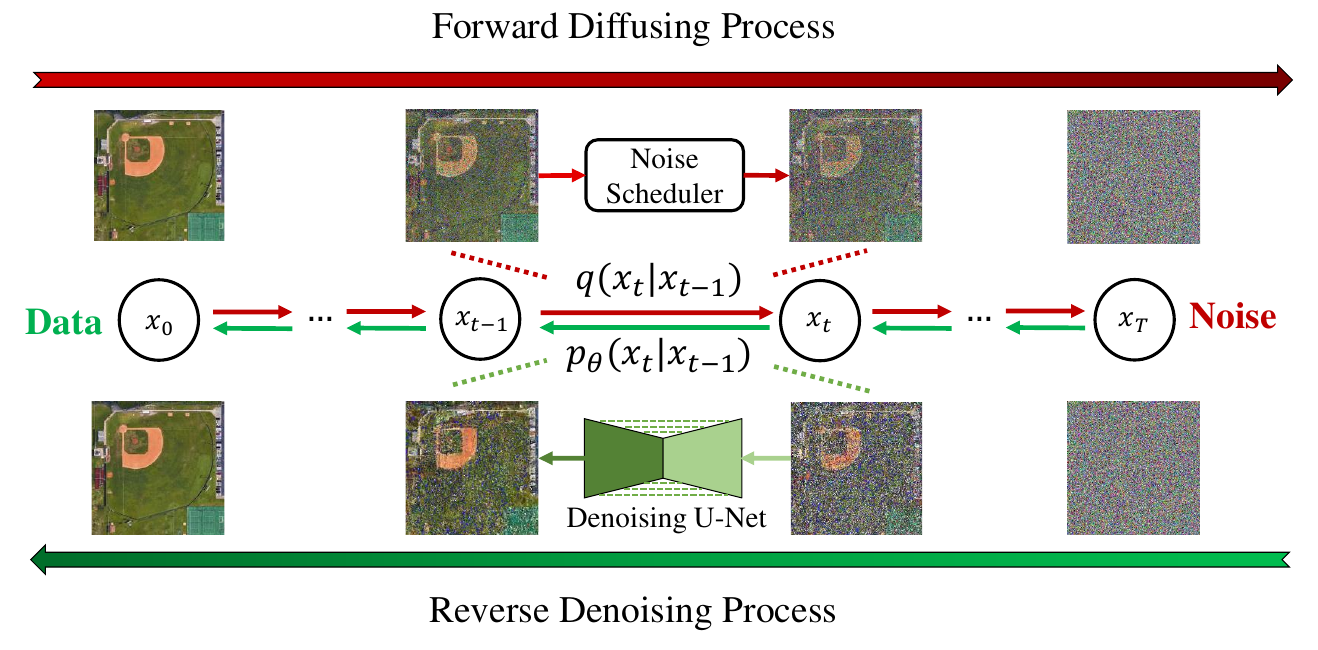}
		\caption{Illustration of the forward and reverse processes of the generative diffusion models in pre-training phase. The forward diffusion process gradually adds Gaussian noise to the images using the noise scheduler, and finally, pure Gaussian noise is generated. After that, the reverse process progressively recovers the noise to reconstruct an image with a denoising U-Net model.}
		\label{fulldiff}
\end{figure*}
    Generative diffusion models have demonstrated superior generalization capabilities in learning domain representations and have been widely applied in vision tasks, such as image synthesis and image translation, within the remote sensing community. Before applying diffusion models to these tasks, a pre-training process is essential. This process enables the models to learn data distributions, handle noise, extract meaningful features, and generalize effectively, subsequently laying a robust foundation for high-quality generative performance and efficient execution of downstream tasks. Therefore, we employ a pre-training process for the DDPM model to learn the heterogeneous data distributions in an adversarial-free domain for each remote sensing dataset.
    
    As shown in Fig \ref{fulldiff}, the pre-training of diffusion models leverages a forward diffusing process and a reverse denoising process, which iteratively adds random noise to the input image with a noise scheduler and denoises them with a denoising U-Net, respectively. According to Eqn. \ref{forwarddiff}. the forward diffusing process is taken by a transitional kernel $q(x_{t}|x_{t-1})$  handcrafted by Gaussian perturbations, in $T$ iterative steps. Let $\mathcal{N}$ be a Gaussian distribution with variance $I$, an iterative step of the diffusing process can be calculated as follows:
    \begin{eqnarray}
		q(x_{t}|x_{t-1})=\mathcal{N}(x_{t};\sqrt{1-\beta_{t}}x_{t-1},\beta_{t}I)),
    \end{eqnarray}
    where $\beta_{t}$ represents the variance schedule for noise addition in the $t$-th diffusion step, which is controlled by the noise scheduler. After that, the DDPM incorporates a learnable Markov chain parameterized jointly by a prior distribution $p(X_{T})=\mathcal{N}(x_{T};0,I)$ and a denoise transitional kernel $p_{\theta}(x_{t-1}|x_{t})$ for the reverse denoising process in Eqn. \ref{reversediff}. A step of the reverse denoising process can be calculated as follows:
    \begin{eqnarray}
		p_{\theta}(x_{t-1}|x_{t})=\mathcal{N}(x_{t-1};\mu_{\theta}(x_{t},t),{\sum}_{\theta}(x_{t},t)),
	\end{eqnarray}
	where the mean $\mu_{\theta}(x_{t},t)$ and variance ${\sum}_{\theta}(x_{t},t)$ are usually calculated from the denoising U-Net parameterized by $\theta$. Notably, the reverse process is linked with the forward process by the form of prior distribution $p(X_{T})$, which is approximately the same as $q(x_{T})$. To simplify the implementation of the forward diffusing process, DDPM utilized a closed form of perturbed representations as follows:
    
    % It can be observed that the prior distribution is linked to the Gaussian distribution in the forward process
    
    % Concerning the reverse diffusion process in Eq. (\ref{reversediff}), DDPM incorporates a learnable Markov chain parameterized jointly by a prior distribution $p(X_{T})=\mathcal{N}(x_{T};0,I)$, approximately the same as $q(x_{T})$, and a reverse transitional kernel $p(x_{t-1}|x_{t})$ as follows:
    \begin{equation}
		q(x_{t}|x_{0})=\mathcal{N}(x_{t};\sqrt{\overline{\alpha}_{t}} x_{0},(1-\overline{\alpha}_{t})I),
	\end{equation}
	\begin{equation}
		\label{simplyforwarddiff}
		x_{t}=\sqrt{\overline{\alpha}_{t}} x_{0}+\sqrt{1-\overline{\alpha}_{t}} \epsilon,
	\end{equation}
	where $\alpha_{t}:=1-\beta_{t}$, $\overline{\alpha}_{t}:=\prod_{s=0}^{t}\alpha_{s}$, and $\epsilon$ is a standard Gaussian noise. Based on the forward and revered processes above, DDPM utilizes a Kullback-Leibler (KL) divergence to optimize the trainable parameters $\theta$ in the denoising network by approximately matching the actual time reversal of the diffusing Markov chain \cite{yang2022diffusion}:
	\begin{equation}
		\mathbf{KL}(q(x_{0},x_{1},...,x_{T})|p_{\theta}(x_{T},x_{T-1},...,x_{0})),
	\end{equation}
	\begin{equation}
		=-\mathbb{E}_{q(x_{0},x_{1},...,x_{T})}[\log p_{\theta}(x_{0},x_{1},...,x_{T})] + const,
	\end{equation}
	\begin{equation}
		\label{vlb}
		=-\mathbb{E}_{q(x_{0},x_{1},...,x_{T})}[-\log p(x_{T})- \sum_{t=1}^{T} \log \frac{p_{\theta}(x_{t-1}|x_{t})}{q(x_{t}|x_{t-1})}] + const,
	\end{equation}
 where the first term in Eq. (\ref{vlb}) represents the variational lower bound of the log-likelihood of the data $x_{0}$, which is commonly considered as a training objective for probabilistic generative models. By further incorporating a step-wise weight schedule $\lambda (t)$, the training objective function can be further derived from KL divergence as:
	\begin{eqnarray}
		\mathbb{E}_{t\sim\mathcal{U}\llbracket 1,T \rrbracket,x_{0}\sim q(x_{0}), \epsilon \sim \mathcal{N}(0,I)}[\lambda (t) \Vert \epsilon - \epsilon_{\theta} (x_{t},t) \Vert^{2}],
	\end{eqnarray}
	where $x_{t}$ can be calculated from $x_{0}$, $\epsilon\sim (0,I)$ is the Gaussian vector for sampling, $\mathcal{U}\llbracket 1,T \rrbracket$ is a uniform set over $\{0,1,2,...,T\}$, and $\epsilon_{\theta}$ represents the denoising U-Net model parameterized by $\theta$ that predicts the noise vector $\epsilon$ given $x_{t}$ and $t$.

	\subsection{Unified Adversarial Purification Framework}
	%	Taking the idea of powerful generative ability in a marginalized domain of the diffusion models, we propose to use a pretrained DDPM model to purify the adversarial samples on RS imagery. The DDPM consists of a forward process injecting Gaussian noise to a clean image and a reverse process that eliminates the noise in the image, which can be considered to destroy the perturbations of the adversarial samples and reconstruct images in the clean domain, respectively. In particular, the forward diffusion can gradually add Gaussian noise to adversarial samples that gradually submerge the adversarial perturbation, and the reverse diffusion can eliminate both Gaussian noise and adversarial samples at the same time and marginalize the prediction in the adversarial-free domain with the pretrained models.
	\begin{figure*}
		\centering
		\includegraphics[width=\linewidth]{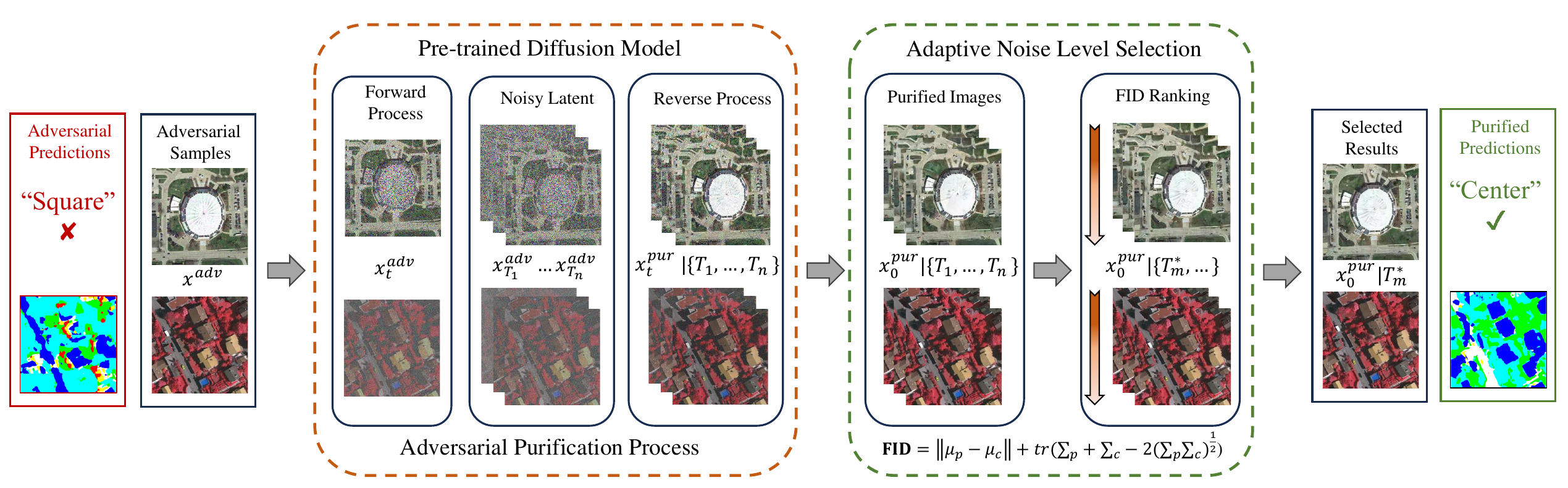}
		\caption{Overview of the proposed UAD-RS adversarial purification framework. The UAD-RS purification process effectively removes adversarial perturbations by first disrupting them with Gaussian noise and then denoising the resulting latent to produce clean images in the forward and reverse process of the pre-trained diffusion model, respectively. Subsequently, the Adaptive Noise Level Selection (ANLS) algorithm identifies the optimal noise level $T_{m}^{*}$ that yields the best purification results, compared to other noise levels, by calculating and ranking their task-guided Frechet Inception Distance (FID) scores. Ultimately, the purified predictions circumvent the influence of the original perturbations, ensuring accurate results.}
		\label{overview}
	\end{figure*}
As introduced in Section \ref{attackdescription}, adversarial examples $x^{adv}$ are generated by adding perturbations $\delta$ to clean data $x$, expressed as $x^{adv} = x + \delta$. These perturbations, created using various attack algorithms, exhibit different patterns and intensities. The proposed UAD-RS aims to purify these perturbations using a unified framework, requiring only a single pre-trained DDPM model for each remote sensing dataset. This approach eliminates the need for multiple training sessions for each type of attack, significantly enhancing the robustness of DNNs against unknown adversarial examples.

Using the pretrained DDPM model, the universal adversarial purification framework eliminates adversarial perturbations through forward diffusing and reverse denoising processes. In the forward diffusing process, adversarial examples are mixed with Gaussian noise, which disrupts the patterns of adversarial perturbations. Following Eqn. \ref{simplyforwarddiff}, the forward diffusion process can be computed as follows:
\begin{equation}
		\label{simplifiedforward}
		x_{T_{m}}^{adv}=\sqrt{\overline{\alpha}_{T_{m}}}x^{adv}+\sqrt{1-\overline{\alpha}_{T_{m}}}\epsilon,
\end{equation}
where $x_{T_{m}}^{adv}$ is the noise latent and $T_{m}$ is the noise level controlling the number of diffusing and denoising steps. As $T_{m}$ increases, the weight factor $\overline{\alpha}_{T_{m}}:=\prod_{s=0}^{T_{m}}\alpha_{s}$ decreases significantly since $\alpha_{s}< 1$, causing Gaussian noise to dominate in the noise latent. A relatively small $T_{m}$ is selected to preserve the texture and details of the input adversarial examples that can be effective information for AI4EO applications. The noise latent $x^{adv}_{T_{m}}$ is then reconstructed into purified samples by the denoising process of $T_{m}$ steps, as follows:
\begin{align}
		p_{\theta^{*}}(x^{pur}_{t-1}|x_{t}^{pur})&=\mathcal{N}(x_{t-1}^{pur};\mu_{\theta^{*}}(x_{t}^{pur},t),{\sum}_{\theta^{*}}(x_{t}^{pur},t)),
	\end{align}
	\begin{equation}
		x^{pur}_{T_{m}}:=x^{adv}_{T_{m}},
	\end{equation}
where $\theta^{*}$ are the parameters of the denoising U-Net from the pretrained DDPM. Each denoising step reconstructs clean representations within the learned adversarial-free data distributions. The purified image $x^{pur}_{0}$ is obtained with adversarial perturbations removed along with the added Gaussian noise. In conclusion, the universal adversarial purification framework improves model robustness against adversarial examples by disrupting adversarial patterns with Gaussian noise and eliminating these noises during the denoising process.

\subsection{Adaptive Noise Level Selection}
In the universal adversarial purification framework, the noise level $T_{m}$ is a critical variable that controls the amount of added Gaussian noise and the effectiveness of the denoising process, directly impacting the performance of adversarial purification. If $T_{m}$ is set too high, Gaussian noise can destroy details and textures, making it impossible to recover an informative image during denoising. Conversely, if $T_{m}$ is too low, the Gaussian noise may be too mild to counteract the adversarial perturbations, leaving them predominant in the noise latent and insufficiently removed by the denoising steps with fewer steps. This phenomenon is demonstrated by the experiments shown in Fig. \ref{ablation_noise_level}.
To maximize the effectiveness of adversarial purification, an appropriate noise level should be selected based on the specific patterns of adversarial perturbations. However, since these patterns vary in intensity and appearance depending on the attack method, finding an optimal noise level for each pattern of adversarial perturbation in universal adversarial examples remains complicated.

To address this challenge, we propose an Adaptive Noise Level Selection (ANSL) module to adaptively determine the optimal noise level $T_{m}$ for purifying universal adversarial examples of heterogeneous patterns with a task-guided FID ranking strategy. The idea comes from the assumption that the purified images with the higher quality should have more consistent deep representations similar to those of the clean images in a well-trained application model, while the adversarial perturbations were initially designed to destroy this consistency \cite{alfarra2022robustness}. 

The ANSL module operates in four steps, summarized in pseudocode in Algorithm \ref{alg:ANLS}. First, an application model $F_{a}$ is trained on the adversarial-free training dataset $\mathcal{D}^{train}$ that is also used for pre-training the DDPM in Section \ref{method:pretrain}. The clean deep representations $\mathcal{F}$ of these samples are obtained using the encoder part of the pre-trained application model $F_{a}^{enc}$. Then, the universal adversarial purification framework $F_{p}$ is applied to a subset of adversarial examples $\mathcal{D}^{adv}$ using a set of noise levels $\mathcal{T}$. The deep representations of the purified samples are then obtained with the feature encoder $F_{a}^{enc}$. Subsequently, two Gaussian distributions, $\mathcal{N}(\mu_{T_{m}}^{pur}, {\scriptstyle\sum}_{T_{m}}^{pur})$ and $\mathcal{N}(\mu, {\scriptstyle\sum})$, are fitted based on the deep representations of purified samples $\mathcal{F}_{T_{m}}^{pur}$ and clean data $\mathcal{F}$, respectively. Finally, the FID value $\gamma_{T_{m}}$ for different noise levels is calculated as follows:
\begin{align}
		\gamma_{T_{m}}=||\mu_{T_{m}}^{pur}-\mu||+
  \mathrm{tr}({\scriptstyle\sum}_{T_{m}}^{pur}+{\scriptstyle\sum}-2\sqrt{{\scriptstyle\sum}_{T_{m}}^{pur}{\scriptstyle\sum}}),
\end{align}
where $\mathrm{tr}(\cdot)$ represents the trace linear algebra operation.

Notably, we only randomly select a small part of the adversarial examples as representatives in ANLS to improve computational efficiency. After determining the optimal noise level, it is applied in the universal adversarial purification framework to purify all the adversarial examples.
\begin{algorithm}
\caption{Adaptive Noise Level Selection}
\label{alg:ANLS}
    \begin{algorithmic}
        \REQUIRE The application model $F_{a}$, the universal adversarial purification framework $F_{p}$, the training dataset $\mathcal{D}^{train}$, a set of adversarial examples $\mathcal{D}^{adv}$, a preset series of noise level $\mathcal{T}$, FID function $d_{F}$
        \ENSURE The optimal noise level $T_{m}^{*}$
        \STATE \textit{\textbf{Step1}: Pre-training application model}
        \STATE Train $F_{a}$ on $\mathcal{D}^{train}$
        \STATE Obtain encoder of the trained model: $F_{a}^{enc} \gets F_{a}$
        \STATE
        \STATE \textit{\textbf{Step2}: Obtaining clean representations}
        \FOR{$x$ in $\mathcal{D}^{train}$}
            \STATE Generate deep representations: $f \gets F_{a}^{enc}(x)$
        \ENDFOR
        \STATE Generate clean representation set: $\mathcal{F} \gets \{f\}$
        \STATE
        \STATE \textit{\textbf{Step3}: Universal adversarial purification}
        \FOR{$T_{m}$ in $\mathcal{T}$}
        \FOR{$x^{adv}$ in $\mathcal{D}^{adv}$}
            \STATE Adversarial purification: $x^{pur}_{T_{m}} \gets F_{p}(x^{adv},T_{m})$
            \STATE Generate deep representations: $f^{pur}_{T_{m}} \gets F_{a}^{enc}(x^{pur}_{T_{m}})$
        \ENDFOR
            \STATE Generate purified representation set: $\mathcal{F}^{pur}_{T_{m}} \gets \{f^{pur}_{T_{m}}\}$
        \ENDFOR
        \STATE
        \STATE \textit{\textbf{Step4}: FID Calculation and Ranking}
        \STATE Fit Gaussian distributions: \\ $\mathcal{N}(\mu_{T_{m}}^{pur}, {\scriptstyle\sum}_{T_{m}}^{pur})\gets \mathcal{F}_{T_{m}}^{pur}$, $\mathcal{N}(\mu, {\scriptstyle\sum})\gets \mathcal{F}$
        \STATE Calculate FID value: \\$\gamma_{T_{m}} \gets d_{F}(\mathcal{N}(\mu_{T_{m}}^{pur}, {\scriptstyle\sum}_{T_{m}}^{pur}), \mathcal{N}(\mu, {\scriptstyle\sum}))$
        \STATE Find optimal noise level: $T_{m}^{*} \gets \argmax_{T_{m} \in \mathcal{T}} \gamma_{T_{m}}$
    \end{algorithmic}
\end{algorithm}

	\section{Experimental Results}\label{sec:experimental results}
\subsection{Dataset Description}
In this article, we evaluate the proposed UAD-RS model using the Universal Adversarial Examples in RS (UAE-RS) dataset \cite{xu2022universal}, which targets the defense of scene classification and semantic segmentation models. The UAE-RS dataset is constructed from four benchmark remote sensing image datasets: two for scene classification (UC-Merced (UCM) and Aerial Image Dataset (AID)) and two for semantic segmentation (Vaihingen and Zurich Summer).
\subsubsection{UCM}
This dataset \cite{yang2010bag} consists of 2,100 overhead scene images sourced from the U.S. Geological Survey (USGS) National Map. The images are categorized into 21 land-use classes, each containing 100 images of size 256 $\times$ 256 pixels with a spatial resolution of 0.3m per pixel in RGB color space.
\subsubsection{AID}
Comprising 10,000 aerial images from 30 scene types collected from Google Earth, the AID dataset \cite{xia2017aid} varies in the number of images per class (ranging from 220 to 420). The dataset features multiple spatial resolutions from 8m to 0.5m per pixel, with all images standardized to a size of 600 $\times$ 600 pixels.
\subsubsection{Vaihingen}
This subset of a semantic segmentation benchmark \cite{cramer2010dgpf} is provided by the International Society for Photogrammetry and Remote Sensing (ISPRS). It contains 33 aerial images annotated with six land cover classes, comprising three bands: near-infrared, red, and green. The images have a spatial resolution of 0.09m, with an average sample size of about 2500 $\times$ 1900 pixels, covering approximately 1.38 square kilometers.
\subsubsection{Zurich Summer}
This urban remote sensing semantic segmentation dataset \cite{volpi2015semantic} includes 20 satellite images annotated with eight land-use classes. The average image size is approximately 1000 $\times$ 1000 pixels with a spatial resolution of 0.62m. Images are captured across four bands: near-infrared, red, green, and blue. In our experiments, we follow the UAE-RS settings, selecting the near-infrared, red, and green channels.

\subsection{Adversarial Attack Settings}
In this article, we evaluate the universal defense using a series of seven representative adversarial attack algorithms. These include both white-box attacks (FGSM \cite{goodfellow2014explaining}, IFGSM \cite{kurakin2016adversarial}, Jitter \cite{schwinn2023exploring}, TPGD \cite{zhang2019theoretically}, and CW \cite{carlini2017towards}) and black-box attacks (Mixcut and Mixup \cite{xu2022universal}) from different technical approaches.

White-box attacks generate adversarial perturbations based on features and gradients extracted from the victim classification models, with their performance heavily dependent on prior knowledge of the DNN structure. In contrast, black-box methods generate adversarial samples without such information by attacking surrogate models independently.
 In our experiments, we perform white-box attacks on each pre-trained baseline classifier and incorporate Mixcut and Mixup attack samples directly from the UAE-RS dataset as black-box attack samples.
Detailed information about these attacks is provided in Section \ref{attackdescription}.

% In this article, a series of seven representative adversarial attack algorithms including both white-box (i.e., FGSM \cite{goodfellow2014explaining}, IFGSM \cite{kurakin2016adversarial}, Jitter \cite{schwinn2023exploring}, TPGD \cite{zhang2019theoretically}, and CW \cite{carlini2017towards}) and black-box attack (i.e., Mixcut and Mixup Attacks \cite{xu2022universal}) from different technique routes are adopted for a comprehensive evaluation of the universal defense. In particular, the white-box attacks can generate adversarial perturbations based on the features and gradients extracted from the victim classification models, and the performance is highly related to prior knowledge of the structure of the DNNs. On the opposite, the black-box methods can directly generate adversarial samples without the information of utilized classification models by attacking the surrogate models in a standalone manner. Detailed information about these attacks has been given in Section \ref{attackdescription}. In our experiments, the white-box attacks are performed for each pre-trained baseline classifier, while we directly incorporate the Mixcut and Mixcut attacked images from the UAE-RS dataset as the black-box attack samples. 

\begin{table}
	\caption{Standard Evaluation Results of the Scene Classification Models.}
	\label{baseline-sc}
	\centering
 \resizebox{\linewidth}{!}{
	\begin{tabular}{ccccc}
		\hline
		\multirow{2}{*}{Datasets} & \multicolumn{4}{c}{DNN Classifiers}                 \\
		& AlexNet & DenseNet-121 & ResNet-18 & RegNetX-400MF \\ \hline
		UCM                       & 91.23   & 96.19       & 96.38    & 95.04            \\
		AID                       & 91.20    & 93.76       & 96.36    & 94.32            \\ \hline
	\end{tabular}}
\end{table}

\subsection{Evaluation Metrics}
%\subsection{Image Synthesis}
\subsubsection{Scene Classification}
The Overall Accuracy (OA$=\mathrm{n}_{\mathrm{correct}}/\mathrm{n}_{\mathrm{total}}$) is adopted for quantitative comparison in the scene classification task, where $\mathrm{n}_{\mathrm{correct}}$ and $\mathrm{n}_{\mathrm{total}}$ represent the amount of correctly classified samples and the size of full test dataset, respectively.
\subsubsection{Semantic Segmentation}
For the quantitative evaluation of pixel-level semantic segmentation results, we employ four commonly used metrics: overall accuracy (OA), precision, recall, and F1-score. These metrics are calculated as follows:
\begin{align}
	\mathrm{OA}&=\frac{\mathrm{TP+TN}}{\mathrm{TP+TN+FP+FN}}, \\
	\mathrm{Precision}^{(c)}&=\mathrm{TP}^{(c)}/\mathrm{(TP^{(c)}+FP^{(c)})}, \\
	\mathrm{Recall}^{(c)}&=\mathrm{TP^{(c)}}/\mathrm{(TP^{(c)}+FN^{(c)})}, \\
	\mathrm{F1}^{(c)}&=2\cdot \frac{\mathrm{Precision}^{(c)}\cdot \mathrm{Recall}^{(c)}}{\mathrm{Precision}^{(c)}+\mathrm{Recall}^{(c)}}, 
\end{align}
where $\mathrm{TP^{(c)}, TN^{(c)}, FP^{(c)}, FN^{(c)}}$ represent the number of pixels that are correctly classified in category $c$, correctly classified in other categories, wrongly classified in category $c$, and wrongly classified in other categories, respectively.

Due to the extensive number of experiment settings, only the most representative evaluation metrics for semantic segmentation (i.e., OA, F1-score) are displayed. For the F1-score, we calculate the average of the metrics among all categories $C$ as the final evaluation measurement (i.e., $\mathrm{F1}=\sum_{c=1}^{C}\mathrm{F1}^{(c)}/C$).

\subsection{Victim Models}
\subsubsection{Scene Classification}
We adopt four widely used scene classification models, AlexNet, ResNet-18, DenseNet-121, and RegNetX-400mf—as victim DNNs to evaluate the performance of the UAD-RS. These variants are chosen for their popularity and effectiveness in many RS studies. The different scales of these models help demonstrate the universal defense performance of the proposed UAD-RS model.

Each model consists of an encoder and a classifier for feature extraction and softmax activation-based prediction, respectively. Initialized with ImageNet pre-trained weights for convergence, each model undergoes 10 epochs of pre-training followed by standard evaluation on the original clean test set. Results are summarized in Table \ref{baseline-sc}.
\subsubsection{Semantic Segmentation}
We employ four popular semantic segmentation models in RS research: FCN-8s, U-Net, PSPNet, and LinkNet. These models use an encoder to extract deep features and a decoder to predict pixel-wise segmentation maps. Initialized with random parameters, we pre-train these models for 100 epochs. The results of the standard evaluation for semantic segmentation are shown in Table \ref{baseline-ss}.
\begin{table}
	\caption{Standard Evaluation Results of the Semantic Segmentation Models.}
	\label{baseline-ss}
	\centering
        \resizebox{\linewidth}{!}{
	\begin{tabular}{ccccccccc}
		\hline
		\multirow{3}{*}{Datasets} & \multicolumn{8}{c}{DNN Classifiers}                                                                               \\
		& \multicolumn{2}{c}{U-Net} & \multicolumn{2}{c}{PSPNet} & \multicolumn{2}{c}{FCN-8s} & \multicolumn{2}{c}{LinkNet} \\
		& OA          & F1          & OA           & F1          & OA           & F1          & OA           & F1           \\ \hline
		Vaihingen                 & 84.19       & 68.85       & 83.59        & 67.10       & 82.61        & 64.63       & 83.19        & 67.12         \\
		Zurich                    & 75.27       & 68.22       & 75.75        & 67.87       & 73.99        & 66.12       & 73.54        & 65.28        \\ \hline
	\end{tabular}}
\end{table}
\subsection{Comparison Methods for Adversarial Defense}
Adversarial purification with unsupervised training is challenging, while most approaches require training based on specific adversarial perturbations. We compare three state-of-the-art methods: Pix2Pix, a GAN-based image-to-image translation model widely used for image enhancement \cite{isola2017image}; Perturbation Seeking Generative Adversarial Networks (PSGAN) \cite{9442932}, a hybrid adversarial defense combining adversarial training and purification; and the task-guided denoising network (TGDN) \cite{xu2022task}, a DNN-based purification approach.

Most adversarial purification methods, like PSGAN, primarily focus on scene classification tasks in RS studies, limiting comparable methods for semantic segmentation experiments. Therefore, for semantic segmentation experiments in this study, we compare only Pix2Pix and TGDN.

\subsection{Implementation Details}
For the proposed UAD-RS model, the generative DDPM was pre-trained under the structure of \textit{diffusers} \cite{von-platen-etal-2022-diffusers} with default training hyperparameters for 500 epochs for each dataset. The number of diffusion steps $T$ in the DDPM was set to $1000$. Regarding the adversarial purification experiments, the noise level $\mathcal{T}$ was set to \{10, 20, 30, ..., 110, 120\} in the ANLS module, and $N=100$ samples were tested with each $T_{m}$ to find the optimal one from the list. As for the adversarial attack part, the implementation of UAE-RS \cite{xu2022universal} GitHub repository was incorporated and utilized in our experiments.

In our experiments, the parameters of the comparison methods were set following the corresponding original articles. All models were implemented using the PyTorch deep learning platform. The experiments were run on the Slurm computational system with eight NVIDIA Tesla A100 GPUs (40 GB of RAM).
	\subsection{Experiments on UCM Dataset}
	\subsubsection{Quantitative Results}
	% \usepackage{multirow}
\begin{table}
		\centering
		\caption{Quantitative comparisons for adversarial defense of scene classification on UCM dataset.}
		\label{UCMresults}
            \resizebox{\linewidth}{!}{
		\begin{tabular}{ccccccccc}
			\hline
			\multirow{2}{*}{Defense}  & \multirow{2}{*}{Victim DNNs} & \multicolumn{7}{c}{Attack Algorithms}                      \\
			&                              & FGSM  & IFGSM & CW    & Jitter & Mixcut & Mixup & TPGD  \\ \hline
			\multirow{4}{*}{None}            & AlexNet                      & 10.95 & 0.00 & 0.00 & 3.05  &  37.05 & 13.14 & 30.48 \\
			& DenseNet-121                  & 56.67& 1.62 & 1.14 & 12.28  & 3.05  & 1.05 & 35.33 \\
			& ResNet-18                     & 33.9 & 0.00 & 0.00 & 6.28  & 8.28  & 3.62 & 35.81 \\
			& RegNetX-400MF             & 34.24 & 0.02 & 0.00 & 4.92  & 23.66  & 24.76 & 29.68 \\ \hline
			\multirow{4}{*}{Pix2Pix}            & AlexNet                      & 17.43 & 0.00 & 0.00 & 13.43   & 52.76 &  21.62  & 31.43 \\
			& DenseNet-121                  & 65.24& 4.19 & 2.57 & 31.71  & 3.71 & 2.19   & 36.95 \\
			& ResNet-18                      & 42.10 & 0.00  & 0.10 & 26.00 & 11.24 & 7.52   & 35.81 \\
			& RegNetX-400MF      & 63.71 & 6.50 & 6.85 & 50.10   & 45.90 & 35.81   & 26.10       \\ \hline
			\multirow{4}{*}{PSGAN}            & AlexNet                      & $\textcolor{red}{84.19}$& 36.95 & 36.76 & $\textcolor{red}{71.52}$   & 46.38 &  29.42  & 41.71 \\
			& DenseNet-121                  & $\textcolor{red}{87.90}$& 29.71 & 30.48 & 80.10  & 6.67 & 4.28   & 53.52 \\
			& ResNet-18                     & $\textcolor{red}{84.10} $& 38.76 & 40.19 & 80.19   & 33.14 & 25.90   & 67.03 \\
			& RegNetX-400MF             & $\textcolor{red}{85.33} $& 44.48  & 49.90 & 83.14 & 65.33 & 54.00   & 55.90 \\ \hline
			\multirow{4}{*}{TGDN}            & AlexNet                      & 46.28 & 30.48 & 31.05 & 53.14  &  $\textcolor{red}{52.10}$ & 44.28 & 48.38 \\
			& DenseNet-121                  & 77.81& 64.76 & 62.57 & 79.43  & 29.62  & 30.10 & 74.48 \\
			& ResNet-18                     & 71.14 & 63.43 & 64.10 & 74.76  & 50.38  & 46.19 & 72.95 \\
			& RegNetX-400MF             & 66.19 & 60.10 & 60.19 & 66.67  & 68.10  & 61.14 & 64.95 \\ \hline
			 & AlexNet          & 65.43 & $\textcolor{red}{53.33}$ & $\textcolor{red}{54.10}$ & 70.10  & 51.24  & $\textcolor{red}{66.28}$ & $\textcolor{red}{64.86}$ \\
			UAD-RS& DenseNet-121      & 81.71 & $\textcolor{red}{68.57}$& $\textcolor{red}{68.19}$ & $\textcolor{red}{82.79}$  & $\textcolor{red}{40.48}$  & $\textcolor{red}{39.05}$ & $\textcolor{red}{75.33}$ \\
			(Ours)& ResNet-18         & 78.00 & $\textcolor{red}{66.48}$& $\textcolor{red}{66.86}$ & $\textcolor{red}{80.48}$  & $\textcolor{red}{54.19}$  & $\textcolor{red}{48.95}$ & $\textcolor{red}{75.52}$ \\
			& RegNetX-400MF & 82.48 & $\textcolor{red}{75.81}$ & $\textcolor{red}{75.81}$ & $\textcolor{red}{84.19}$  & $\textcolor{red}{72.57}$  & $\textcolor{red}{62.19}$ & $\textcolor{red}{80.28}$ \\ \hline
		\end{tabular}
            }
	\end{table}
        \begin{table}
	\centering
	\caption{Quantitative comparisons for adversarial defense of scene classification on AID dataset.}
		\resizebox{\linewidth}{!}{
	\begin{tabular}{ccccccccc}
		\hline
		\multirow{2}{*}{Defense}  & \multirow{2}{*}{Victim DNNs} & \multicolumn{7}{c}{Attack Algorithms}                      \\
		&                              & FGSM  & IFGSM & CW    & Jitter & Mixcut & Mixup & TPGD  \\ \hline
		\multirow{4}{*}{None}            & AlexNet   & 12.82 & 0.00 & 0.00 & 2.32  &  28.36 & 3.64 & 24.10 \\
		& DenseNet-121                  & 46.08& 0.02 & 0.02 & 4.74  & 0.02  & 0.02 & 24.44 \\
		& ResNet-18                     & 15.30 & 0.00 & 0.00 & 3.68  & 2.64  & 0.14 & 34.50 \\
		& RegNetX-400MF             & 34.24 & 0.02 & 0.00 & 4.92  & 23.66  & 24.76 & 29.68 \\ \hline
		\multirow{4}{*}{Pix2Pix}            & AlexNet   & 21.54 & 0.10 & 0.06 & 22.00  & 43.62 & 10.04 & 27.10 \\
		& DenseNet-121                  & 58.96& 3.74 & 2.82 & 43.12  & 0.30  & 0.20 & 27.14 \\
		& ResNet-18                     & 31.12 & 0.28 & 0.24 & 32.62  & 3.84  & 2.70 & 34.88 \\
		& RegNetX-400MF        & 51.90 & 4.78  & 4.96  & 49.78  & 40.70 & 36.50 &34.40      \\ \hline
		\multirow{4}{*}{PSGAN}            & AlexNet   & $\textcolor{red}{79.96}$ & 42.46 & 42.24 & 69.08  & 33.92 & 16.84 & 43.54 \\
		& DenseNet-121                  & $\textcolor{red}{83.12}$& 58.22 & 58.00 & $\textcolor{red}{82.62}$  & 18.48  & 18.46 & $\textcolor{red}{69.50}$ \\
		& ResNet-18                     & $\textcolor{red}{80.52}$ & $\textcolor{red}{61.26}$  & $\textcolor{red}{60.86}$  & $\textcolor{red}{81.76}$  & 33.16 & 29.60 &$\textcolor{red}{72.60}$ \\
		& RegNetX-400MF             & $\textcolor{red}{82.08}$ & 50.56 & 54.18 & $\textcolor{red}{80.58}$  & 57.78  & 52.42 & 65.22 \\ \hline
		\multirow{4}{*}{TGDN}            & AlexNet    & 26.42 & 18.24 & 18.36 & 28.78  &  22.88 & 18.28 & 27.30 \\
		& DenseNet-121                  & 41.96& 34.08 & 33.86 & 41.94  & 14.46  & 14.18 & 38.74 \\
		& ResNet-18                     & 32.12 & 27.70 & 28.68 & 34.72  & 19.76  & 18.28 & 33.90 \\
		& RegNetX-400MF             & 34.44 & 30.14 & 30.66 & 35.84  & 35.24  & 32.16 & 33.78 \\ \hline
		 & AlexNet          & 68.68 & $\textcolor{red}{57.12}$& $\textcolor{red}{57.40}$& $\textcolor{red}{72.56}$  & $\textcolor{red}{47.98}$ & $\textcolor{red}{30.12}$ & $\textcolor{red}{59.16}$ \\
		UAD-RS& DenseNet-121      & 75.02& $\textcolor{red}{59.36}$ & $\textcolor{red}{59.08}$ & 75.36  & $\textcolor{red}{31.54}$  & $\textcolor{red}{32.10}$ & 63.04 \\
		(Ours)& ResNet-18         & 66.52 & 56.44 & 57.34 & 69.28 & $\textcolor{red}{34.84}$  & $\textcolor{red}{34.50}$ & 62.16 \\
		& RegNetX-400MF & 76.88 & $\textcolor{red}{67.72} $& $\textcolor{red}{68.20}$ &78.50  & $\textcolor{red}{59.42}$  & $\textcolor{red}{53.28}$ & $\textcolor{red}{73.00}$ \\ \hline
	\end{tabular}
		}
	\label{AIDresults}
\end{table}
 The quantitative evaluation of the proposed UAD-RS method and its competitors for scene classification on the UCM dataset is shown in Table \ref{UCMresults}. The UAD-RS model achieved the best adversarial defense performance with the highest OA across most classifiers and attacks. Although PSGAN achieved promising results against FGSM attacks, it suffers significant performance loss against more complex and stronger attacks, especially Mixcut and Mixup. The TGDN method outperforms the other two comparators in some cases but still lags significantly behind UAD-RS. Moreover, Pix2Pix does not provide satisfactory defense results, offering only slight improvements in classifier performance against most attacks.

\subsubsection{Qualitative Results}
\begin{figure}
		\centering
		\begin{minipage}{0.16\linewidth}
			\vspace{3pt}
			\centerline{\includegraphics[width=\textwidth]{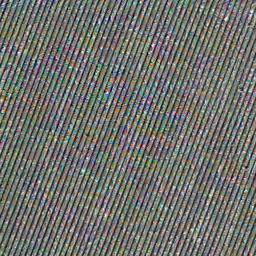}}
			\vspace{3pt}
			\centerline{\includegraphics[width=\textwidth]{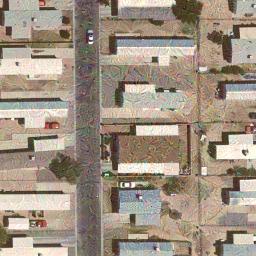}}
			\vspace{3pt}
			\centerline{\includegraphics[width=\textwidth]{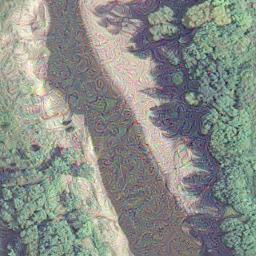}}
			\vspace{3pt}
			\centerline{\includegraphics[width=\textwidth]{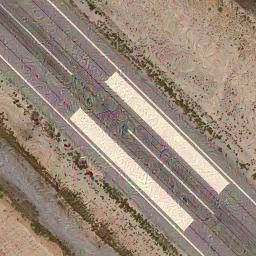}}
			\vspace{3pt}
			\centerline{\includegraphics[width=\textwidth]{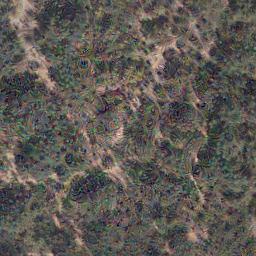}}
			\vspace{3pt}
			\centerline{\includegraphics[width=\textwidth]{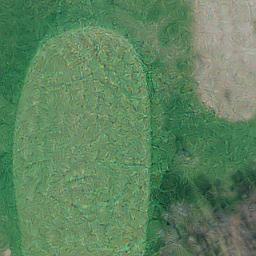}}
			\vspace{3pt}
			\centerline{(a)}
		\end{minipage}
		\hspace{-5pt}
		\begin{minipage}{0.16\linewidth}
			\vspace{3pt}
			\centerline{\includegraphics[width=\textwidth]{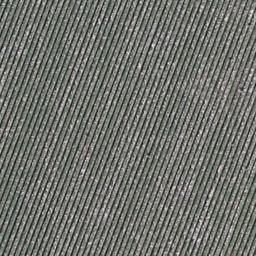}}
			\vspace{3pt}
			\centerline{\includegraphics[width=\textwidth]{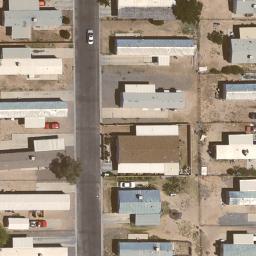}}
			\vspace{3pt}
			\centerline{\includegraphics[width=\textwidth]{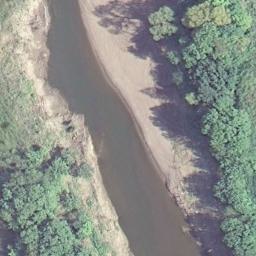}}
			\vspace{3pt}
			\centerline{\includegraphics[width=\textwidth]{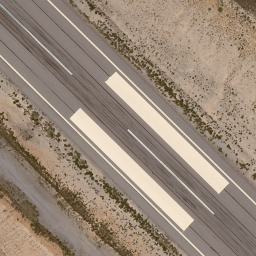}}
			\vspace{3pt}
			\centerline{\includegraphics[width=\textwidth]{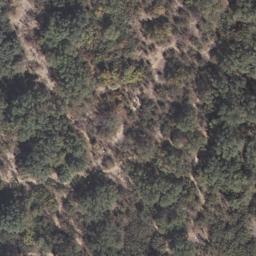}}
			\vspace{3pt}
			\centerline{\includegraphics[width=\textwidth]{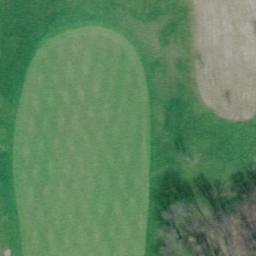}}
			\vspace{3pt}
			\centerline{(b)}
		\end{minipage}
		\hspace{-5pt}
		\begin{minipage}{0.16\linewidth}
			\vspace{3pt}
			\centerline{\includegraphics[width=\textwidth]{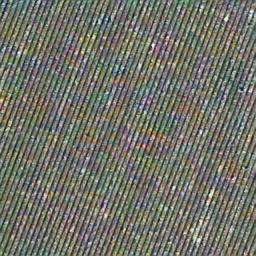}}
			\vspace{3pt}
			\centerline{\includegraphics[width=\textwidth]{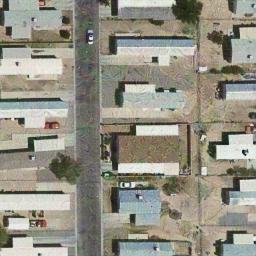}}
			\vspace{3pt}
			\centerline{\includegraphics[width=\textwidth]{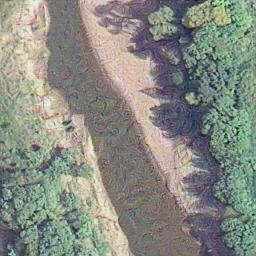}}
			\vspace{3pt}
			\centerline{\includegraphics[width=\textwidth]{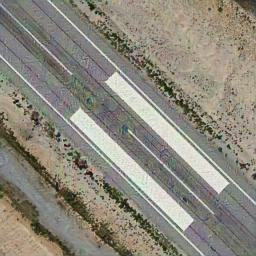}}
			\vspace{3pt}
			\centerline{\includegraphics[width=\textwidth]{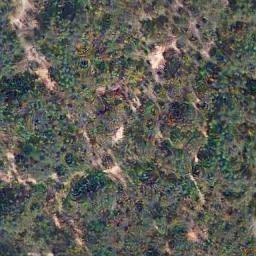}}
			\vspace{3pt}
			\centerline{\includegraphics[width=\textwidth]{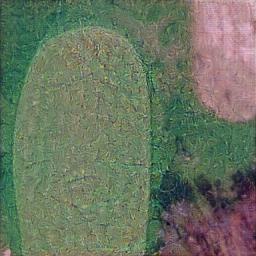}}
			\vspace{3pt}
			\centerline{(c)}
		\end{minipage}
		\hspace{-5pt}
		\begin{minipage}{0.16\linewidth}
			\vspace{3pt}
			\centerline{\includegraphics[width=\textwidth]{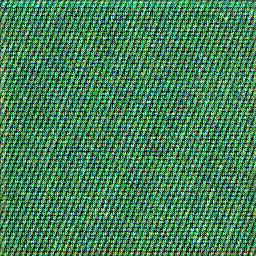}}
			\vspace{3pt}
			\centerline{\includegraphics[width=\textwidth]{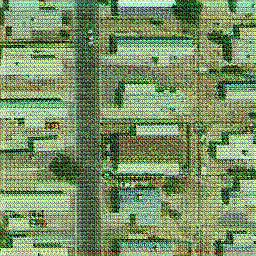}}
			\vspace{3pt}
			\centerline{\includegraphics[width=\textwidth]{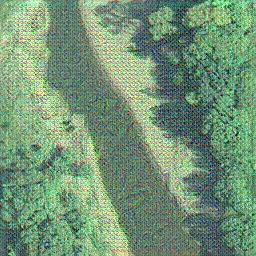}}
			\vspace{3pt}
			\centerline{\includegraphics[width=\textwidth]{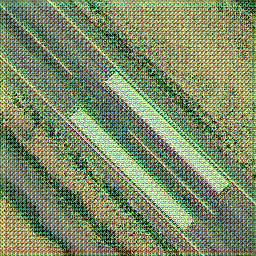}}
			\vspace{3pt}
			\centerline{\includegraphics[width=\textwidth]{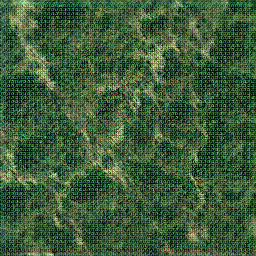}}
			\vspace{3pt}
			\centerline{\includegraphics[width=\textwidth]{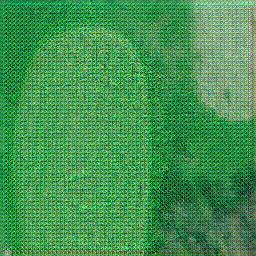}}
			\vspace{3pt}
			\centerline{(d)}
		\end{minipage}
		\hspace{-5pt}
		\begin{minipage}{0.16\linewidth}
			\vspace{3pt}
			\centerline{\includegraphics[width=\textwidth]{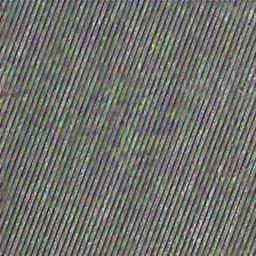}}
			\vspace{3pt}
			\centerline{\includegraphics[width=\textwidth]{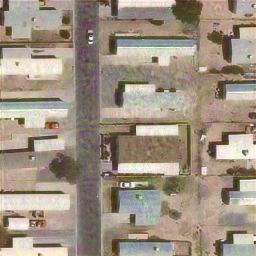}}
			\vspace{3pt}
			\centerline{\includegraphics[width=\textwidth]{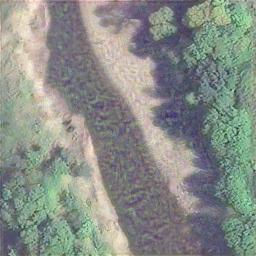}}
			\vspace{3pt}
			\centerline{\includegraphics[width=\textwidth]{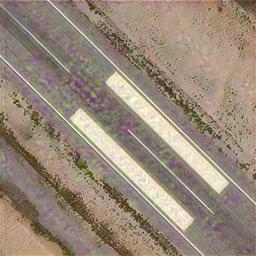}}
			\vspace{3pt}
			\centerline{\includegraphics[width=\textwidth]{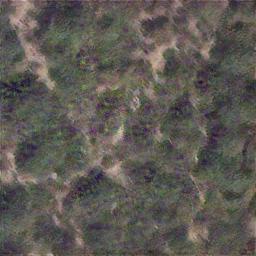}}
			\vspace{3pt}
			\centerline{\includegraphics[width=\textwidth]{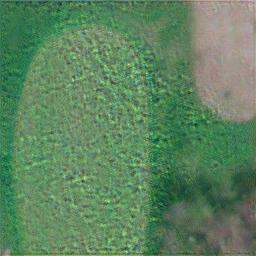}}
			\vspace{3pt}
			\centerline{(e)}
		\end{minipage}
		\hspace{-5pt}
		\begin{minipage}{0.16\linewidth}
			\vspace{3pt}
			\centerline{\includegraphics[width=\textwidth]{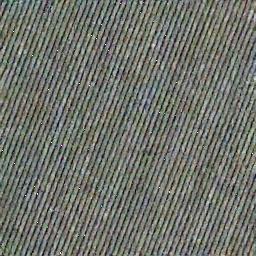}}
			\vspace{3pt}
			\centerline{\includegraphics[width=\textwidth]{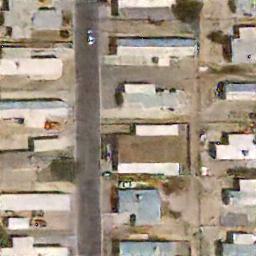}}
			\vspace{3pt}
			\centerline{\includegraphics[width=\textwidth]{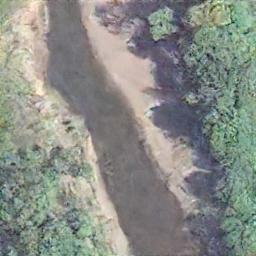}}
			\vspace{3pt}
			\centerline{\includegraphics[width=\textwidth]{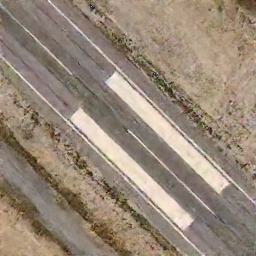}}
			\vspace{3pt}
			\centerline{\includegraphics[width=\textwidth]{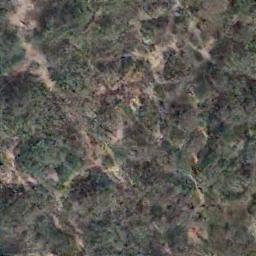}}
			\vspace{3pt}
			\centerline{\includegraphics[width=\textwidth]{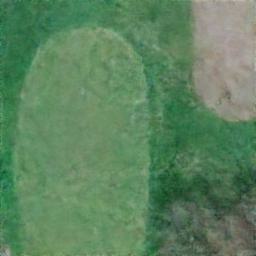}}
			\vspace{3pt}
			\centerline{(f)}
		\end{minipage}
		\caption{Qualitative comparison for adversarial purification on the UCM dataset. (a) Adversarial samples. (b) Ground truth. (c)-(f) Purified results obtained by (c) Pix2Pix. (d) PSGAN. (e) TGDN. (f) UAD-RS.} 
		\label{UCMQuali}
	\end{figure}

Fig. \ref{UCMQuali} presents six examples of adversarial purification results from various methods applied to the UCM dataset. These adversarial samples were generated using robust attacks (e.g., CW, Mixup, Mixcut) to facilitate a clear comparison. The results show that Pix2Pix still exhibits some adversarial perturbations and suffers from color distortion across the entire image. PSGAN's results display a significant green mask in several patches, likely due to the cross-entropy loss guidance used during auxiliary adversarial training. Although TGDN preserves most background features, it loses texture information, resulting in blurred images, particularly noticeable in features such as rivers and plants. In contrast, the proposed UAD-RS method achieves finer spatial detail in the background and maintains better spectral consistency with the ground truth.
	\subsection{Experiments on AID Dataset}
\subsubsection{Quantitative Results}
The proposed UAD-RS demonstrates superior performance in most adversarial attack and classifier combinations on the AID dataset, as shown in Table \ref{AIDresults}. Although PSGAN achieves marginally better results than UAD-RS in some instances, particularly with the ResNet-18 classifier, it struggles to deliver effective performance against higher-intensity attacks on other classifiers. In contrast, the Pix2Pix and TGDN methods consistently fail to provide reliable defenses for a large-scale RS dataset like AID.

\subsubsection{Qualitative Results}
\begin{figure}[]
	\centering
	\begin{minipage}{0.16\linewidth}
		\vspace{3pt}
		\centerline{\includegraphics[width=\textwidth]{figures/Defense/Scene_Classification/AID/Adv/airport_22.png}}
		\vspace{3pt}
		\centerline{\includegraphics[width=\textwidth]{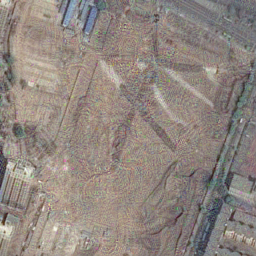}}
		\vspace{3pt}
		\centerline{\includegraphics[width=\textwidth]{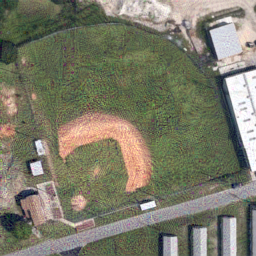}}
		\vspace{3pt}
		\centerline{\includegraphics[width=\textwidth]{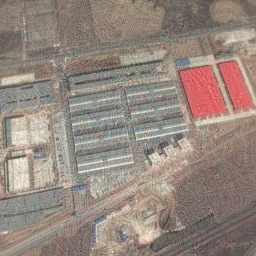}}
		\vspace{3pt}
		\centerline{\includegraphics[width=\textwidth]{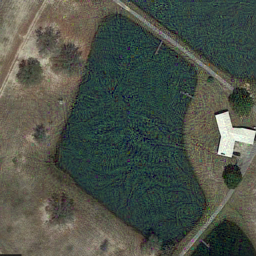}}
		\vspace{3pt}
		\centerline{\includegraphics[width=\textwidth]{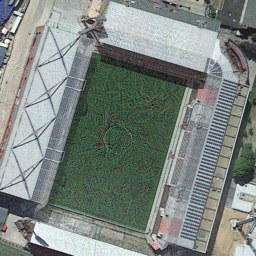}}
		\vspace{3pt}
		\centerline{(a)}
	\end{minipage}
	\hspace{-5pt}
	\begin{minipage}{0.16\linewidth}
		\vspace{3pt}
		\centerline{\includegraphics[width=\textwidth]{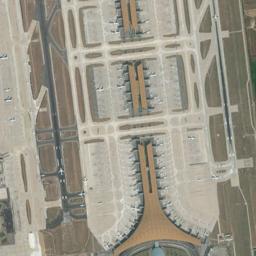}}
		\vspace{3pt}
		\centerline{\includegraphics[width=\textwidth]{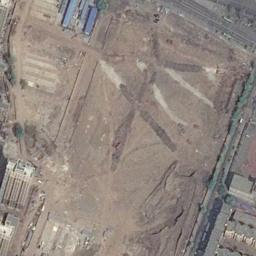}}
		\vspace{3pt}
		\centerline{\includegraphics[width=\textwidth]{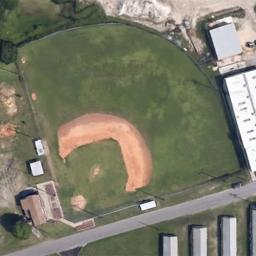}}
		\vspace{3pt}
		\centerline{\includegraphics[width=\textwidth]{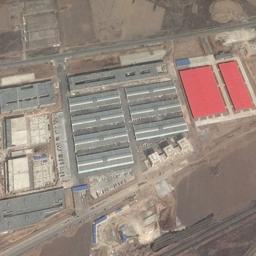}}
		\vspace{3pt}
		\centerline{\includegraphics[width=\textwidth]{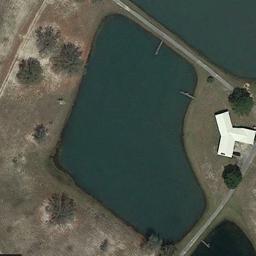}}
		\vspace{3pt}
		\centerline{\includegraphics[width=\textwidth]{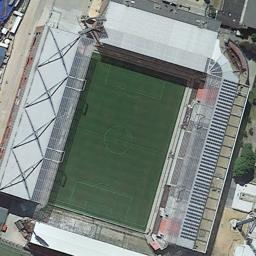}}
		\vspace{3pt}
		\centerline{(b)}
	\end{minipage}
	\hspace{-5pt}
	\begin{minipage}{0.16\linewidth}
		\vspace{3pt}
		\centerline{\includegraphics[width=\textwidth]{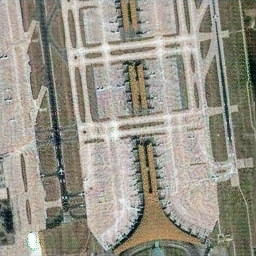}}
		\vspace{3pt}
		\centerline{\includegraphics[width=\textwidth]{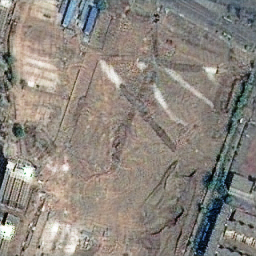}}
		\vspace{3pt}
		\centerline{\includegraphics[width=\textwidth]{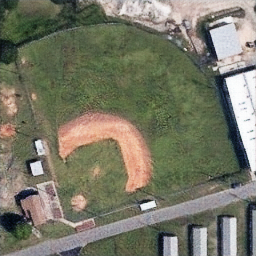}}
		\vspace{3pt}
		\centerline{\includegraphics[width=\textwidth]{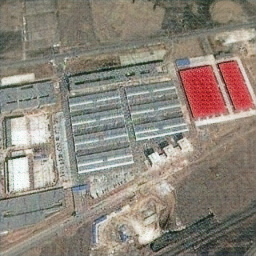}}
		\vspace{3pt}
		\centerline{\includegraphics[width=\textwidth]{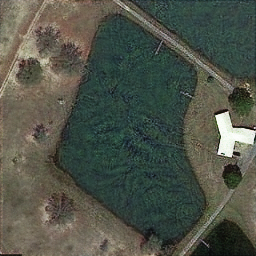}}
		\vspace{3pt}
		\centerline{\includegraphics[width=\textwidth]{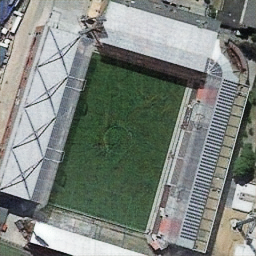}}
		\vspace{3pt}
		\centerline{(c)}
	\end{minipage}
	\hspace{-5pt}
	\begin{minipage}{0.16\linewidth}
		\vspace{3pt}
		\centerline{\includegraphics[width=\textwidth]{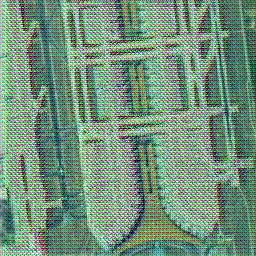}}
		\vspace{3pt}
		\centerline{\includegraphics[width=\textwidth]{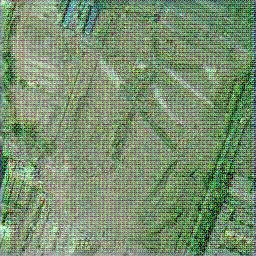}}
		\vspace{3pt}
		\centerline{\includegraphics[width=\textwidth]{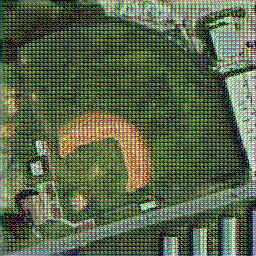}}
		\vspace{3pt}
		\centerline{\includegraphics[width=\textwidth]{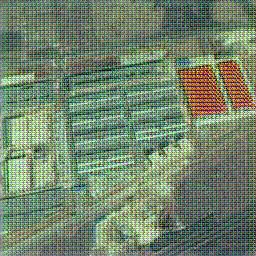}}
		\vspace{3pt}
		\centerline{\includegraphics[width=\textwidth]{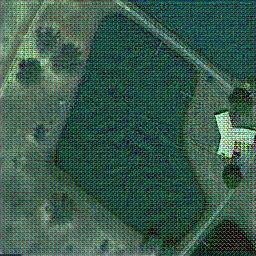}}
		\vspace{3pt}
		\centerline{\includegraphics[width=\textwidth]{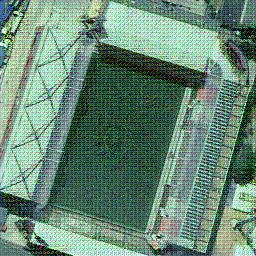}}
		\vspace{3pt}
		\centerline{(d)}
	\end{minipage}
	\hspace{-5pt}
	\begin{minipage}{0.16\linewidth}
		\vspace{3pt}
		\centerline{\includegraphics[width=\textwidth]{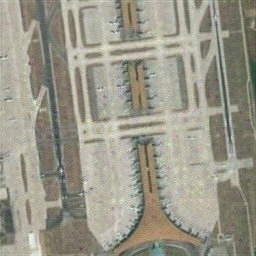}}
		\vspace{3pt}
		\centerline{\includegraphics[width=\textwidth]{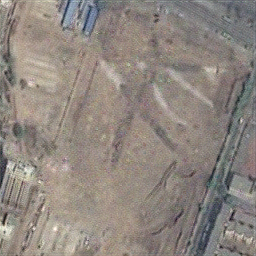}}
		\vspace{3pt}
		\centerline{\includegraphics[width=\textwidth]{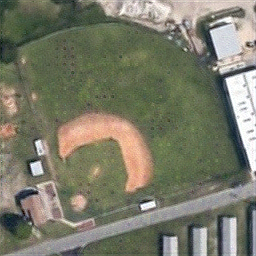}}
		\vspace{3pt}
		\centerline{\includegraphics[width=\textwidth]{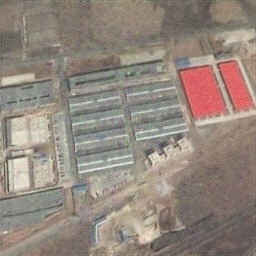}}
		\vspace{3pt}
		\centerline{\includegraphics[width=\textwidth]{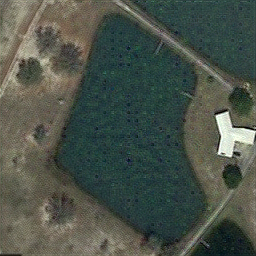}}
		\vspace{3pt}
		\centerline{\includegraphics[width=\textwidth]{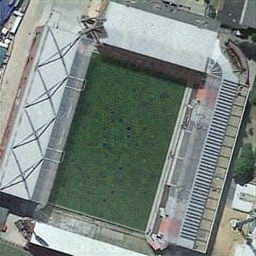}}
		\vspace{3pt}
		\centerline{(e)}
	\end{minipage}
	\hspace{-5pt}
	\begin{minipage}{0.16\linewidth}
		\vspace{3pt}
		\centerline{\includegraphics[width=\textwidth]{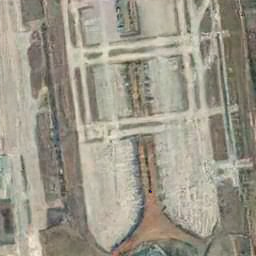}}
		\vspace{3pt}
		\centerline{\includegraphics[width=\textwidth]{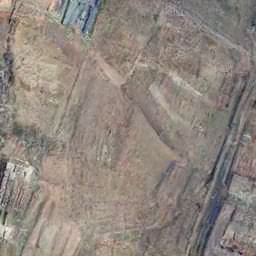}}
		\vspace{3pt}
		\centerline{\includegraphics[width=\textwidth]{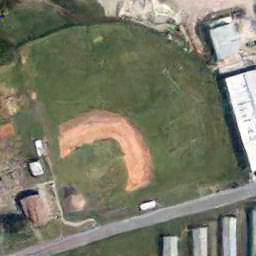}}
		\vspace{3pt}
		\centerline{\includegraphics[width=\textwidth]{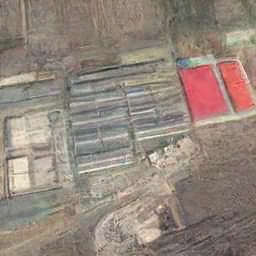}}
		\vspace{3pt}
		\centerline{\includegraphics[width=\textwidth]{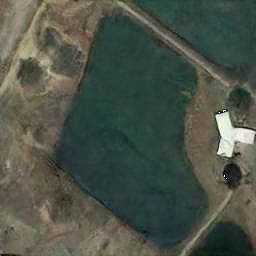}}
		\vspace{3pt}
		\centerline{\includegraphics[width=\textwidth]{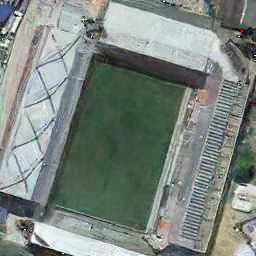}}
		\vspace{3pt}
		\centerline{(f)}
	\end{minipage}
	\caption{Qualitative comparison for adversarial purification on the AID dataset. (a) Adversarial samples. (b) Ground truth. (c)-(f) Purified results obtained by (c) Pix2Pix. (d) PSGAN. (e) TGDN. (f) UAD-RS.} 
	\label{AIDQuali}
\end{figure}

The qualitative results of UAD-RS and the comparison methods for adversarial purification on the AID dataset are presented in Fig. \ref{AIDQuali}. Pix2Pix often fails to remove adversarial perturbations, sometimes integrating them into the texture features. PSGAN tends to produce a color mask effect, failing to accurately restore the original images. While TGDN generates visually acceptable results, it introduces slight noise artifacts, such as black dots, particularly in flat areas. In contrast, UAD-RS delivers the best qualitative results, closely resembling the ground truth and successfully restoring key object information.
 \subsection{Experiments on Vaihingen Dataset}
\subsubsection{Quantitative Results}
Table \ref{Vaihingenquanti} presents the quantitative comparison for semantic segmentation on the Vaihingen dataset. Among the applied attacks, FGSM and Jitter have a minor impact on the performance of segmentation DNNs, while the stronger attacks significantly degrade the segmentation results. The UAD-RS model outperforms all competitors, achieving the highest OA and F1 metrics for adversarial sample purification. The Pix2Pix method offers protection against mild attacks but shows limited improvement against stronger attacks. TGDN effectively defends against most adversarial attacks but falls short in providing satisfactory purification, particularly against intensive attacks.

\begin{table*}[]
	\centering
	\caption{Quantitative comparisons for adversarial defense of semantic segmentation on Vaihingen dataset.}
	\label{Vaihingenquanti}
	\setlength\tabcolsep{5pt}
 \resizebox{\textwidth}{!}{
	\begin{tabular}{cccccccccccccccc}
		\hline
		\multirow{3}{*}{Datasets}        & \multirow{3}{*}{Victims DNNs} & \multicolumn{14}{c}{Attack Algorithms}                                                                                                                                                                                                         \\
		&                               & \multicolumn{2}{c}{FGSM}        & \multicolumn{2}{c}{IFGSM}       & \multicolumn{2}{c}{CW}          & \multicolumn{2}{c}{Jitter}      & \multicolumn{2}{c}{Mixcut}      & \multicolumn{2}{c}{Mixup}       & \multicolumn{2}{c}{TPGD}        \\
		&                               & OA             & F1             & OA             & F1             & OA             & F1             & OA             & F1             & OA             & F1             & OA             & F1             & OA             & F1             \\ \hline
		\multirow{4}{*}{No Defense}      & U-Net                         & 73.11          & 58.59          & 33.63          & 21.59          & 17.89          & 10.89          & 69.67          & 55.67          & 22.00          & 16.50          & 24.45          & 16.62          & 40.65          & 28.36          \\
		& PSPNet                        & 78.87          & 62.83          & 64.28          & 49.36          & 59.22          & 46.01          & 77.96          & 62.05          & 47.66          & 30.53          & 48.20          & 30.88          & 70.62          & 55.13          \\
		& FCN-8s                        & 72.45          & 54.38          & 40.20          & 27.01          & 17.93          & 11.13          & 72.31          & 54.29          & 38.12          & 25.29          & 36.66          & 24.34          & 54.06          & 38.78          \\
		& LinkNet                       & 75.8           & 59.39          & 43.28          & 30.11          & 28.86          & 19.82          & 74.93          & 58.88          & 40.61          & 29.58          & 33.11          & 21.15          & 50.34          & 36.41          \\ \hline
		\multirow{4}{*}{Pix2Pix}         & U-Net                         & 71.43          & 55.17          & 52.30           & 37.18          & 45.43          & 30.66          & 71.20          & 54.97          & 44.79          & 30.36          & 41.92          & 27.15          & 57.85          & 42.45          \\
		& PSPNet                        & 79.03          & 63.02          & 66.95          & 52.41          & 63.06          & 49.31          & 78.79          & 62.72          & 55.76          & 37.90          & 53.25          & 35.16          & 70.72          & 56.02          \\
		& FCN-8s                        & 72.89          & 55.34          & 48.78          & 31.96          & 35.20          & 20.92          & 73.10          & 55.43          & 47.15          & 30.14          & 46.43          & 30.57          & 58.84          & 41.36          \\
		& LinkNet                       & 78.56          & 62.39          & 61.27          & 47.22          & 55.49          & 42.91          & 78.35          & 62.21          & 62.32          & 47.27          & 57.00          & 41.74          & 66.42          & 51.85          \\ \hline
		\multirow{4}{*}{TGDN}            & U-Net                         & 66.67          & 52.86          & 53.00          & 39.15          & 46.54          & 32.08          & 66.29          & 52.45          & 49.21          & \textcolor{red}{40.51} & 38.28          & 24.39          & 55.61          & 41.67          \\
		& PSPNet                        & 80.85          & 64.71          & 67.78          & 52.54          & 64.10          & 48.19          & 80.37          & 64.20           & 59.01          & 40.60          & 56.66          & 38.34          & 70.48          & 55.35          \\
		& FCN-8s                        & 63.26          & 48.38          & 51.74          & 34.86          & 41.87          & 25.89          & 65.20          & 50.05          & 48.18          & 30.87          & 48.22          & 31.55          & 57.91          & 41.39          \\
		& LinkNet                       & 69.85          & 54.25          & 59.65          & 44.54          & 56.46          & 42.70          & 69.69          & 54.09          & 64.42          & 50.20          & 60.53          & \textcolor{red}{46.69} & 61.50          & 46.13          \\ \hline
		 & U-Net                         & \textcolor{red}{77.86} & \textcolor{red}{62.47} & \textcolor{red}{70.62} & \textcolor{red}{56.76} & \textcolor{red}{66.90} & \textcolor{red}{52.53} & \textcolor{red}{77.67} & \textcolor{red}{62.53} & \textcolor{red}{54.83} & 39.12          & \textcolor{red}{56.82} & \textcolor{red}{40.59} & \textcolor{red}{74.49} & \textcolor{red}{60.02} \\
		UAD-RS& PSPNet                        & \textcolor{red}{81.80} & \textcolor{red}{65.24} & \textcolor{red}{79.54} & \textcolor{red}{63.48} & \textcolor{red}{77.81} & \textcolor{red}{61.95} & \textcolor{red}{81.75} & \textcolor{red}{65.20} & \textcolor{red}{70.60} & \textcolor{red}{51.72} & \textcolor{red}{72.15} & \textcolor{red}{53.99} & \textcolor{red}{81.68} & \textcolor{red}{65.13} \\
		(Ours)& FCN-8s                        & \textcolor{red}{75.21} & \textcolor{red}{57.33} & \textcolor{red}{68.74} & \textcolor{red}{50.77} & \textcolor{red}{58.16} & \textcolor{red}{37.03} & \textcolor{red}{74.61} & \textcolor{red}{56.88} & \textcolor{red}{61.64} & \textcolor{red}{43.57} & \textcolor{red}{62.83} & \textcolor{red}{44.76} & \textcolor{red}{75.02} & \textcolor{red}{56.67} \\
		& LinkNet                       & \textcolor{red}{81.34} & \textcolor{red}{65.32} & \textcolor{red}{71.74} & \textcolor{red}{56.53} & \textcolor{red}{75.81} & \textcolor{red}{60.51} & \textcolor{red}{81.34} & \textcolor{red}{65.37} & \textcolor{red}{71.22} & \textcolor{red}{54.22} & \textcolor{red}{61.03} & 44.34          & \textcolor{red}{80.10} & \textcolor{red}{64.42} \\ \hline
	\end{tabular}}
\end{table*}
 \subsubsection{Qualitative Results}
\begin{figure*}[]
		\centering
		\includegraphics[page=1,scale=0.8]{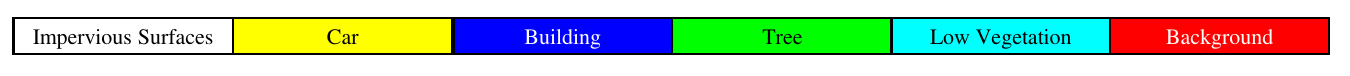}
		\begin{minipage}{0.10\linewidth}
			\vspace{3pt}
			\centerline{\includegraphics[width=\textwidth, angle=90]{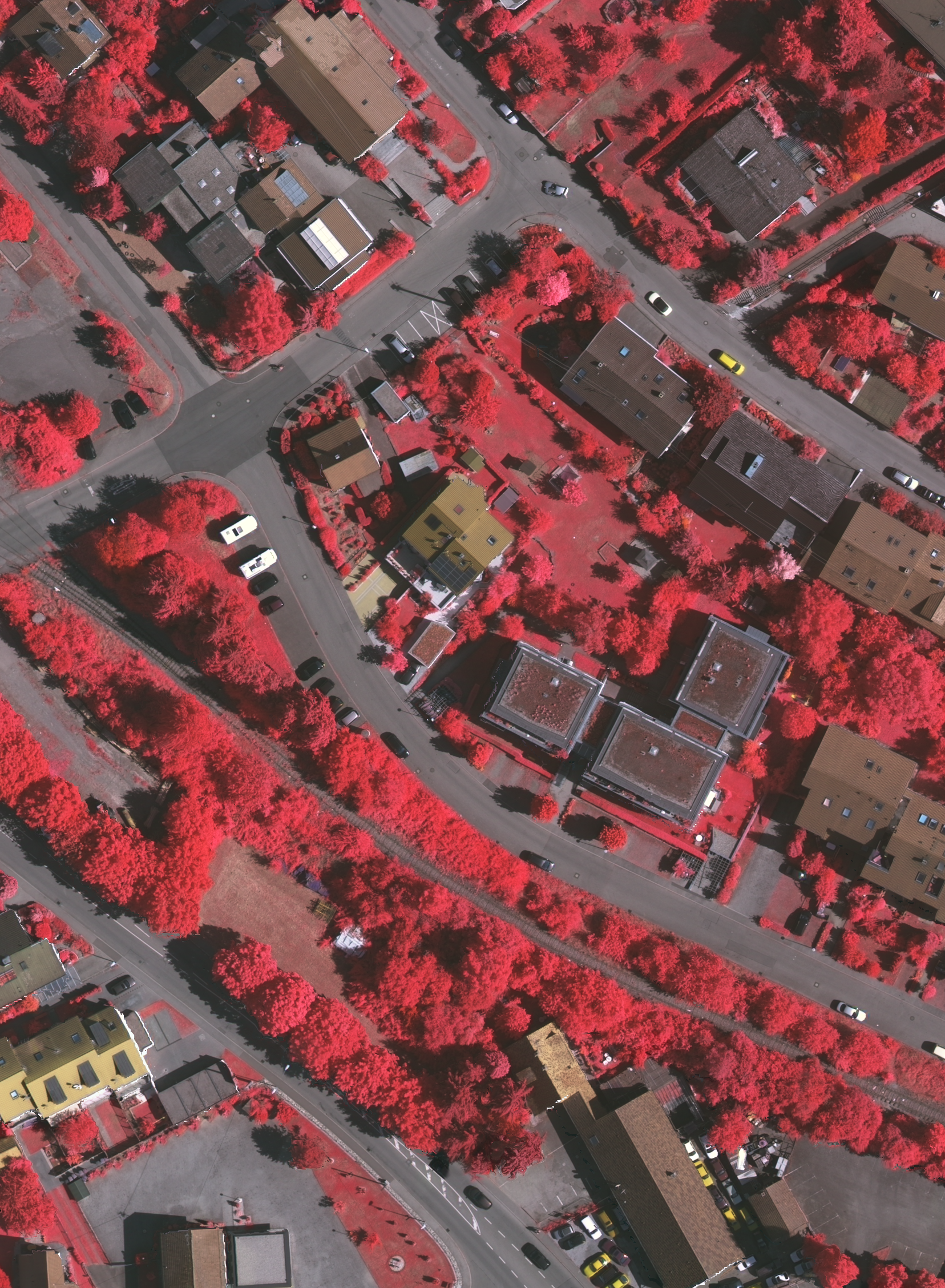}}
			\vspace{3pt}
			\centerline{\includegraphics[width=\textwidth, angle=90]{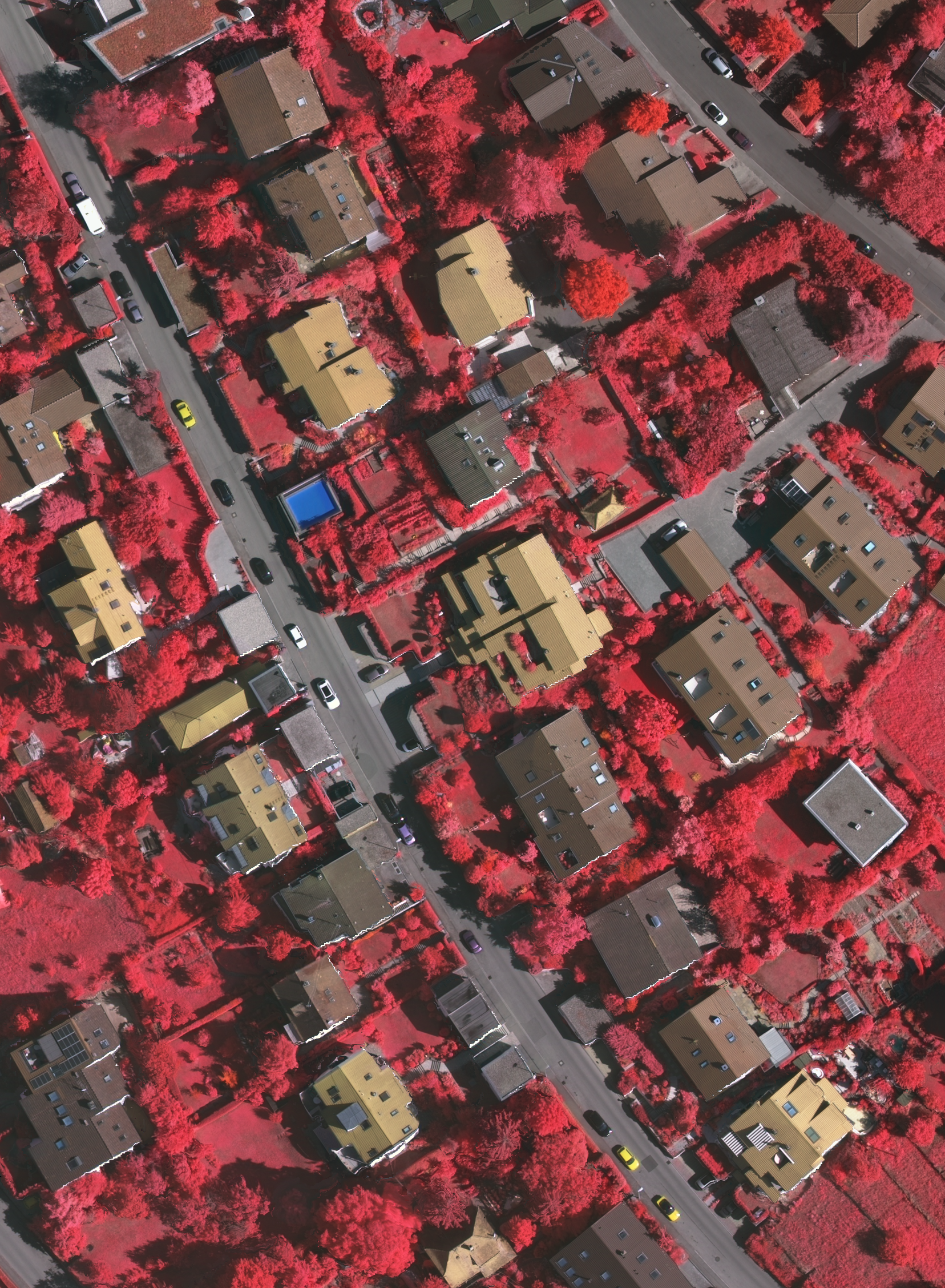}}
			\vspace{3pt}
			\centerline{\includegraphics[width=\textwidth, angle=90]{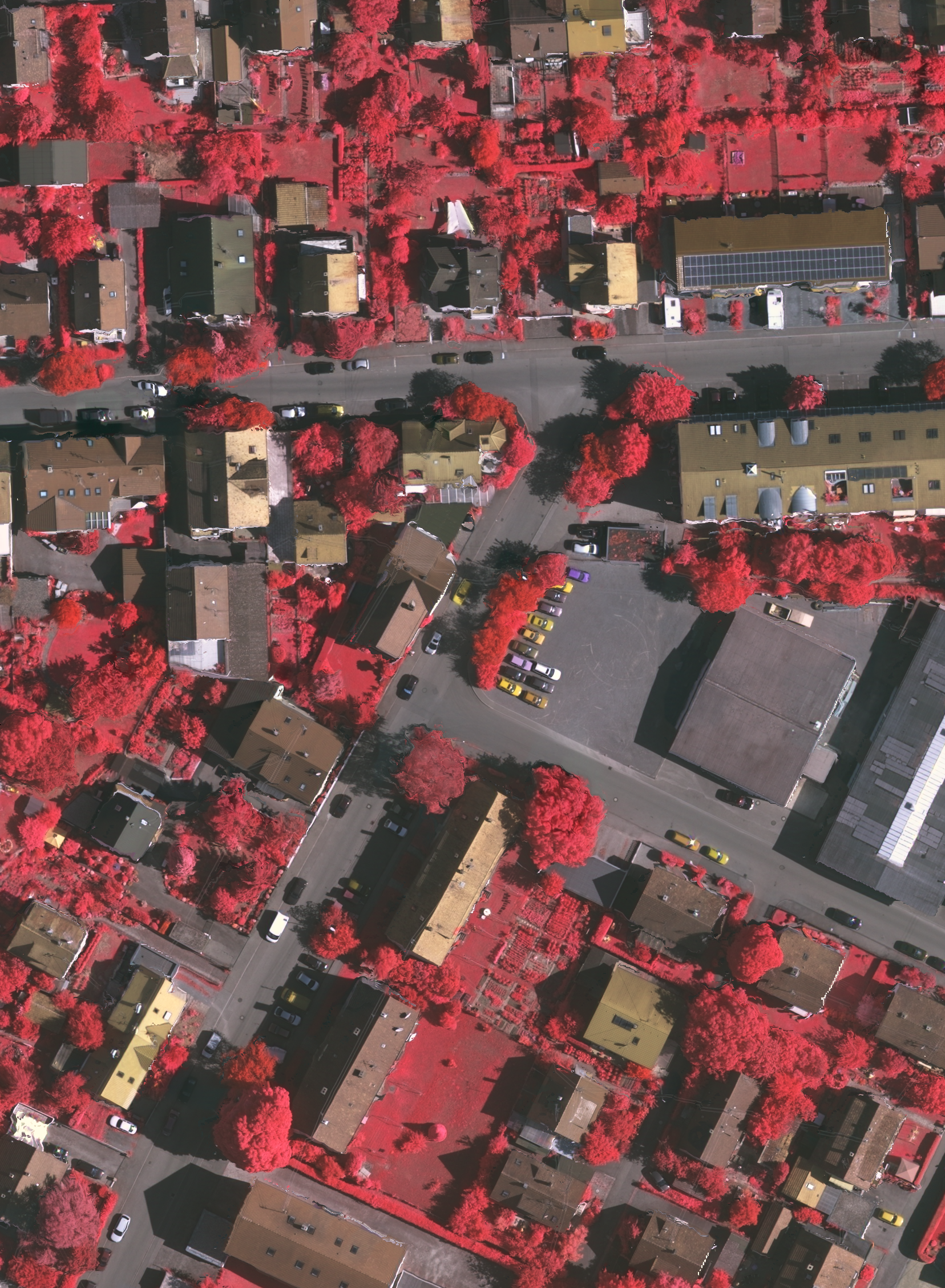}}
			\vspace{3pt}
			\centerline{\includegraphics[width=\textwidth, angle=90]{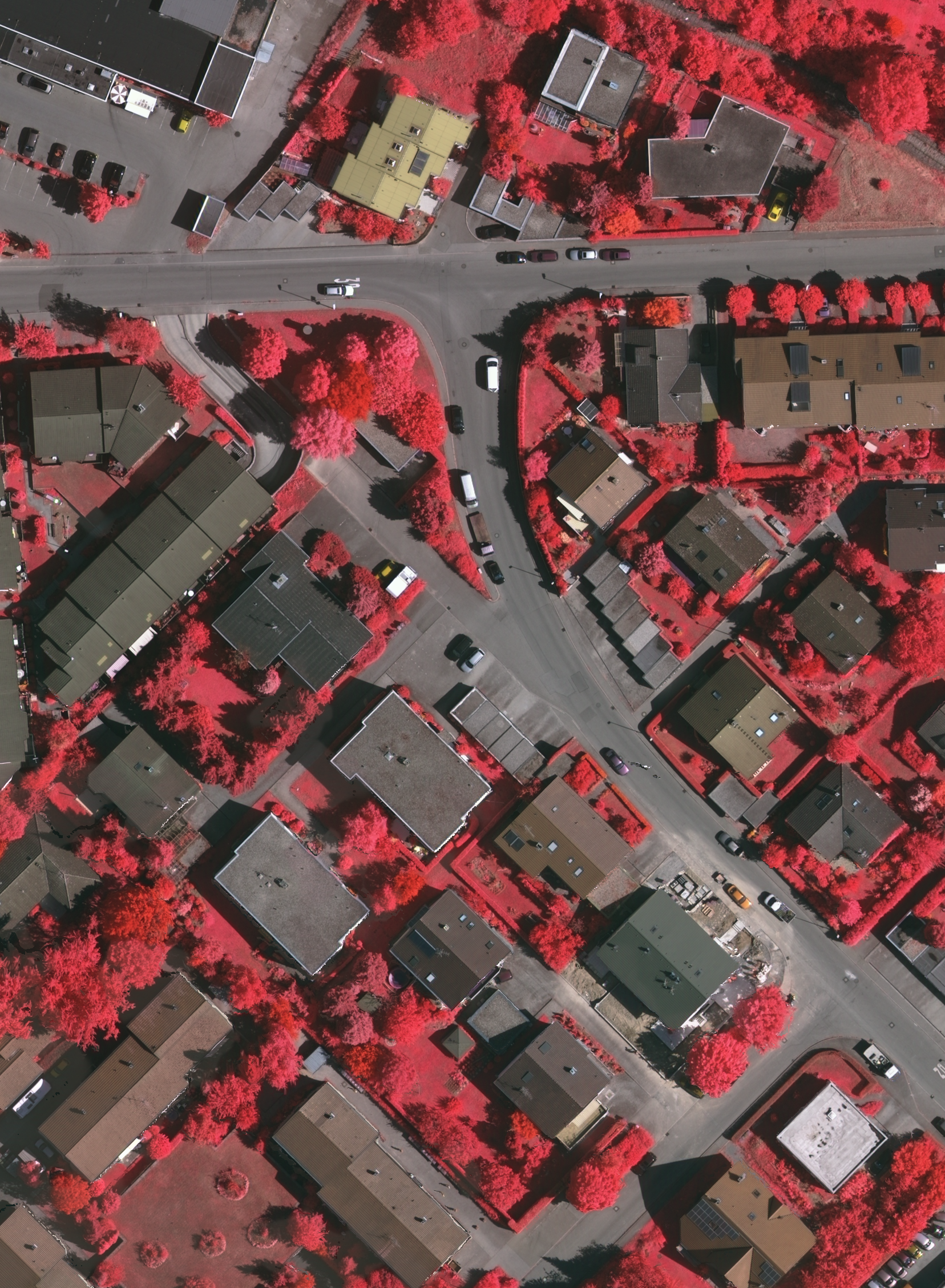}}
			\vspace{3pt}
			\centerline{\includegraphics[scale=0.031, angle=90]{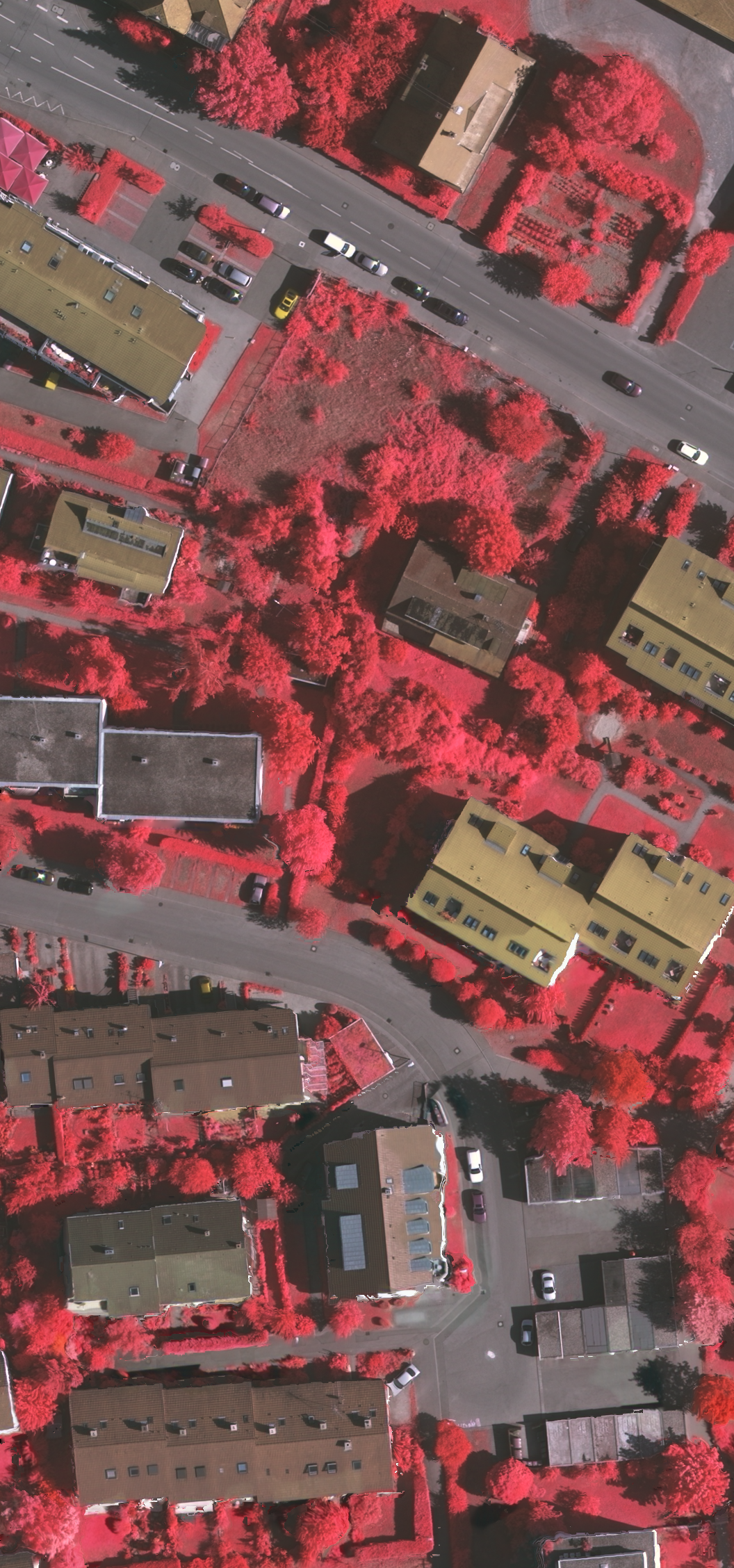}}
			\vspace{3pt}
			\centerline{(a)}
		\end{minipage}
		\hspace{15pt}
		\begin{minipage}{0.10\linewidth}
			\vspace{3pt}
			\centerline{\includegraphics[width=\textwidth, angle=90]{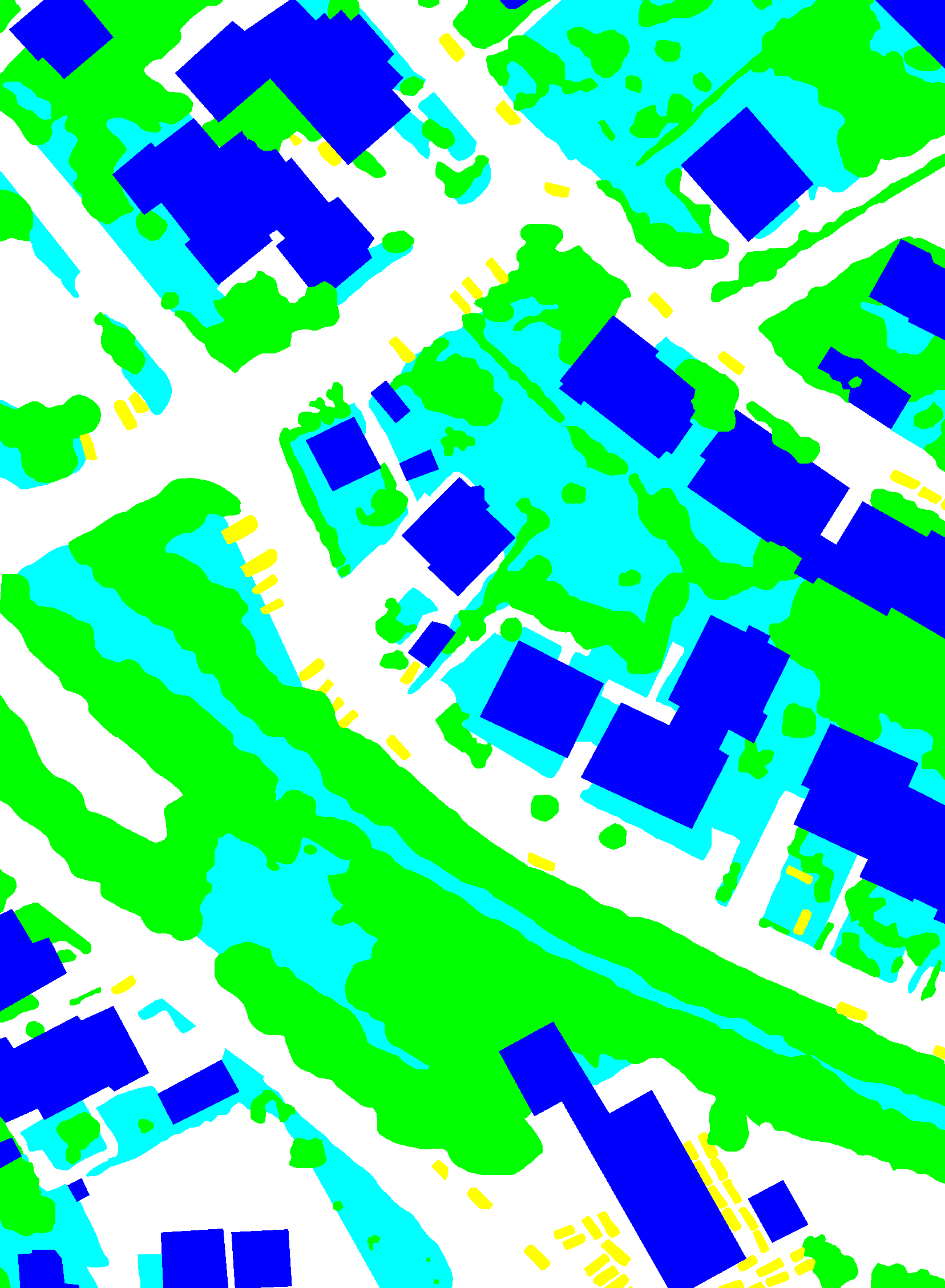}}
			\vspace{3pt}
			\centerline{\includegraphics[width=\textwidth, angle=90]{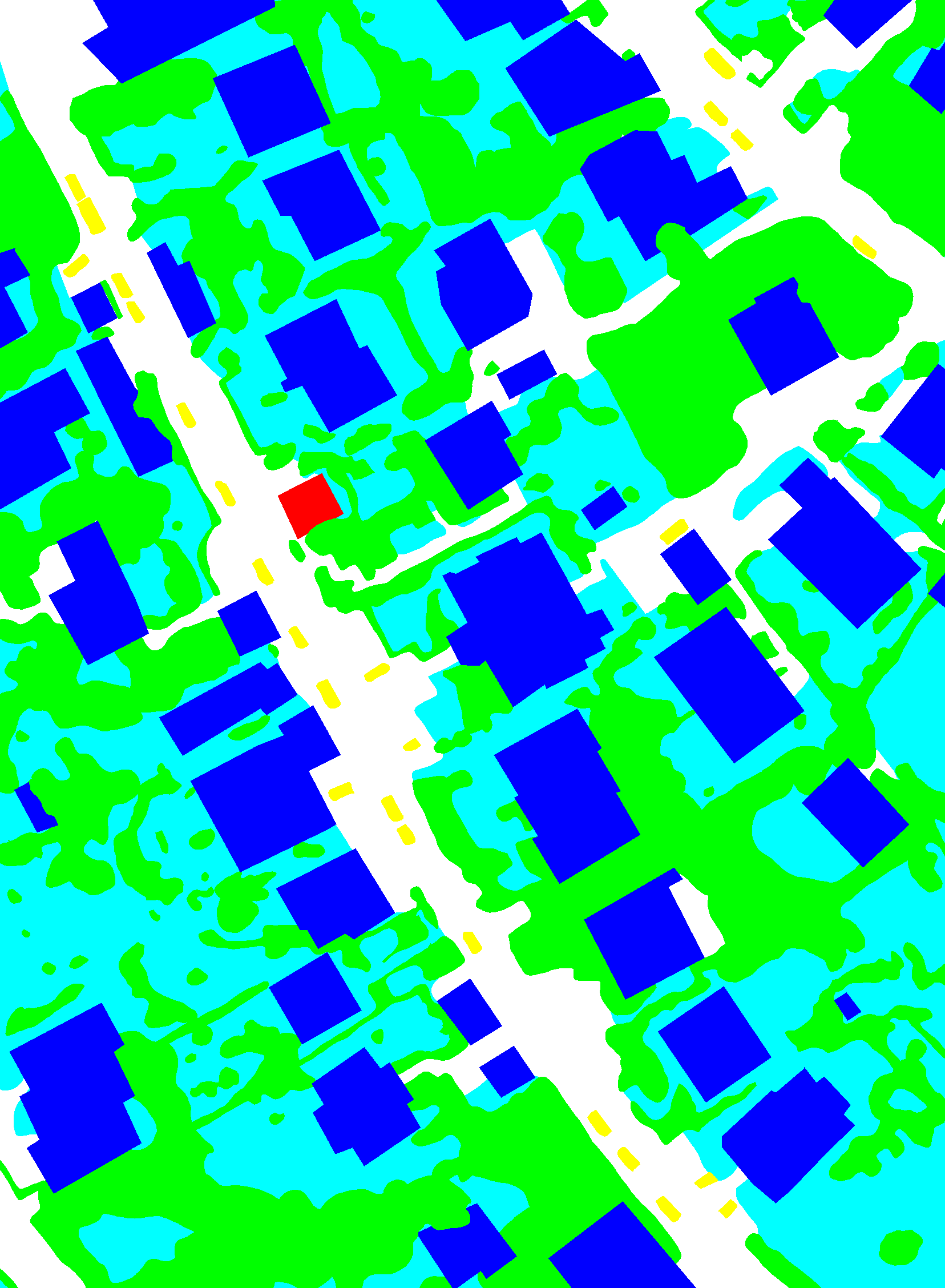}}
			\vspace{3pt}
			\centerline{\includegraphics[width=\textwidth, angle=90]{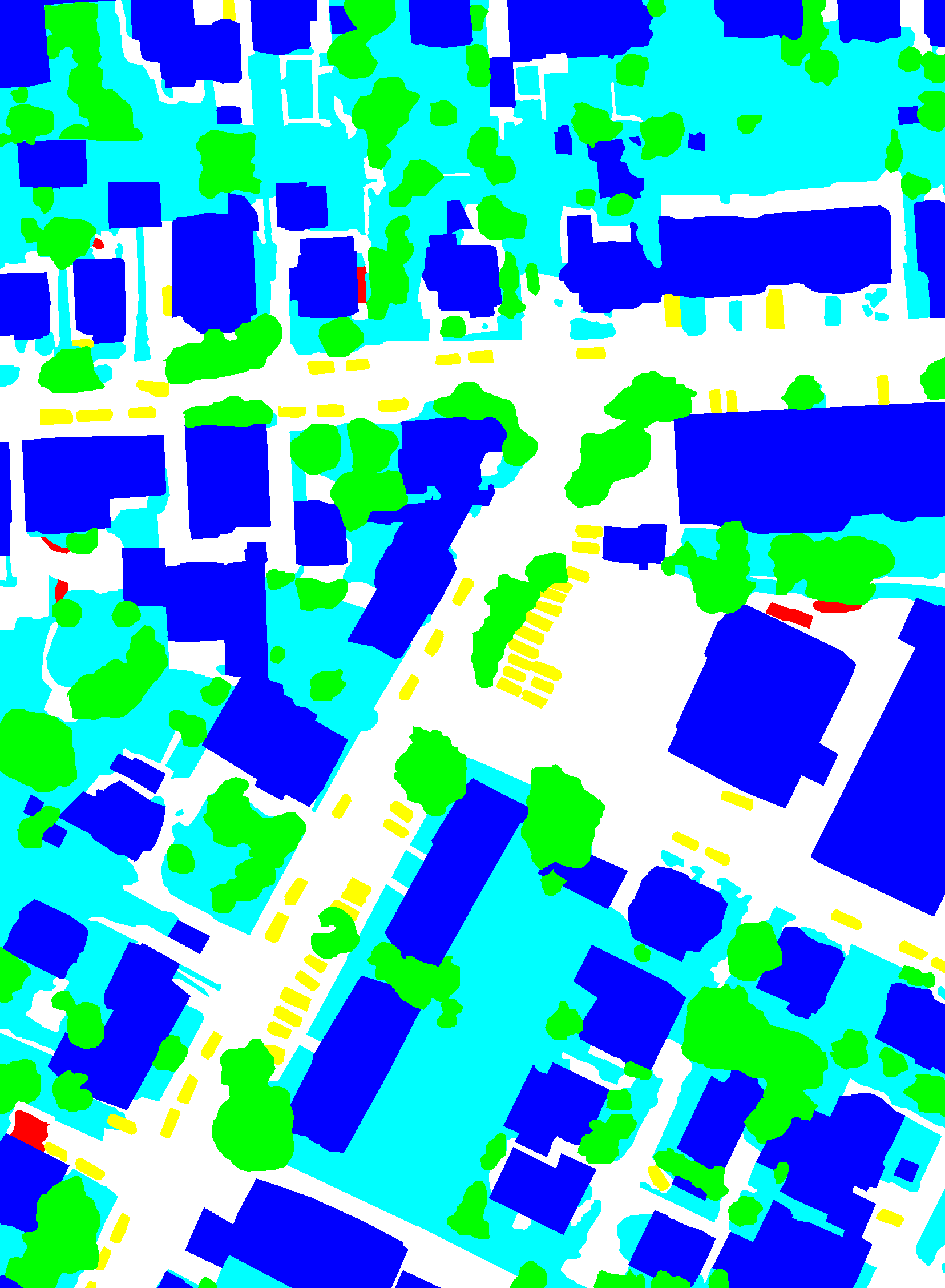}}
			\vspace{3pt}
			\centerline{\includegraphics[width=\textwidth, angle=90]{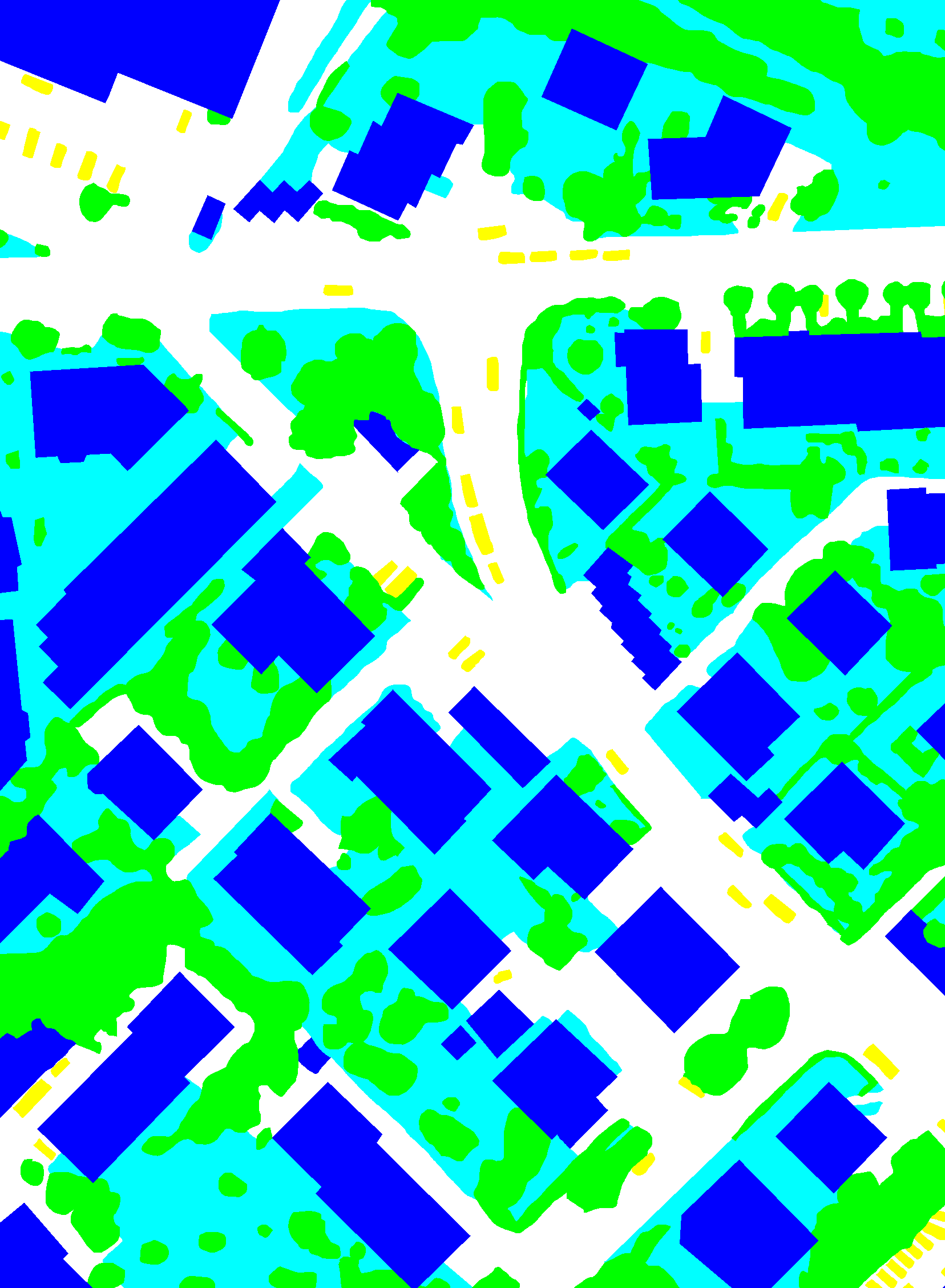}}
			\vspace{3pt}
			\centerline{\includegraphics[scale=0.031, angle=90]{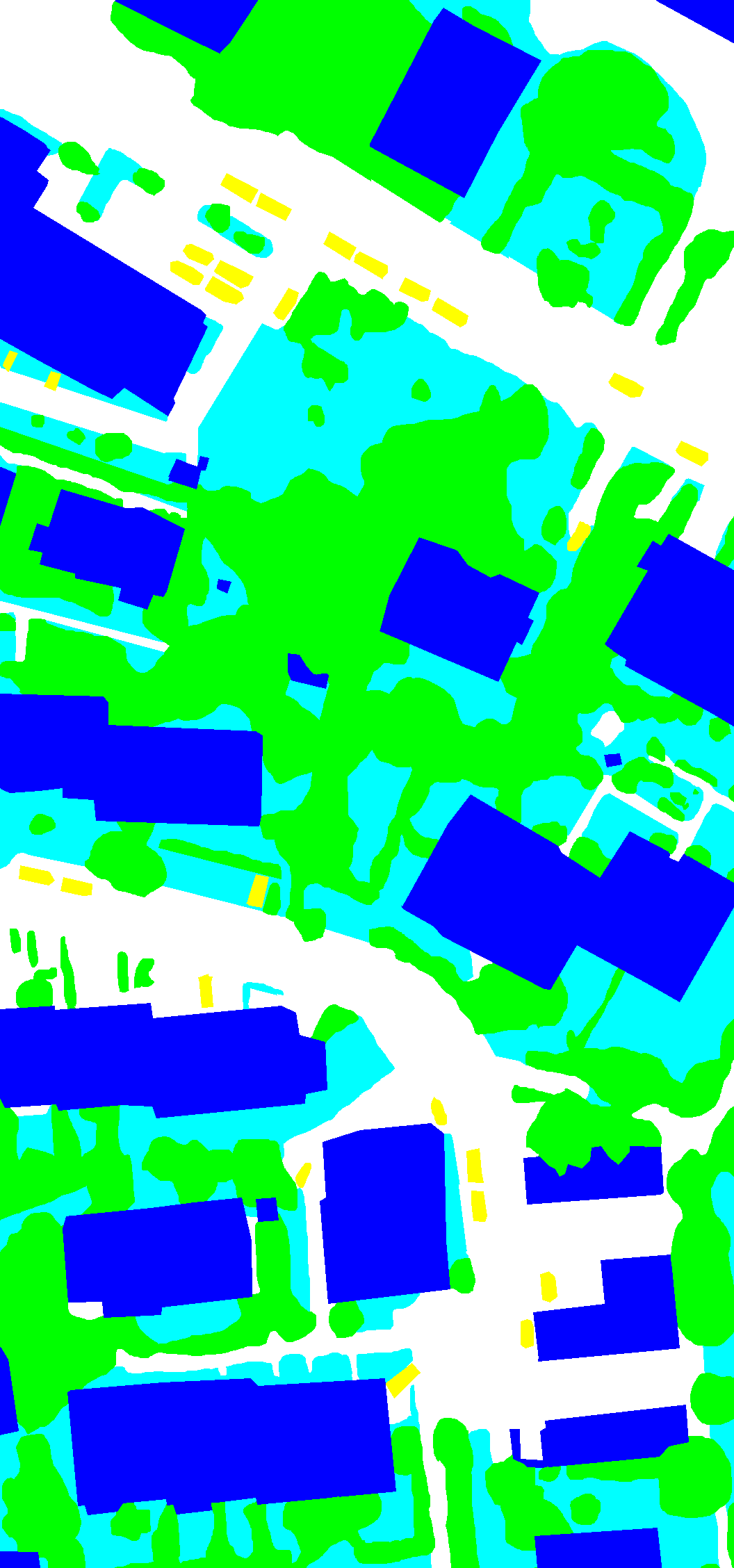}}
			\vspace{3pt}
			\centerline{(b)}
		\end{minipage}
		\hspace{15pt}
		\begin{minipage}{0.10\linewidth}
			\vspace{3pt}
			\centerline{\includegraphics[width=\textwidth, angle=90]{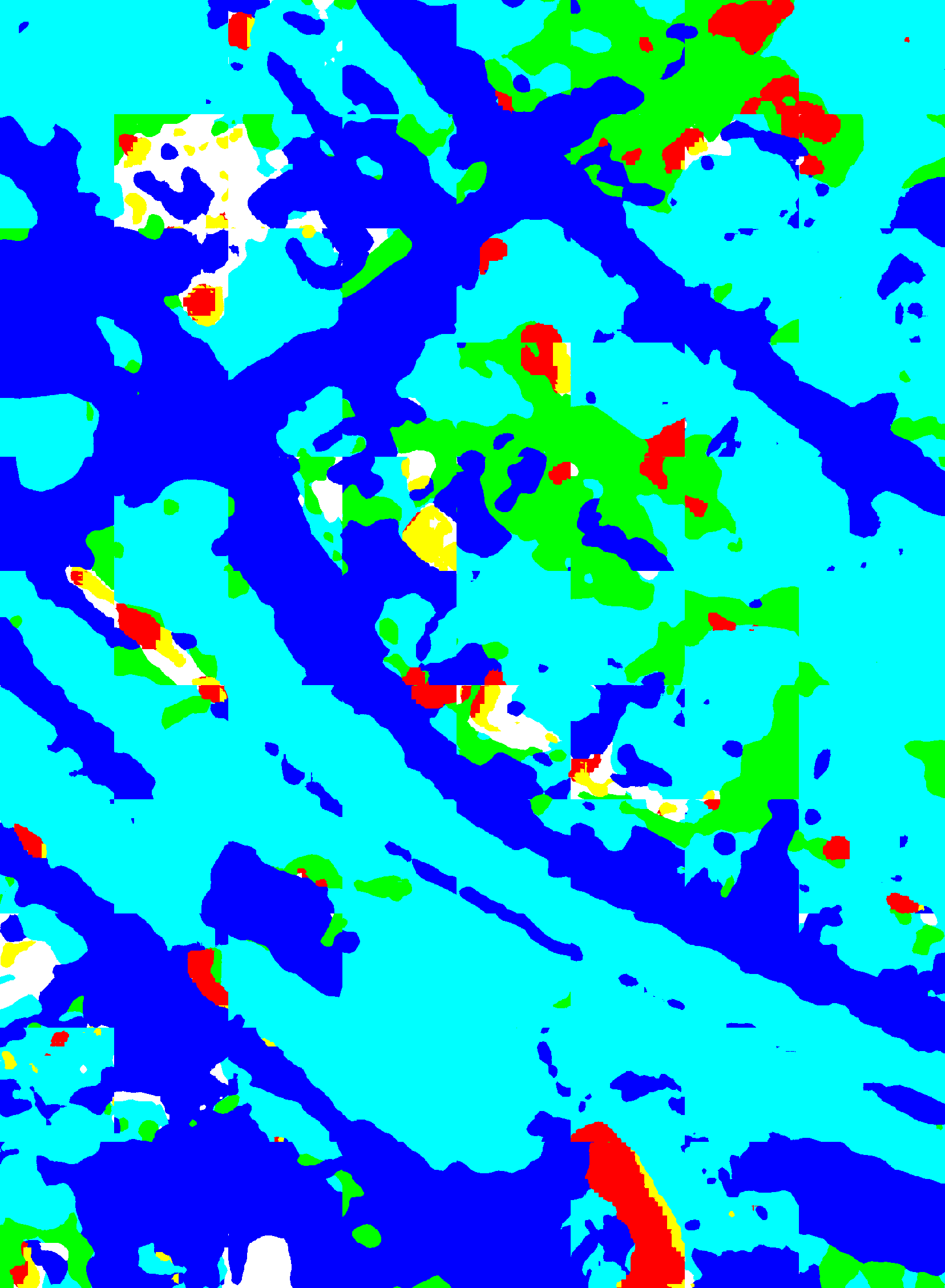}}
			\vspace{3pt}
			\centerline{\includegraphics[width=\textwidth, angle=90]{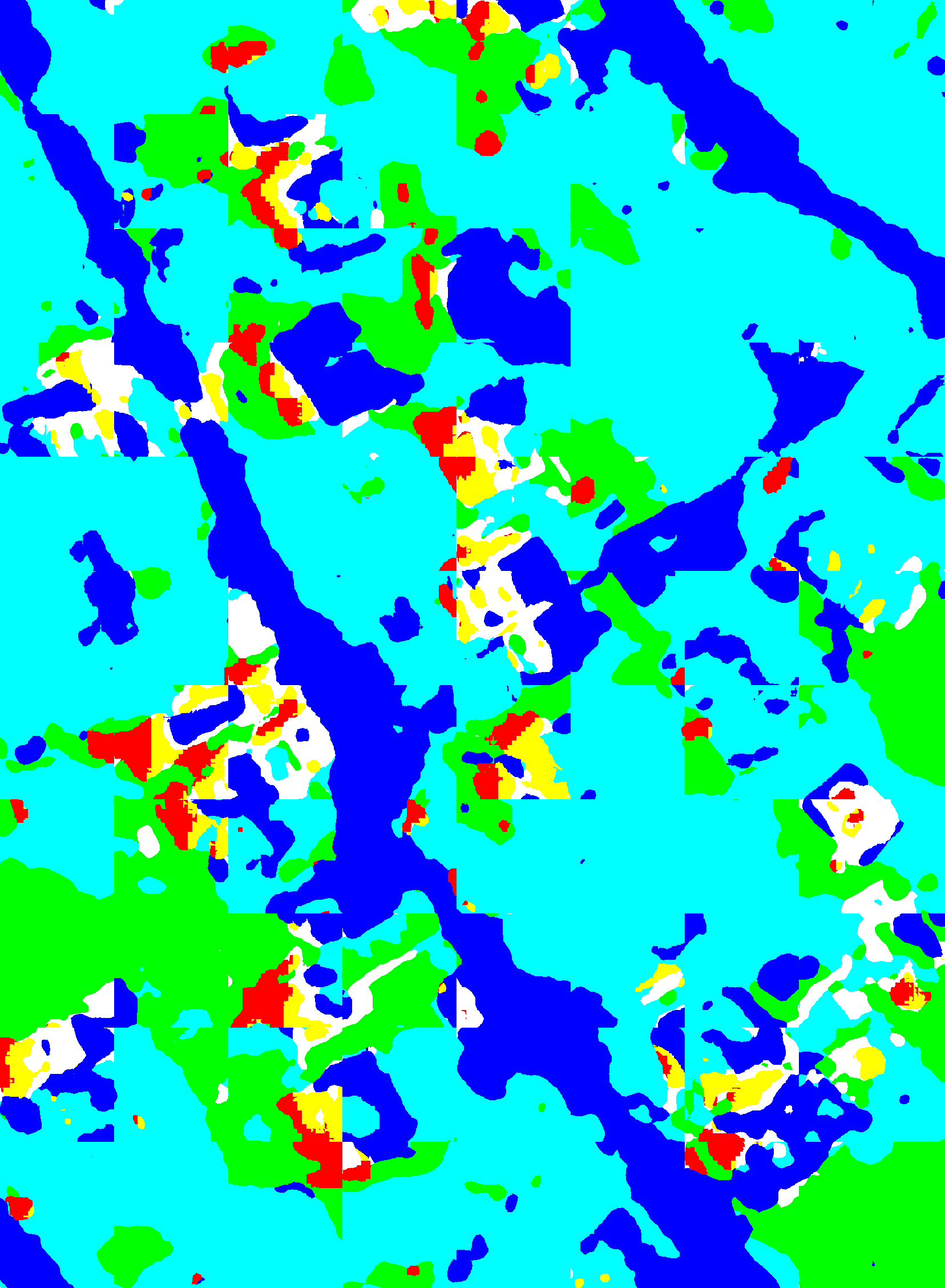}}
			\vspace{3pt}
			\centerline{\includegraphics[width=\textwidth, angle=90]{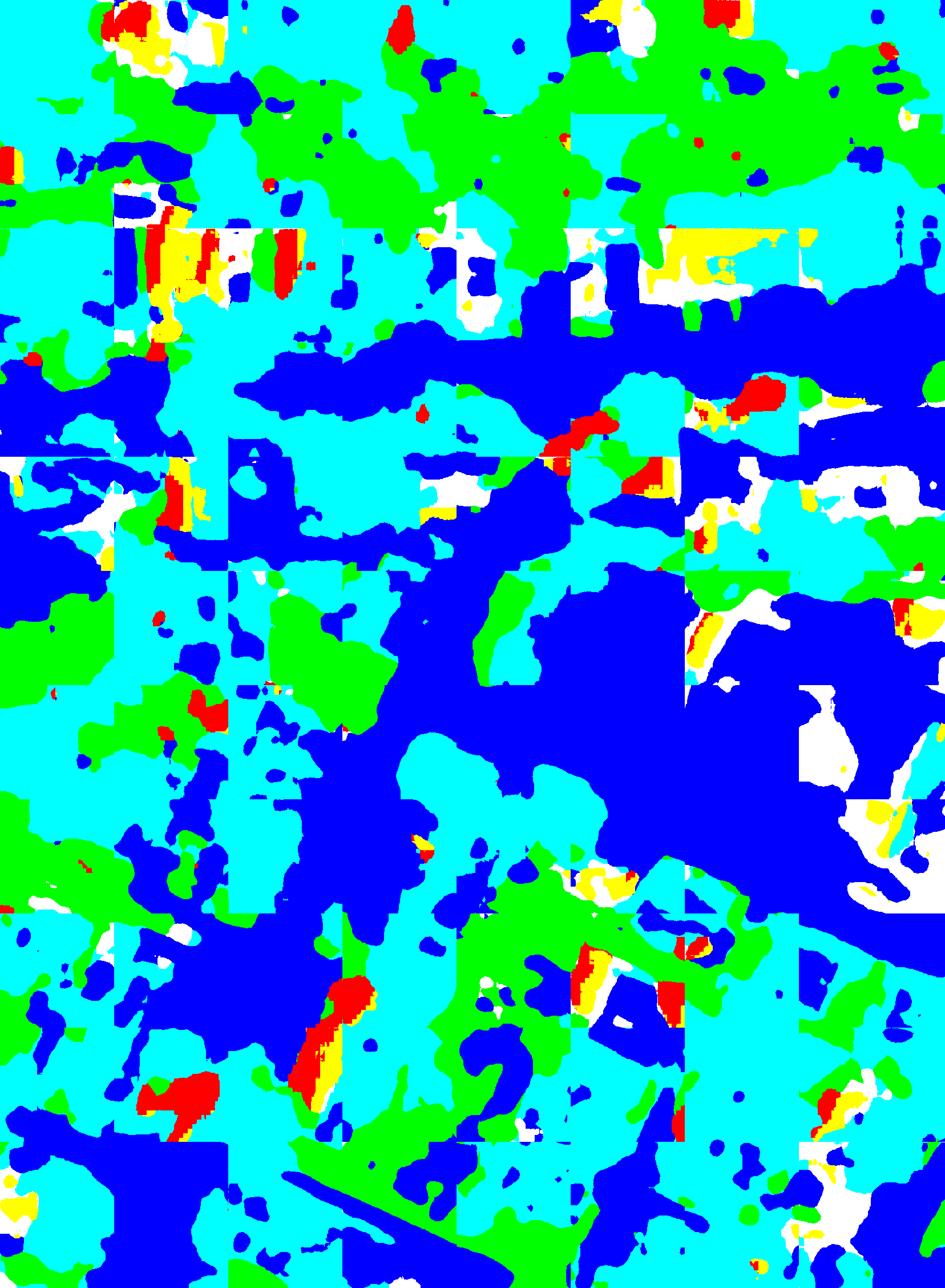}}
			\vspace{3pt}
			\centerline{\includegraphics[width=\textwidth, angle=90]{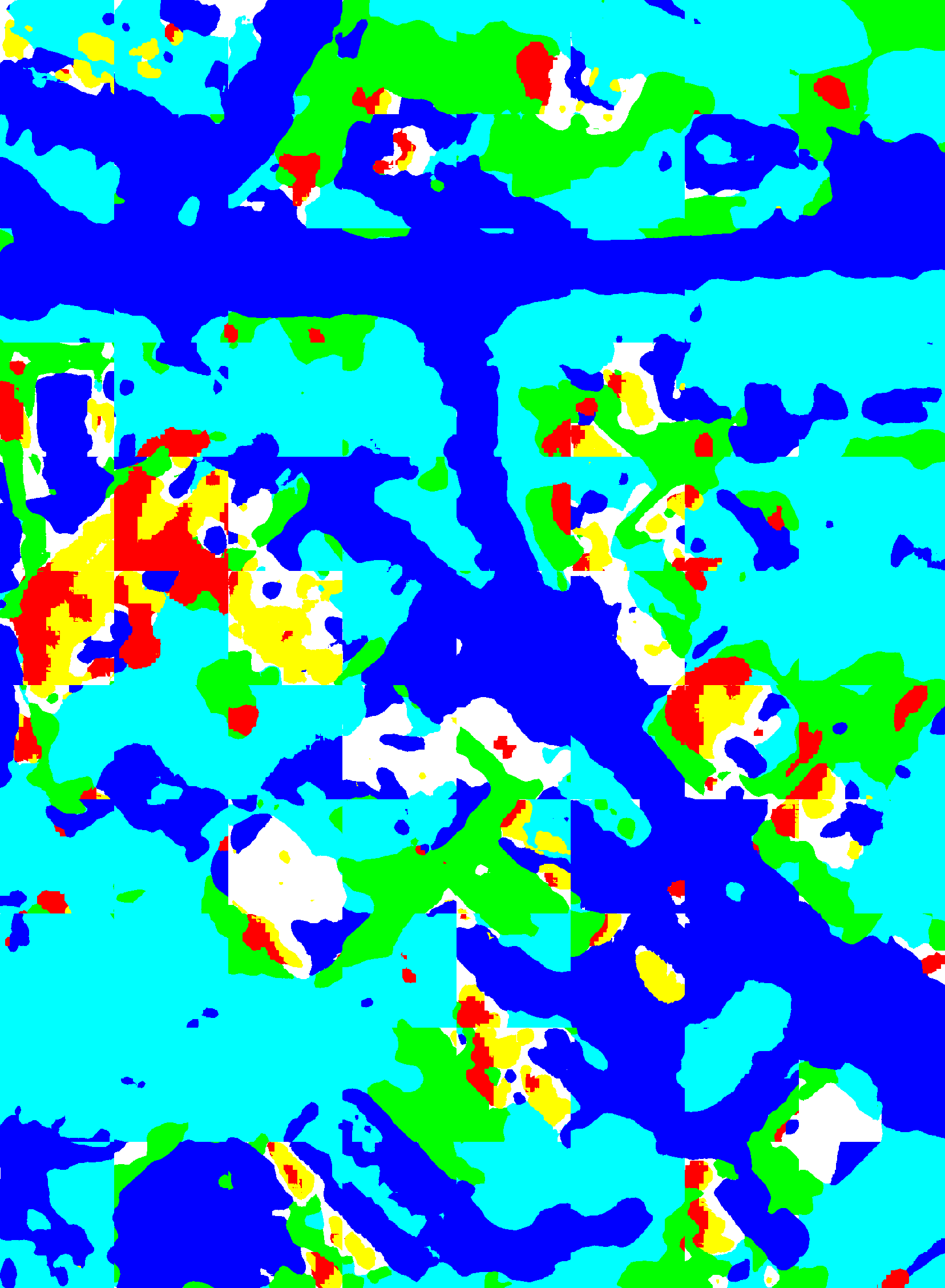}}
			\vspace{3pt}
			\centerline{\includegraphics[scale=0.031, angle=90]{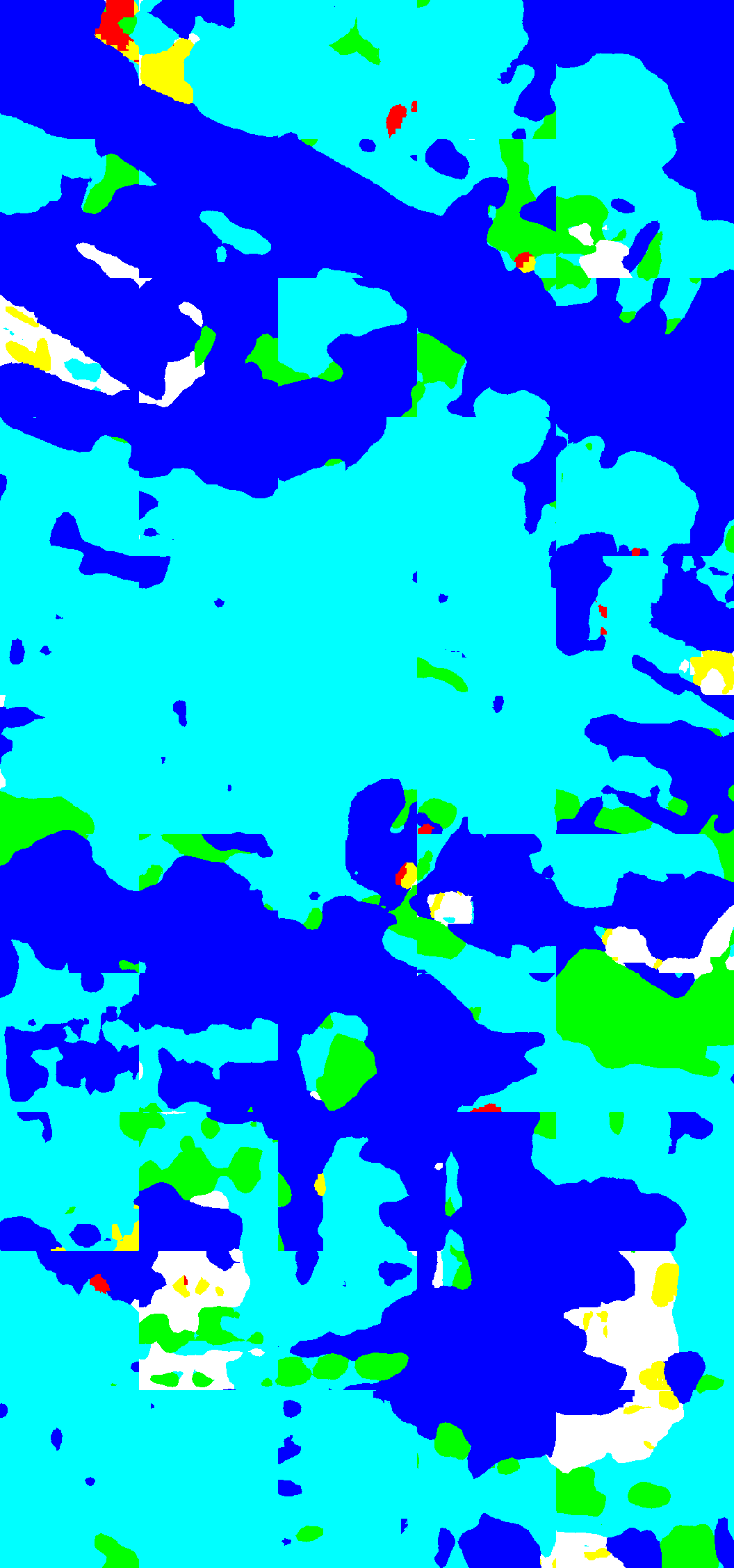}}
			\vspace{3pt}
			\centerline{(c)}
		\end{minipage}
		\hspace{15pt}
		\begin{minipage}{0.10\linewidth}
			\vspace{3pt}
			\centerline{\includegraphics[width=\textwidth, angle=90]{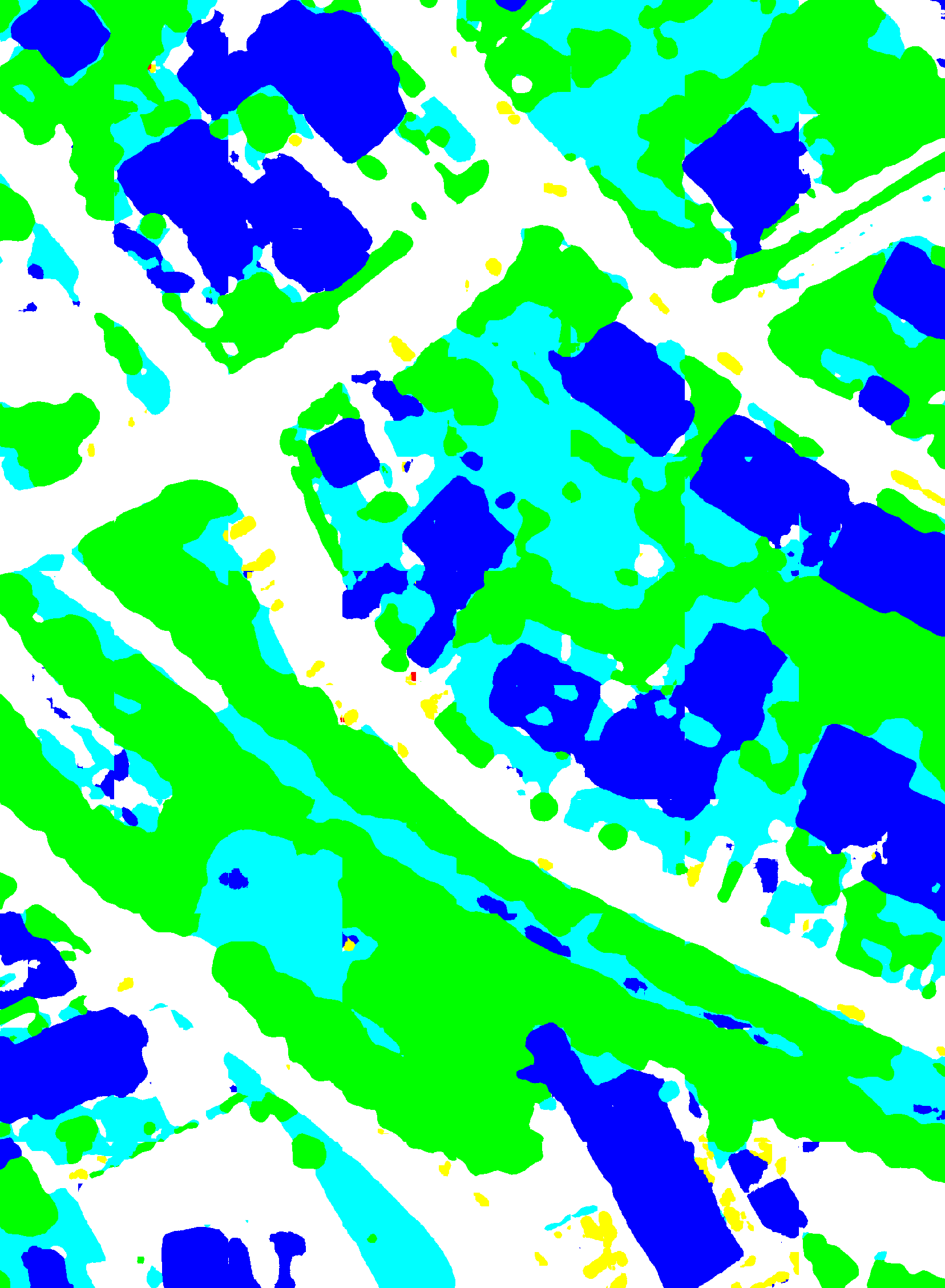}}
			\vspace{3pt}
			\centerline{\includegraphics[width=\textwidth, angle=90]{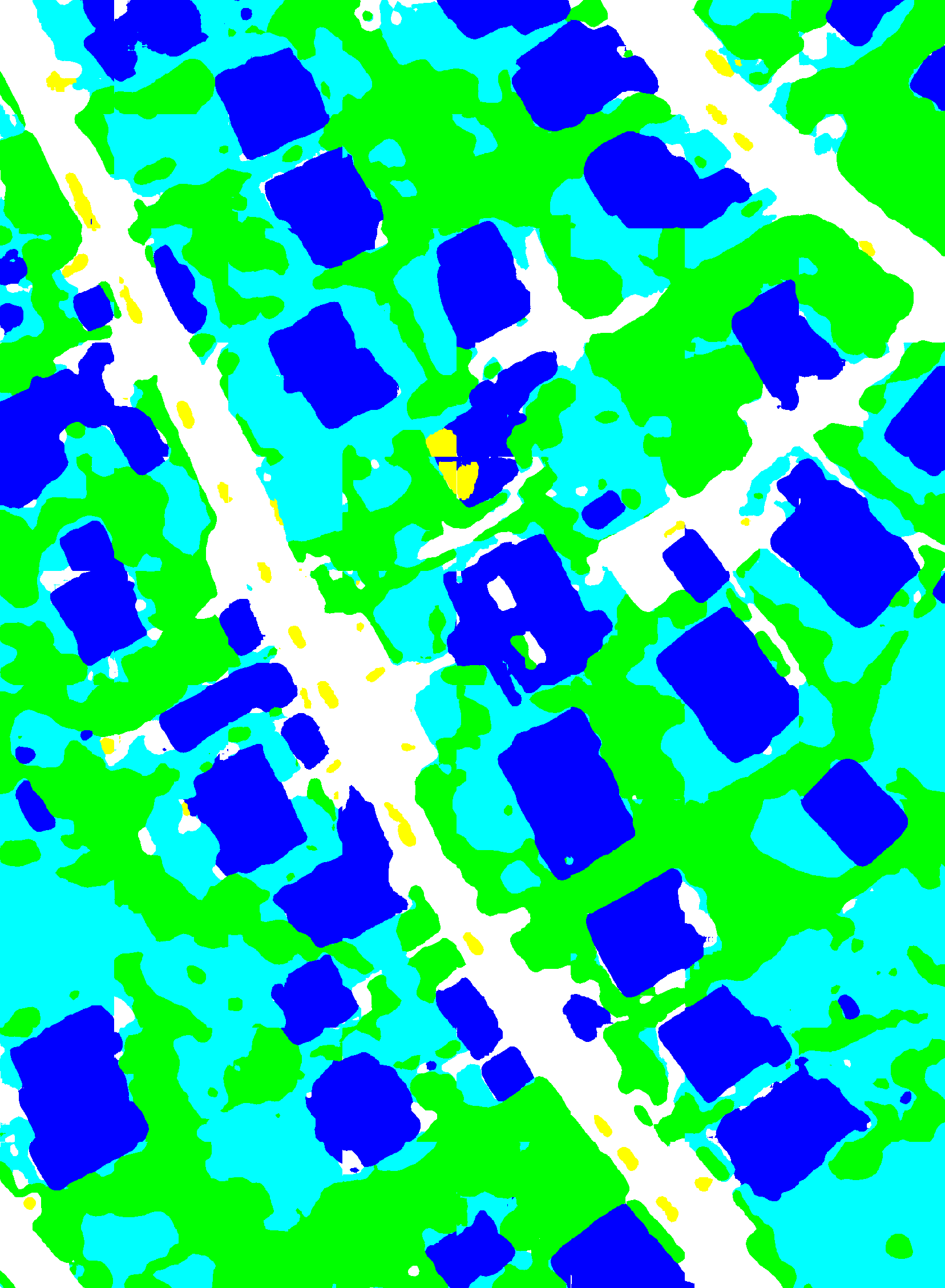}}
			\vspace{3pt}
			\centerline{\includegraphics[width=\textwidth, angle=90]{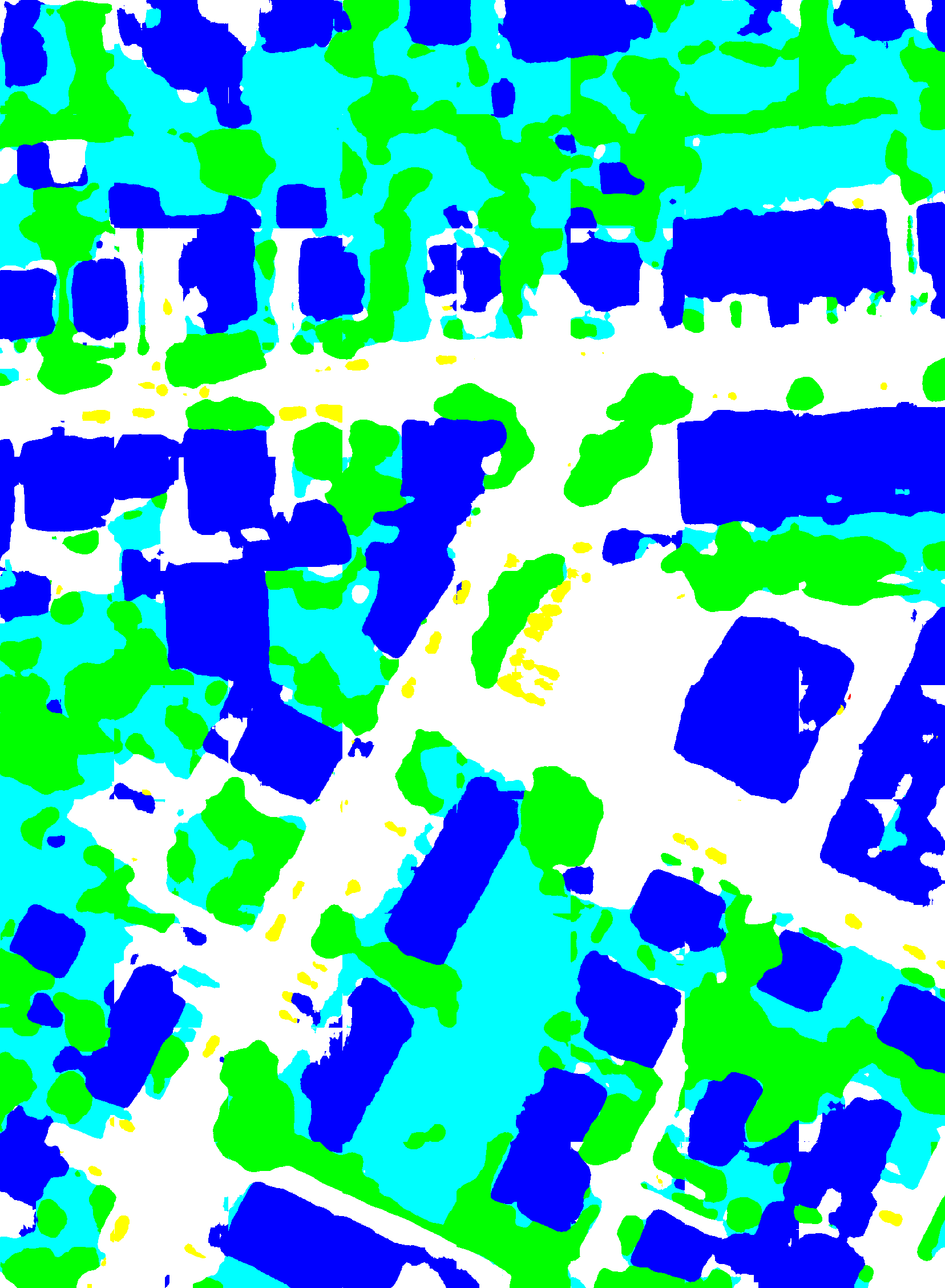}}
			\vspace{3pt}
			\centerline{\includegraphics[width=\textwidth, angle=90]{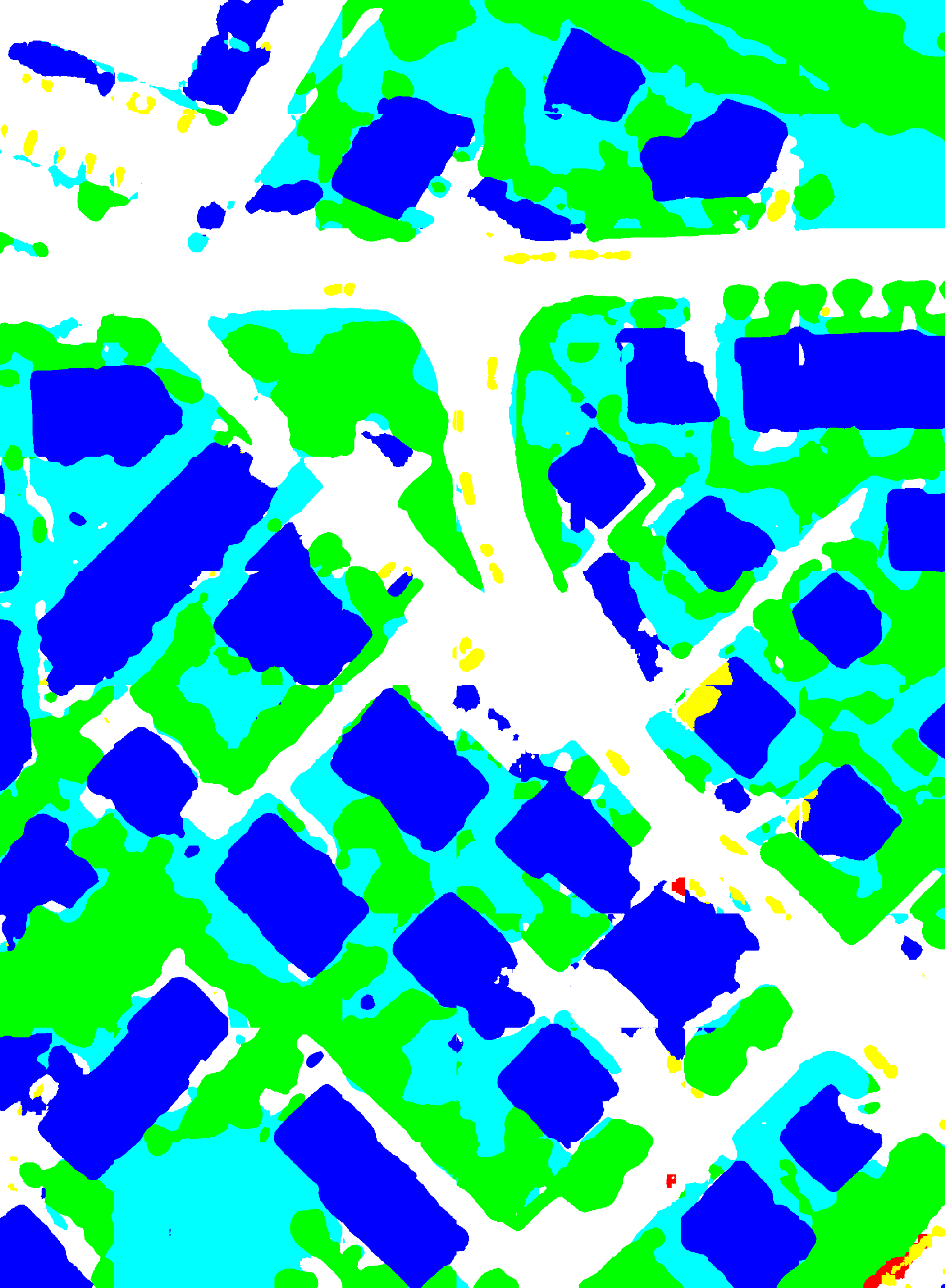}}
			\vspace{3pt}
			\centerline{\includegraphics[scale=0.031, angle=90]{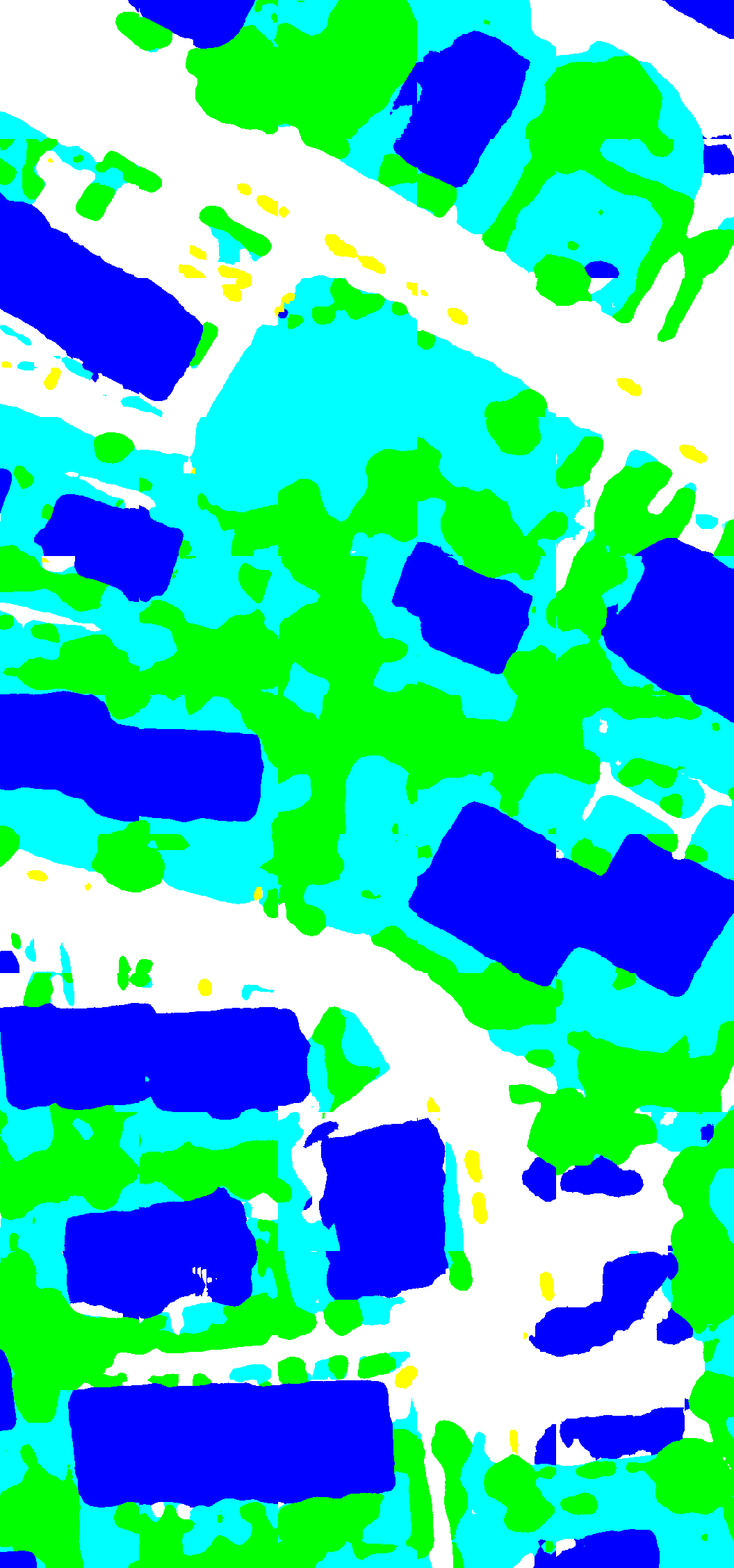}}
			\vspace{3pt}
			\centerline{(d)}
		\end{minipage}
		\hspace{15pt}
		\begin{minipage}{0.10\linewidth}
			\vspace{3pt}
			\centerline{\includegraphics[width=\textwidth, angle=90]{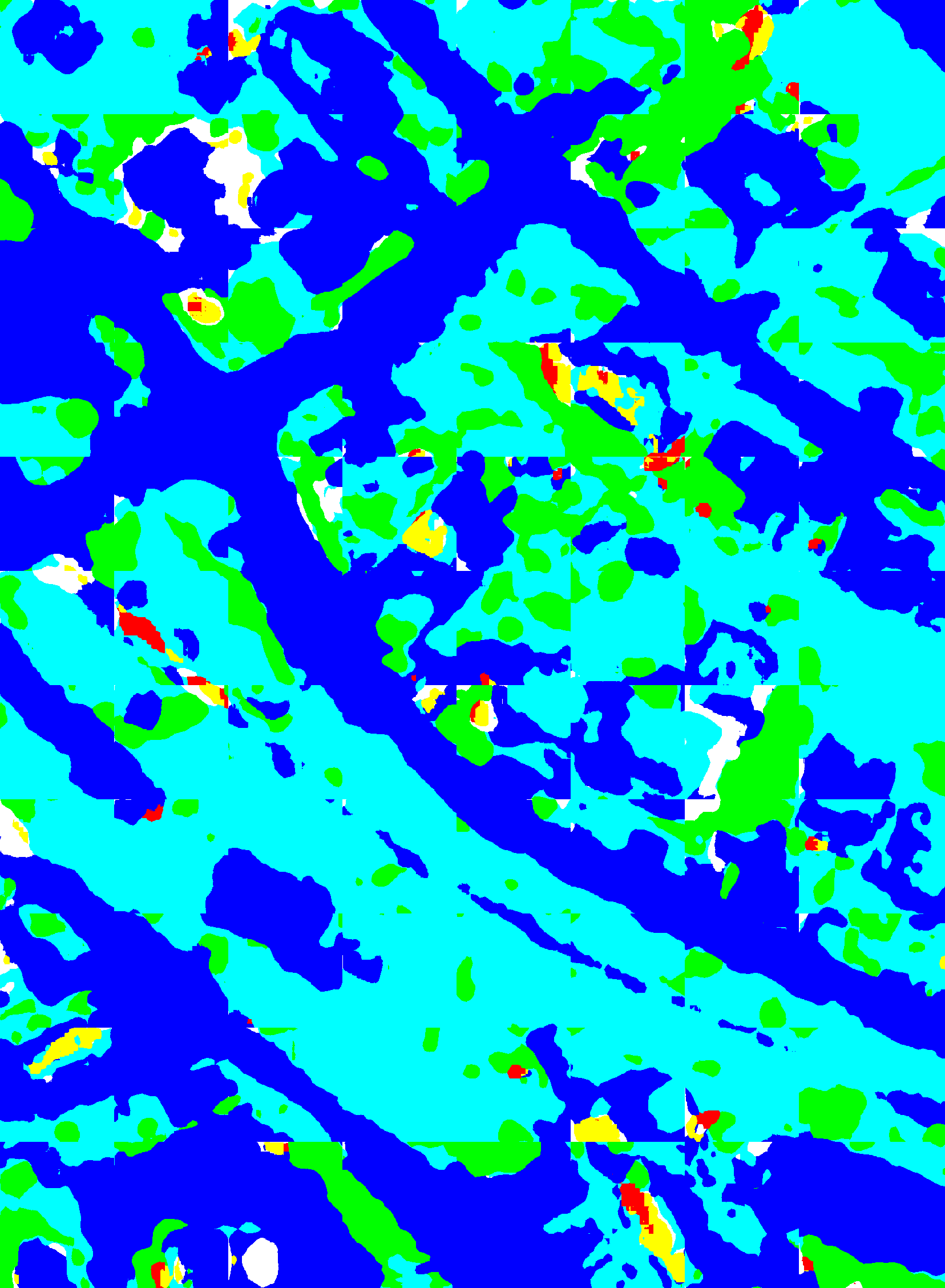}}
			\vspace{3pt}
			\centerline{\includegraphics[width=\textwidth, angle=90]{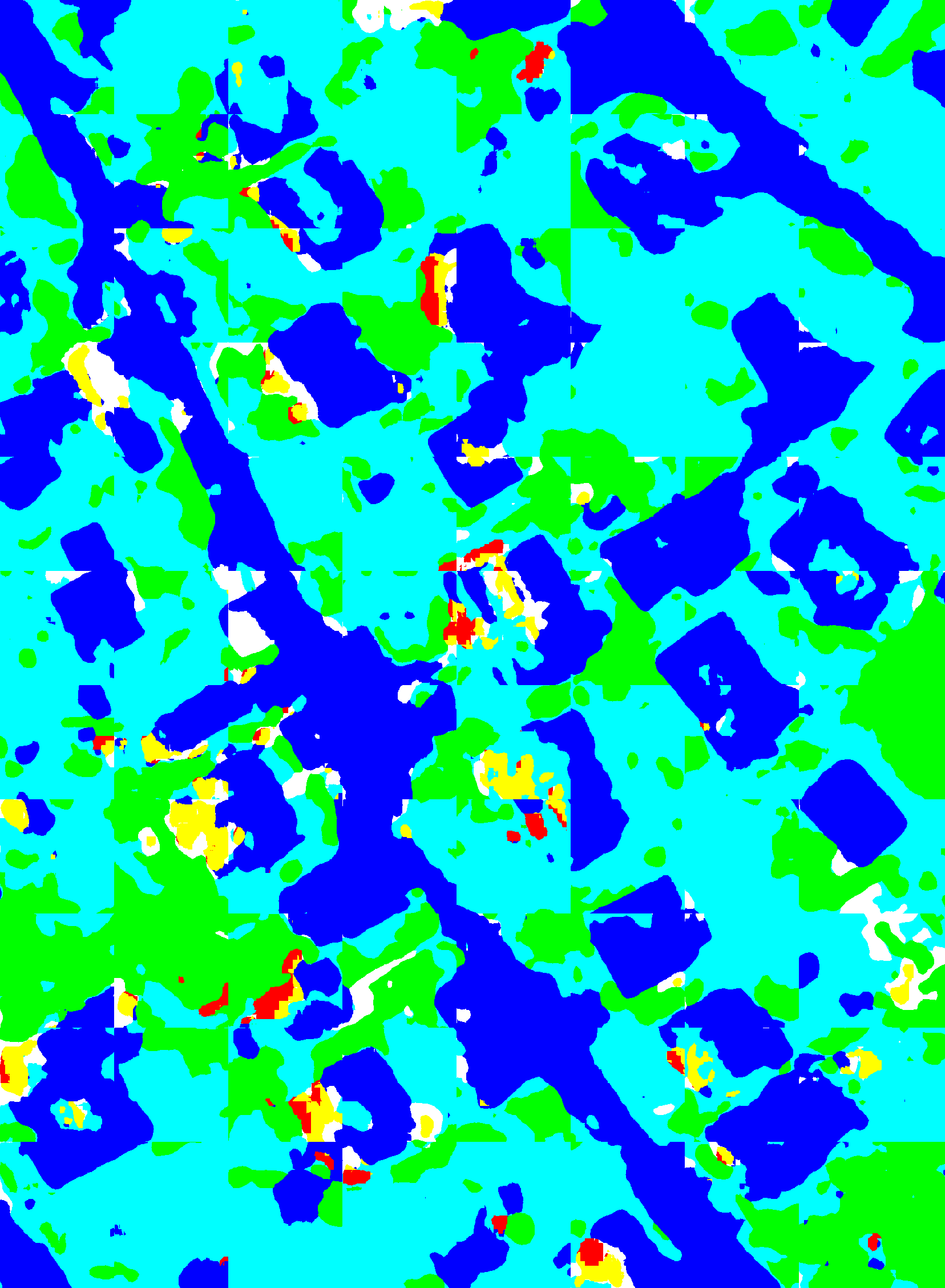}}
			\vspace{3pt}
			\centerline{\includegraphics[width=\textwidth, angle=90]{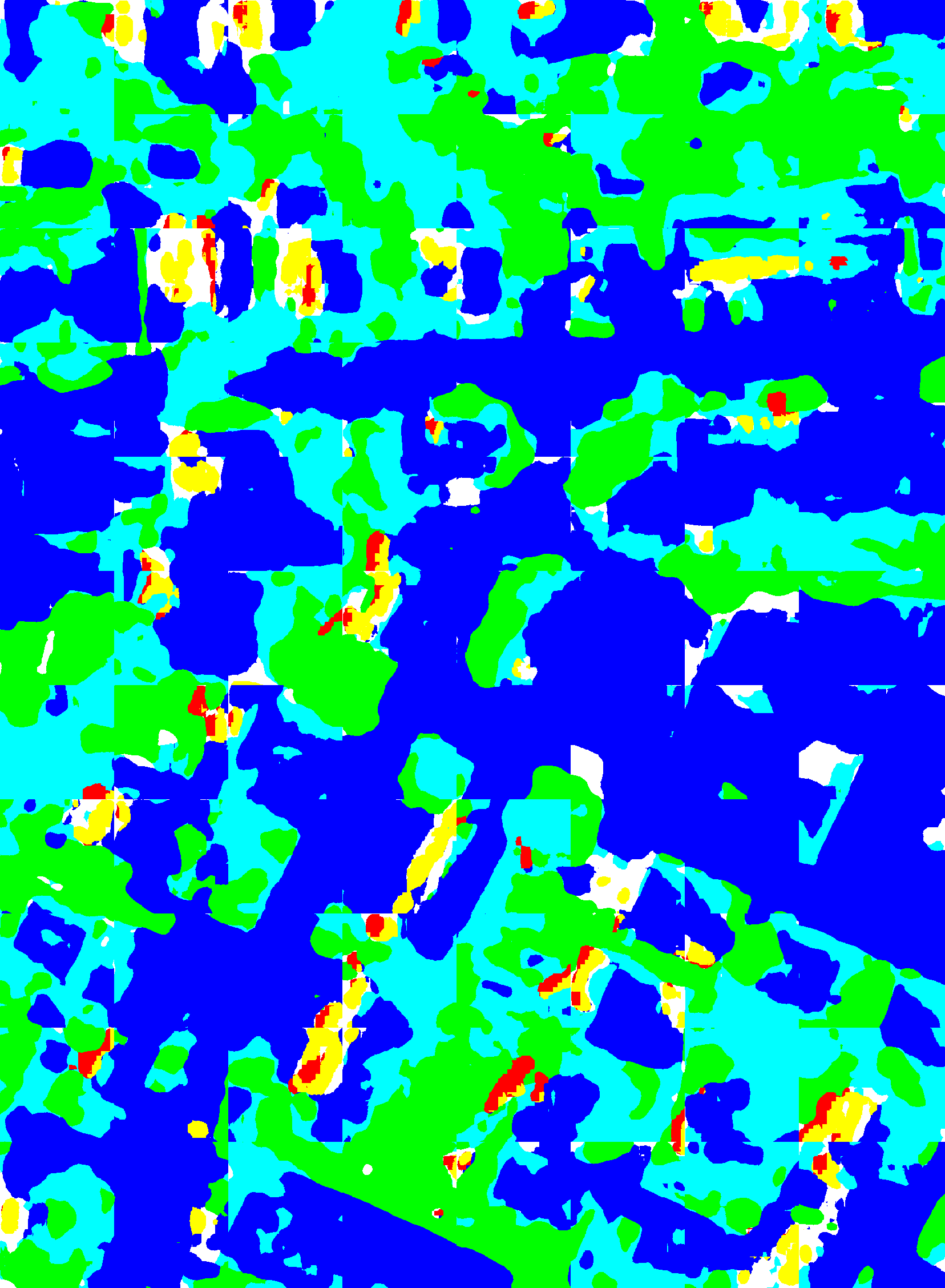}}
			\vspace{3pt}
			\centerline{\includegraphics[width=\textwidth, angle=90]{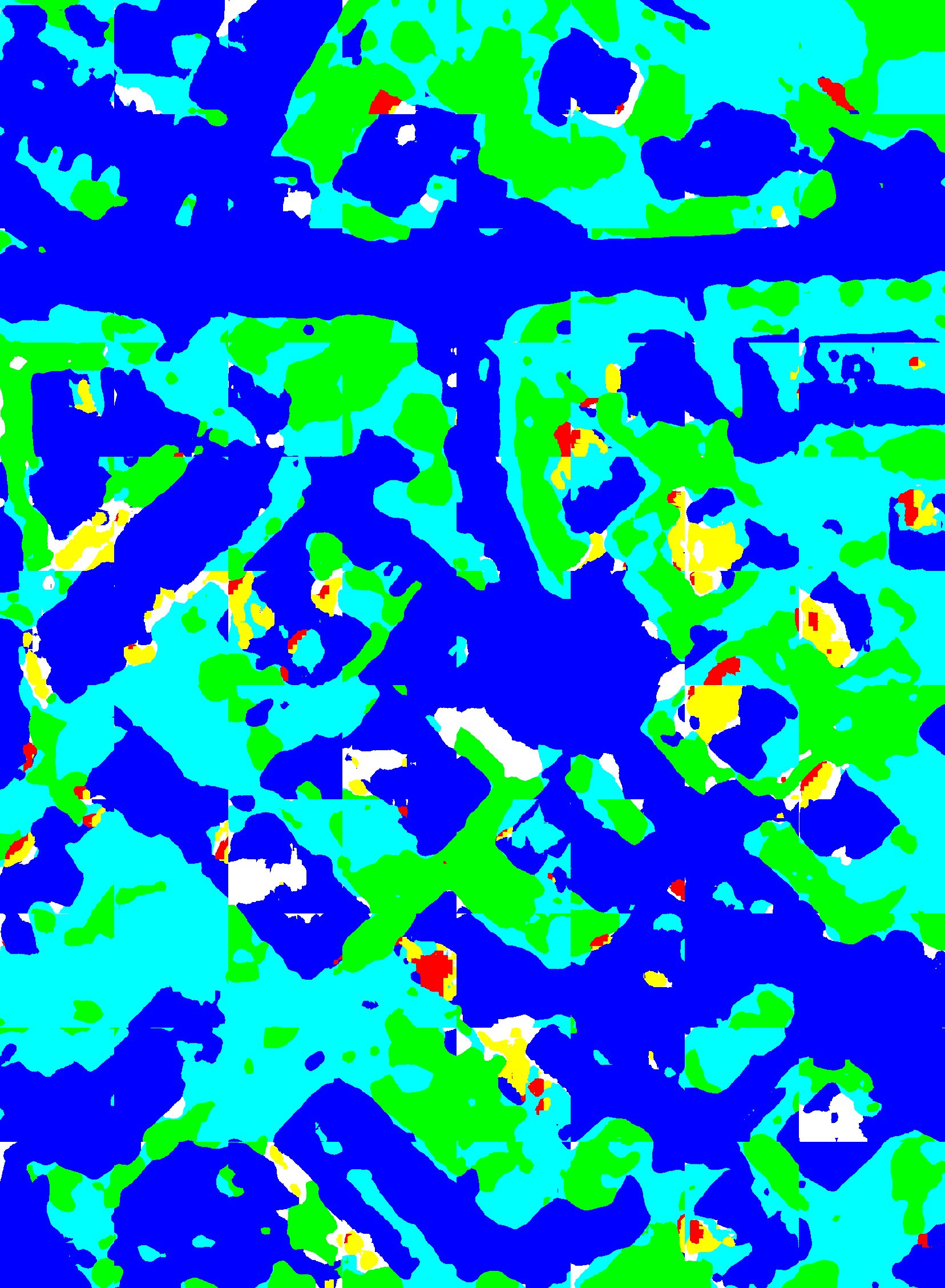}}
			\vspace{3pt}
			\centerline{\includegraphics[scale=0.031, angle=90]{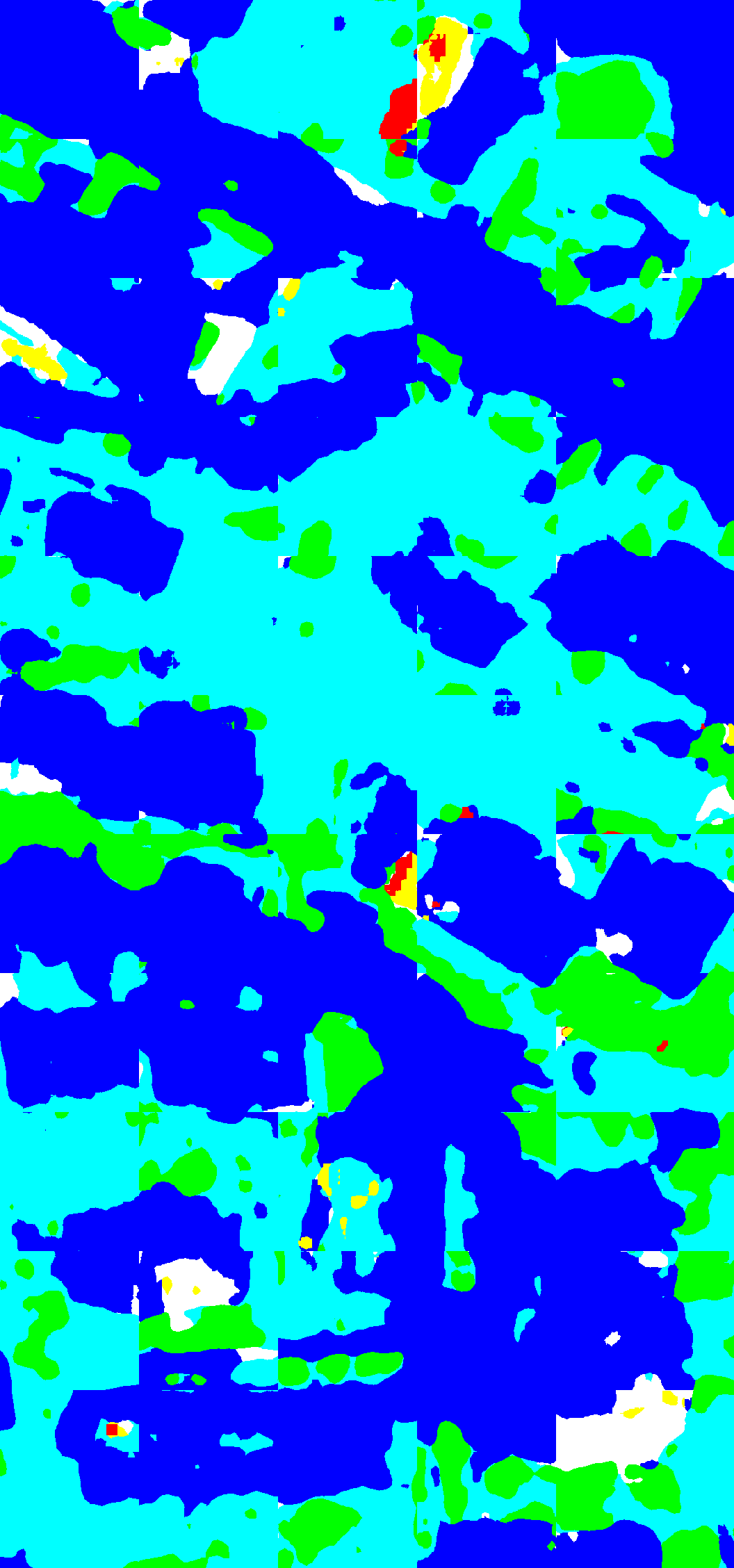}}
			\vspace{3pt}
			\centerline{(e)}
		\end{minipage}
		\hspace{15pt}
		\begin{minipage}{0.10\linewidth}
			\vspace{3pt}
			\centerline{\includegraphics[width=\textwidth, angle=90]{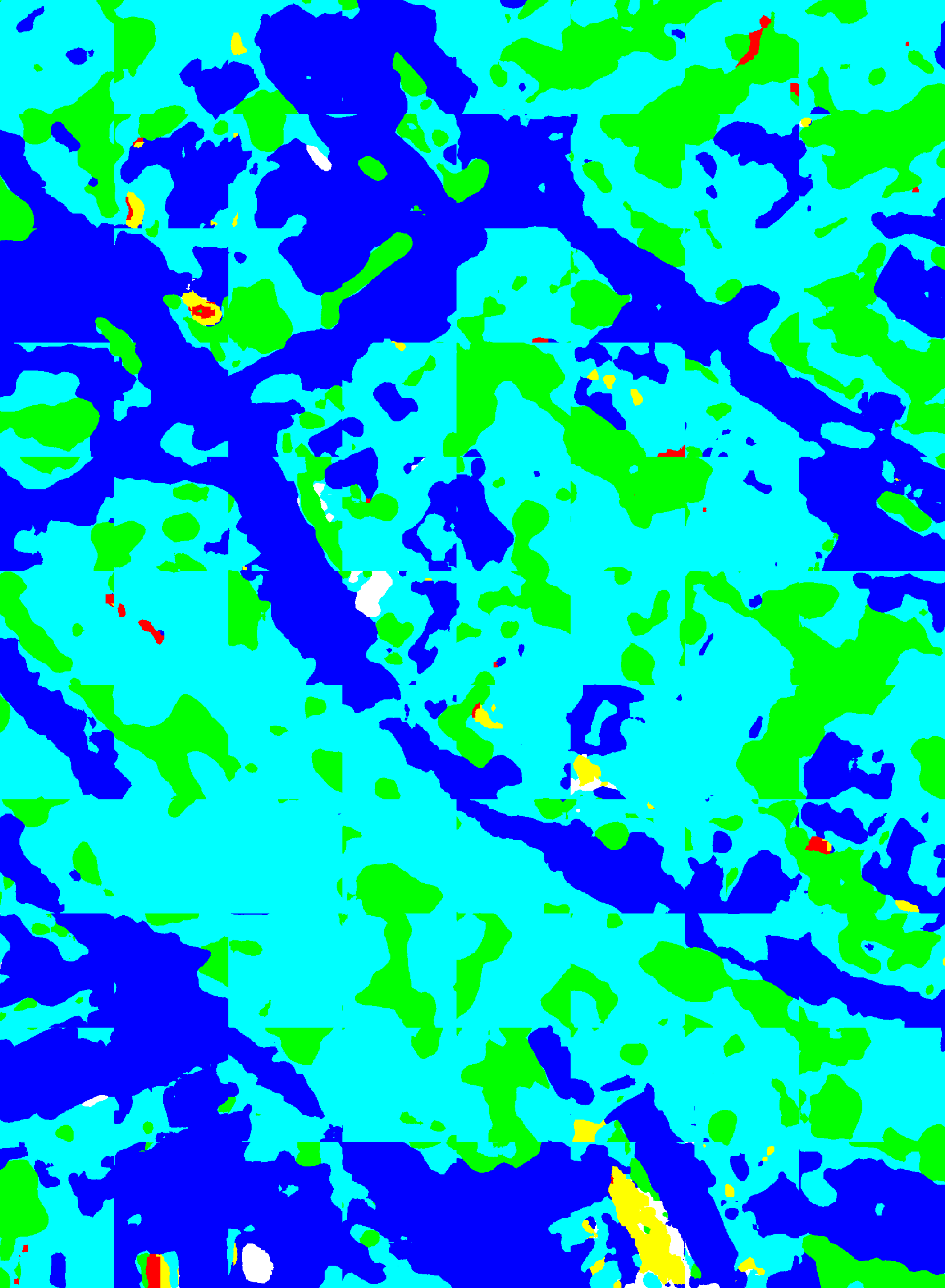}}
			\vspace{3pt}
			\centerline{\includegraphics[width=\textwidth, angle=90]{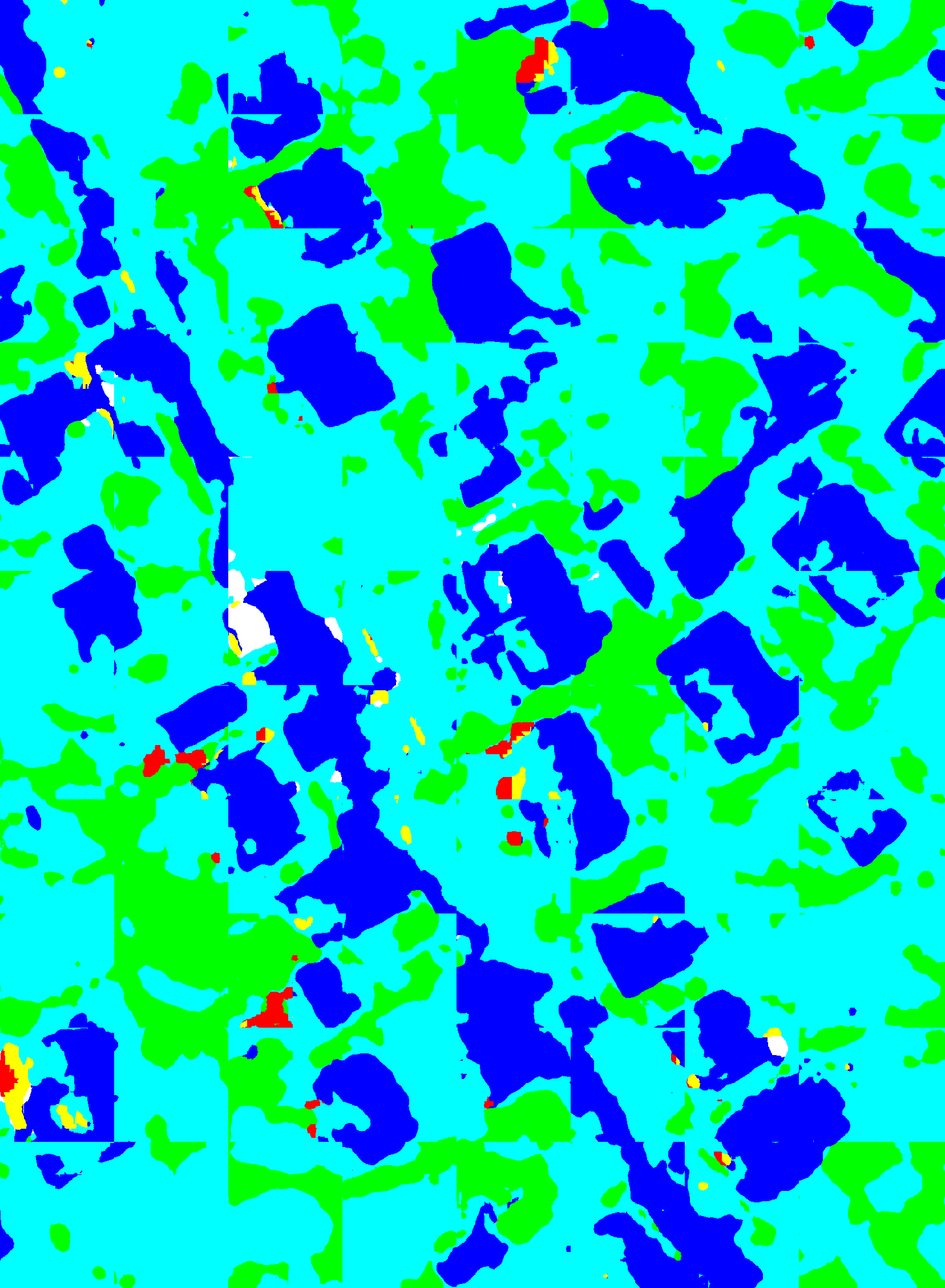}}
			\vspace{3pt}
			\centerline{\includegraphics[width=\textwidth, angle=90]{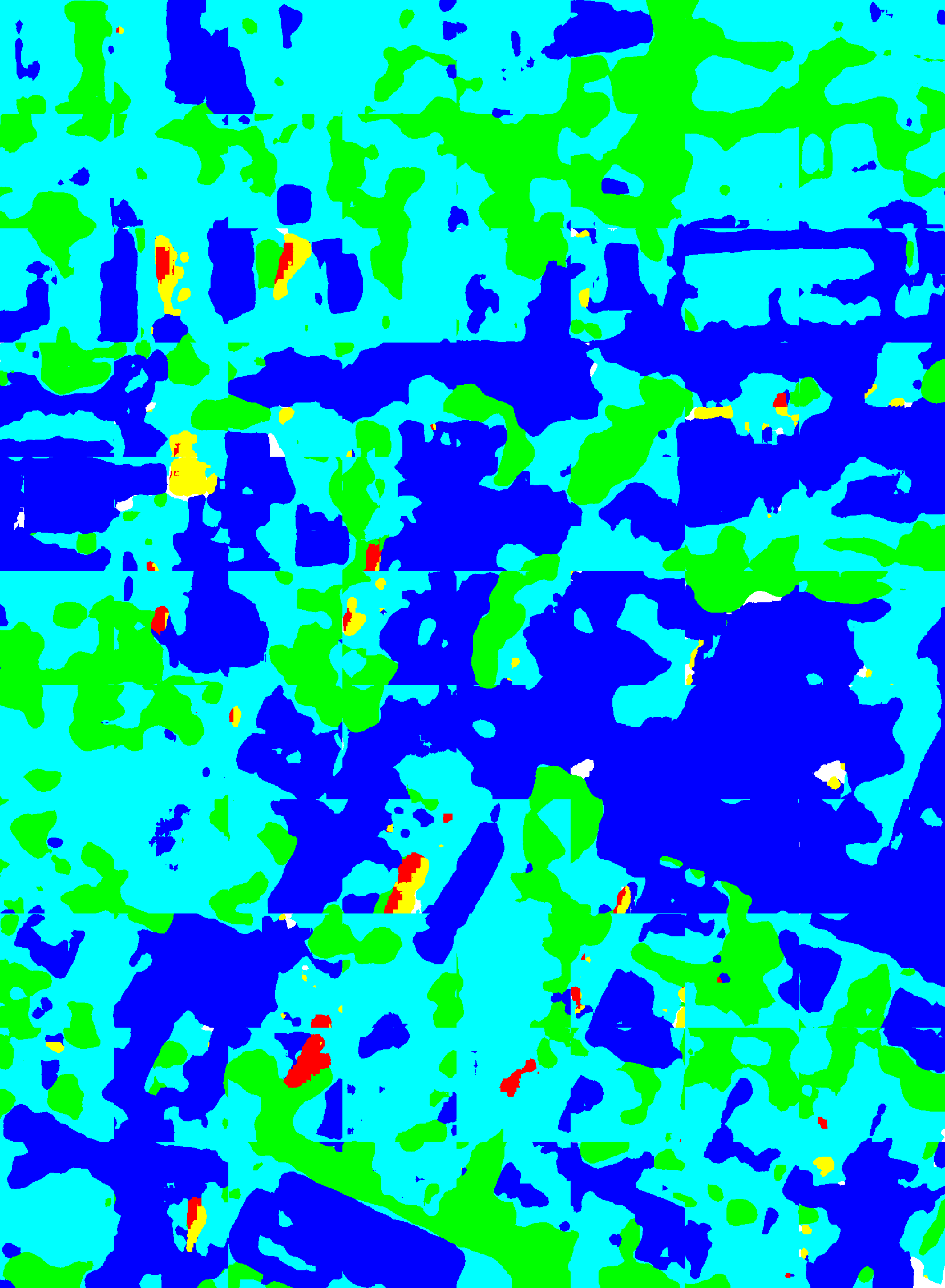}}
			\vspace{3pt}
			\centerline{\includegraphics[width=\textwidth, angle=90]{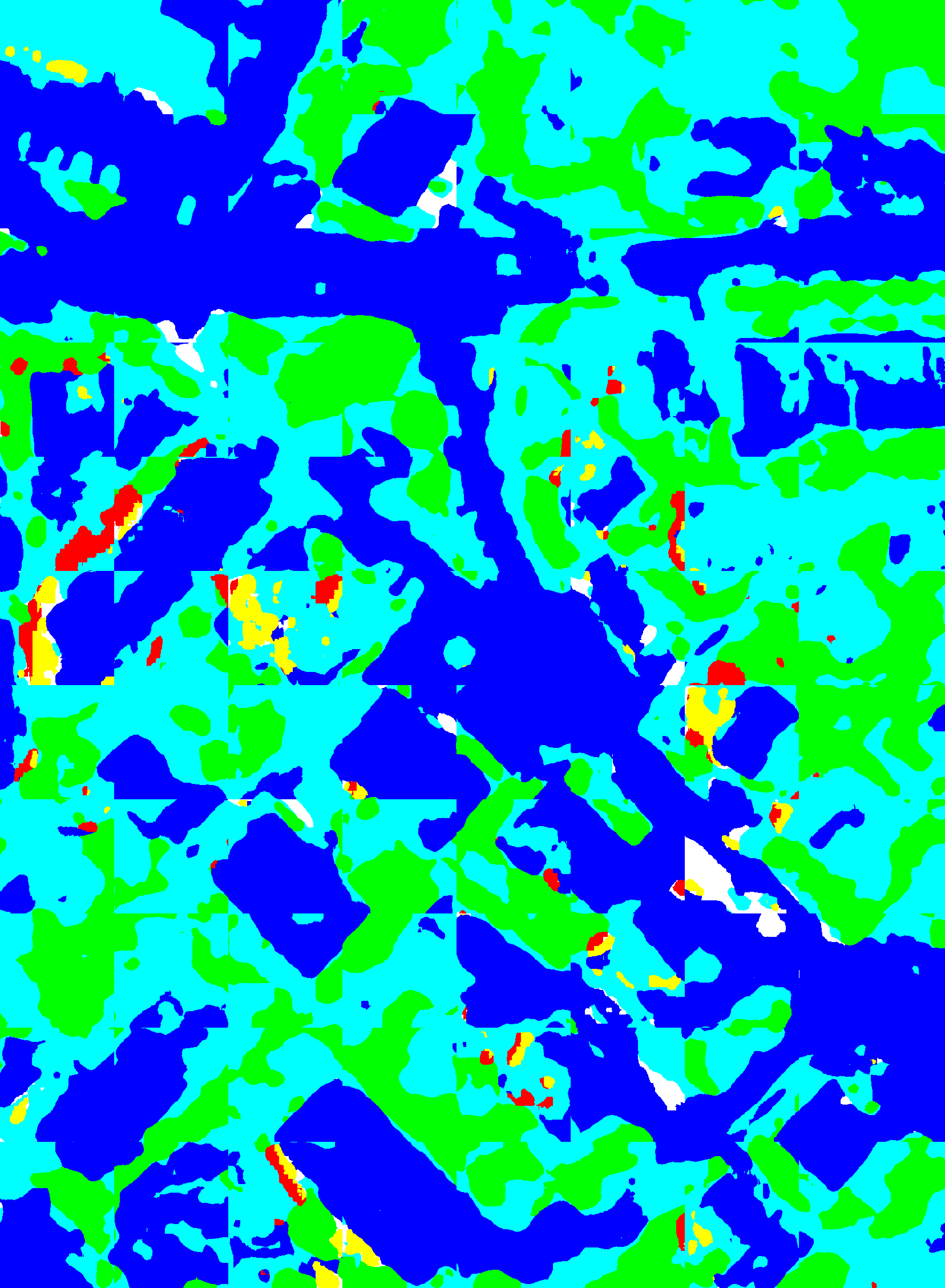}}
			\vspace{3pt}
			\centerline{\includegraphics[scale=0.031, angle=90]{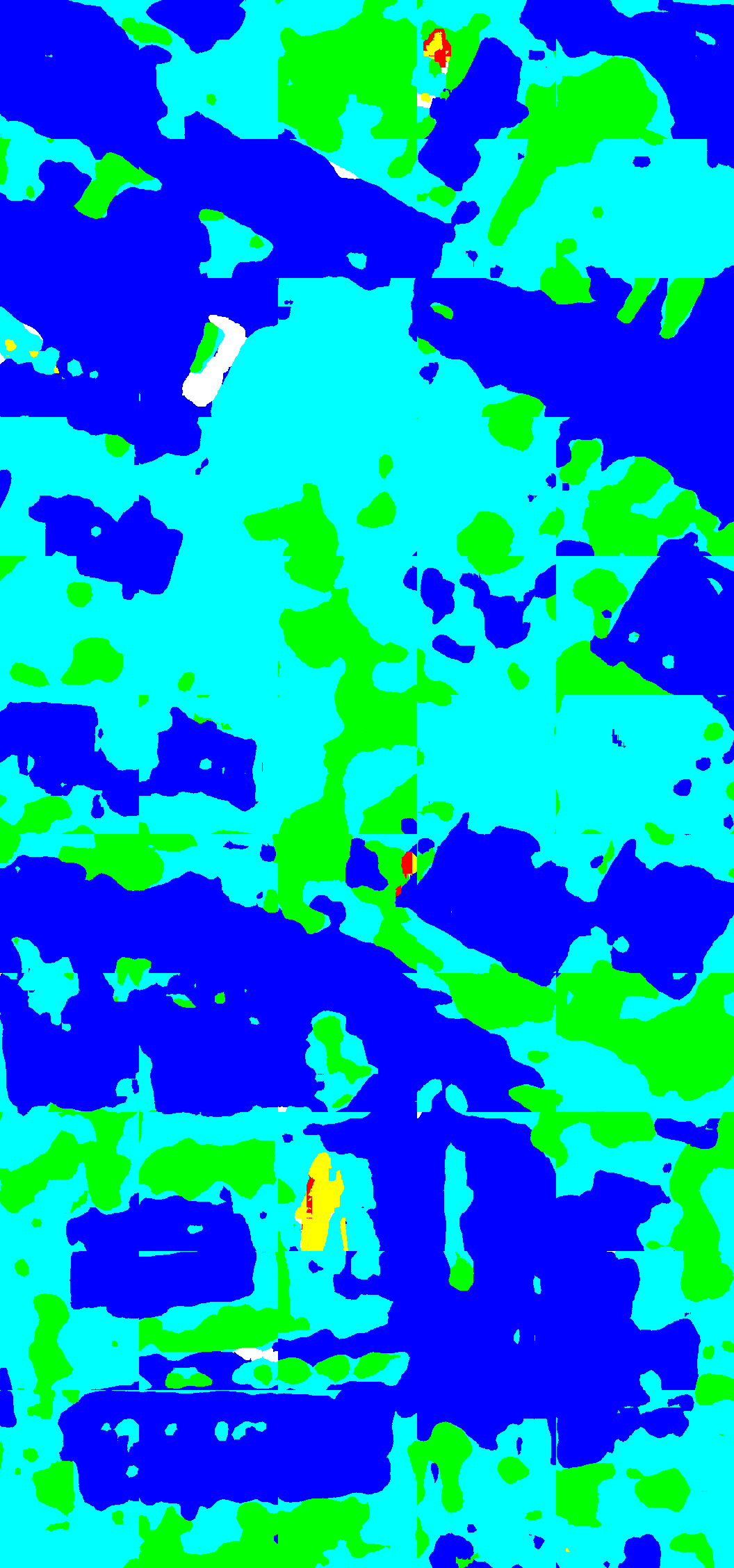}}
			\vspace{3pt}
			\centerline{(f)}
		\end{minipage}
		\hspace{15pt}
		\begin{minipage}{0.10\linewidth}
			\vspace{3pt}
			\centerline{\includegraphics[width=\textwidth, angle=90]{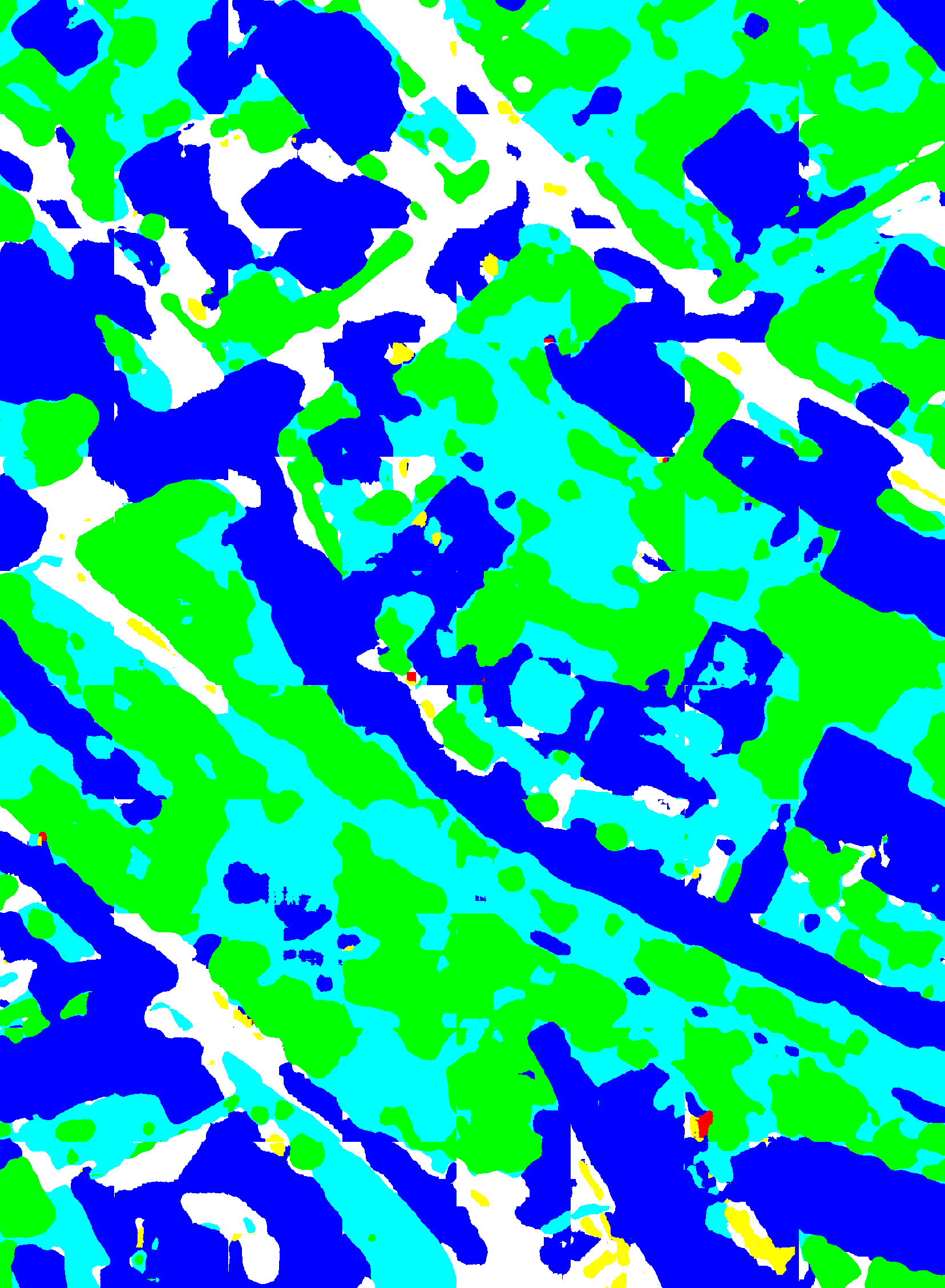}}
			\vspace{3pt}
			\centerline{\includegraphics[width=\textwidth, angle=90]{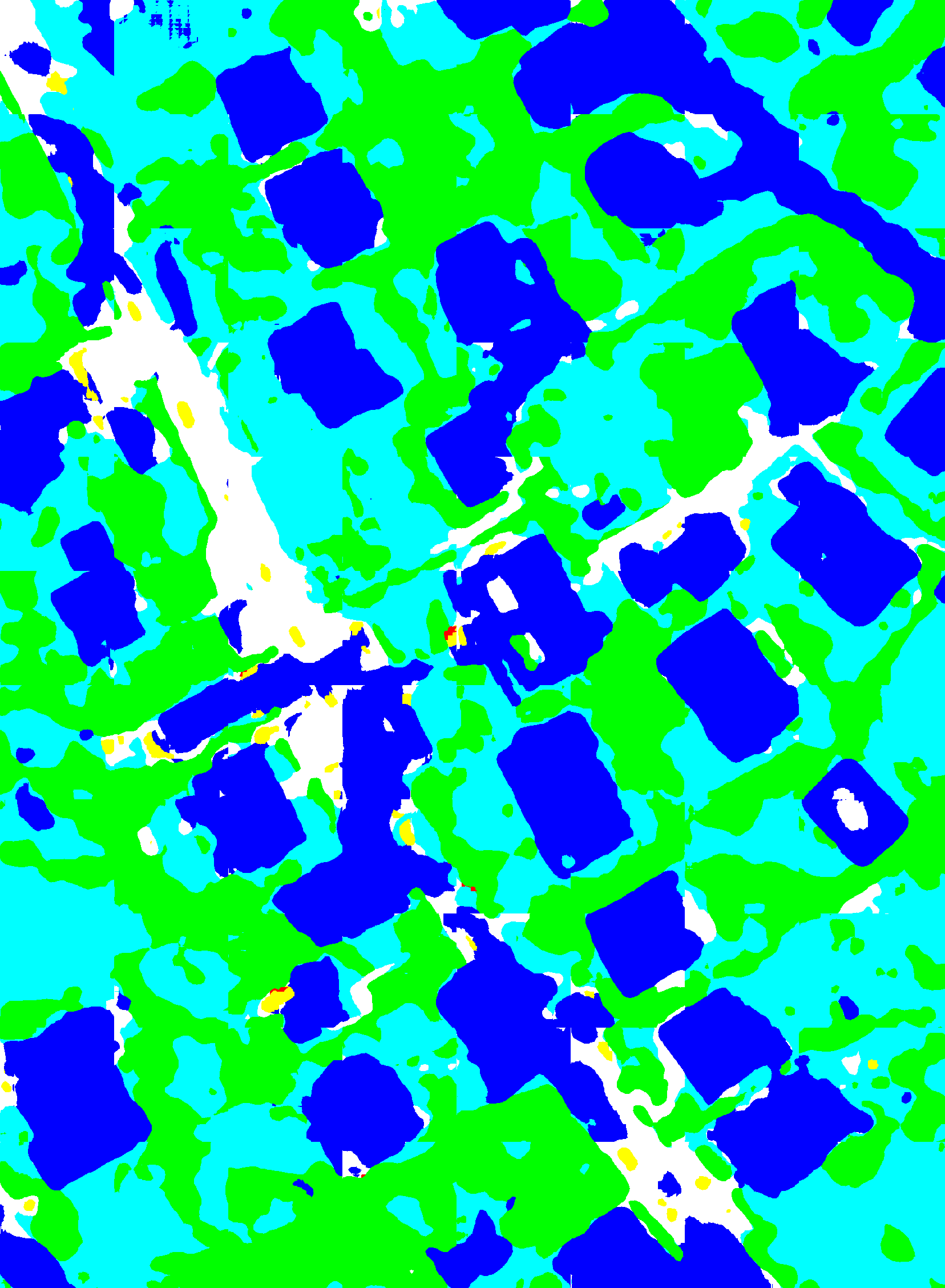}}
			\vspace{3pt}
			\centerline{\includegraphics[width=\textwidth, angle=90]{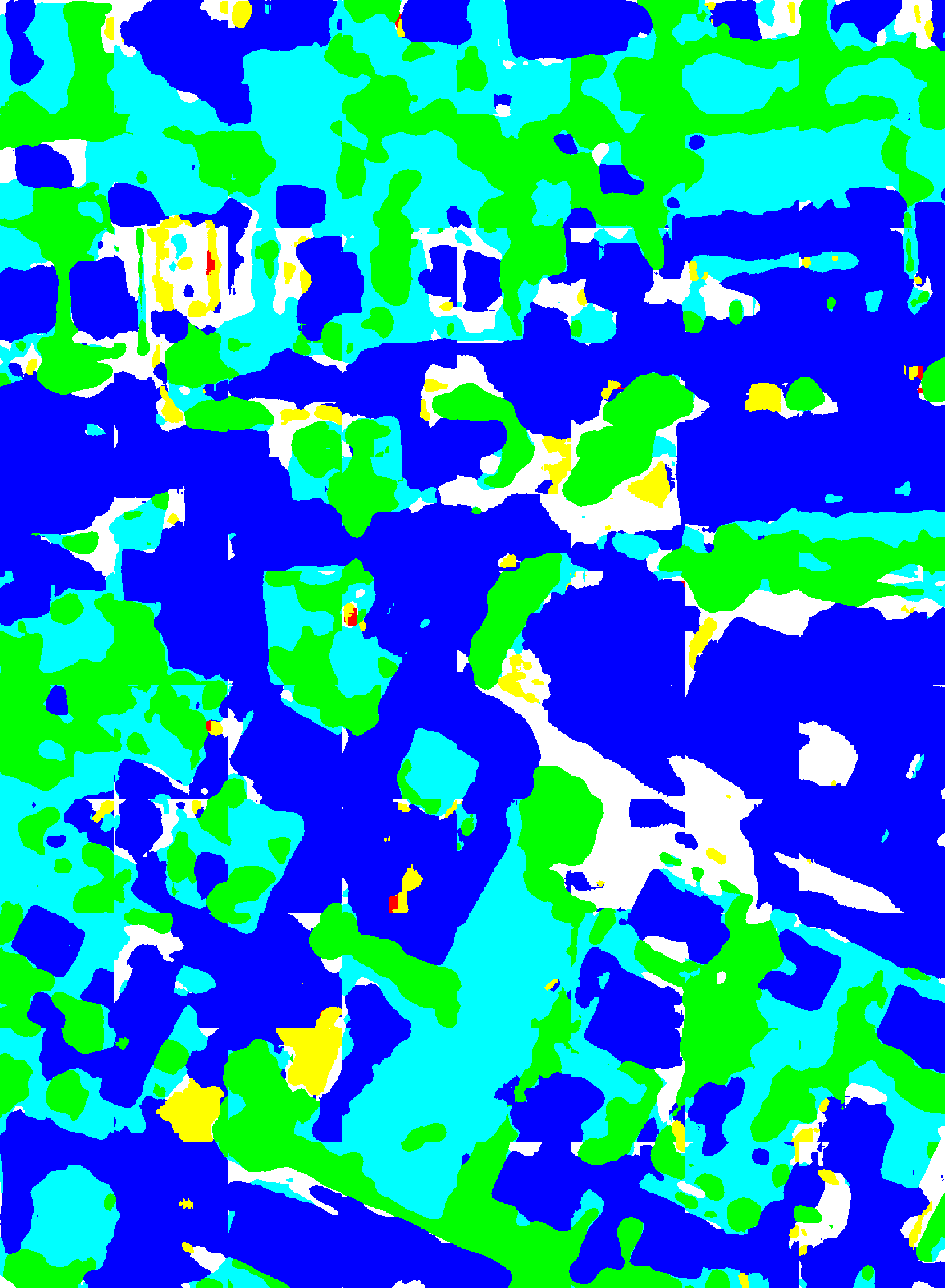}}
			\vspace{3pt}
			\centerline{\includegraphics[width=\textwidth, angle=90]{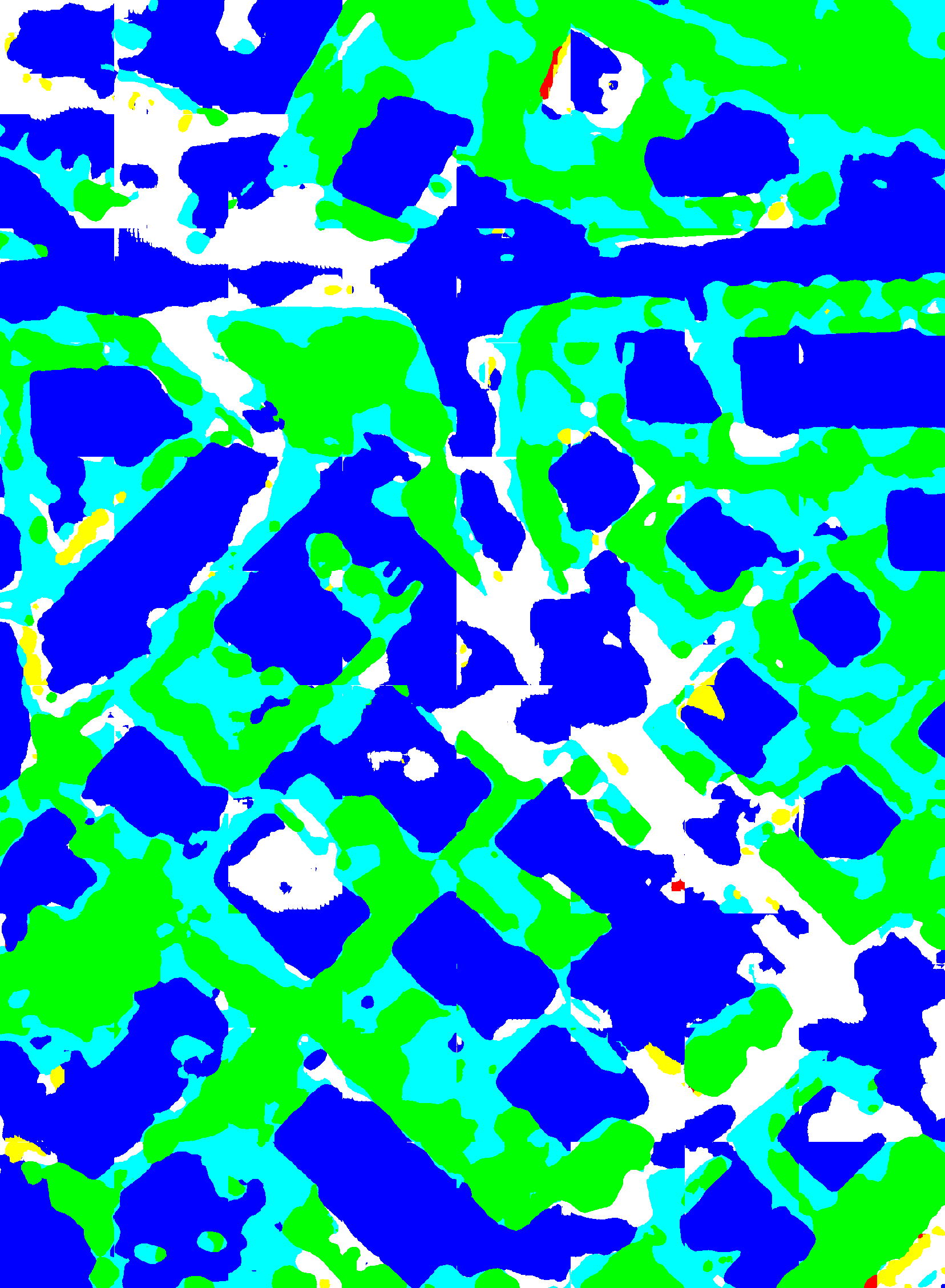}}
			\vspace{3pt}
			\centerline{\includegraphics[scale=0.031, angle=90]{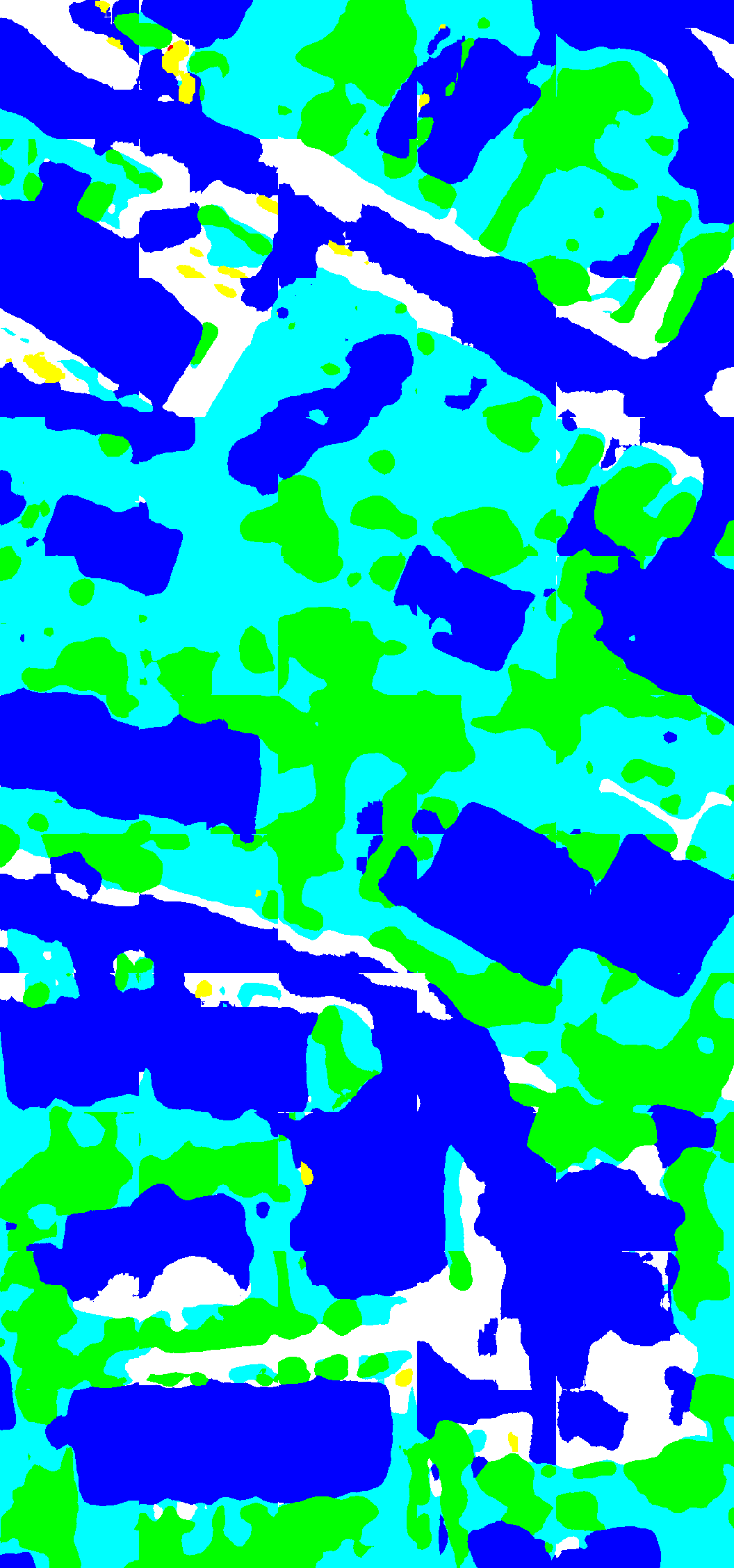}}
			\vspace{3pt}
			\centerline{(g)}
		\end{minipage}
		\caption{Qualitative comparison for adversarial purification on Vaihingen dataset. (a) Image inputs. (b) Ground truth. (c)-(d) Segmetation maps obtained from (c) Adversarial Samples. (d) Clean images. (e)-(g) Segmentation maps of the purified results obtained by (e) Pix2Pix. (f) TGDN. (g) UAD-RS.} 
		\label{VaiQuali}
	\end{figure*}

Figure \ref{VaiQuali} presents the qualitative comparison of adversarial purification and semantic segmentation results on the Vaihingen dataset. Samples are generated using the most powerful attack (CW) based on the quantitative results in Table \ref{Vaihingenquanti}. The predictions of attacked adversarial samples show misclassifications in most areas compared to clean samples. Pix2Pix and TGDN methods partially improve segmentation maps for some categories, but many patches still show residual perturbations. In contrast, UAD-RS purified results successfully restore object edges and background information, closely resembling normal segmentation maps.
	\subsection{Experiments on Zurich Summer Dataset}
 \subsubsection{Quantitative Results}
Table \ref{Zurichquanti} presents the quantitative comparisons of UAD-RS and its competitors for semantic segmentation on the Zurich Summer dataset. Similar to the Vaihingen dataset, adversarial attack algorithms produced varying success rates on the Zurich Summer dataset, with FGSM and Jitter generating mild perturbations and others producing more intensive ones. UAD-RS consistently achieved the best results across most experimental settings, exhibiting the highest OA and F1 values. Pix2Pix provided limited enhancement for victim DNN performance, while TGDN achieved marginally better results than UAD-RS in some attack settings but yielded mediocre results overall.
	% \usepackage{multirow}
\begin{table*}[]
	\centering
	\setlength\tabcolsep{5pt}
	\caption{Quantitative comparisons for adversarial defense of semantic segmentation on Zurich dataset.}
	\label{Zurichquanti}
    \resizebox{\textwidth}{!}{
	\begin{tabular}{cccccccccccccccc}
		\hline
		\multirow{3}{*}{Datasets}        & \multirow{3}{*}{Victims DNNs} & \multicolumn{14}{c}{Attack Algorithms}                                                                                                                                                                                                         \\
		&                               & \multicolumn{2}{c}{FGSM}        & \multicolumn{2}{c}{IFGSM}       & \multicolumn{2}{c}{CW}          & \multicolumn{2}{c}{Jitter}      & \multicolumn{2}{c}{Mixcut}      & \multicolumn{2}{c}{Mixup}       & \multicolumn{2}{c}{TPGD}        \\
		&                               & OA             & F1             & OA             & F1             & OA             & F1             & OA             & F1             & OA             & F1             & OA             & F1             & OA             & F1             \\ \hline
		\multirow{4}{*}{No Defense}      & U-Net                         & 65.63          & 59.90           & 43.26          & 40.44          & 33.53          & 31.07          & 64.95          & 59.35          & 41.69          & 26.41          & 42.63          & 34.39          & 53.76          & 50.56          \\
		& PSPNet                        & 71.72          & 64.53          & 61.40          & 55.76          & 56.55          & 51.51          & 71.32          & 64.24          & 58.17          & 49.72          & 54.88          & 47.93          & 67.88          & 61.77          \\
		& FCN-8s                        & 66.56          & 58.76          & 51.21          & 45.29          & 36.45          & 29.16          & 65.09          & 57.89          & 64.02          & 55.14          & 60.29          & 47.76          & 58.46          & 51.70          \\
		& LinkNet                       & 67.11          & 59.44          & 50.42          & 44.33          & 44.94          & 38.45          & 66.93          & 59.37          & 55.62          & 40.60          & 50.99          & 35.45          & 59.39          & 53.12          \\ \hline
		\multirow{4}{*}{Pix2Pix}         & U-Net                         & 70.06          & 61.53          & 61.55          & 52.52          & 58.74          & 48.83          & 70.24          & 62.20           & 54.80           & 38.19          & 51.83          & 39.61          & 66.39          & 57.44          \\
		& PSPNet                        & 70.35          & 61.35          & 64.79          & 55.64          & 61.74          & 51.99          & 70.51          & 61.59          & \textcolor{red}{62.36} & 48.59          & 51.71          & 41.04          & 68.56          & 58.83          \\
		& FCN-8s                        & 69.47          & 60.23          & 58.91          & 50.08          & 49.55          & 37.20          & 69.09          & 60.77          & 57.95          & 47.13          & 57.55          & 46.20          & 65.67          & 56.21          \\
		& LinkNet                       & 67.98          & 57.58          & 60.70          & 49.9           & 58.01          & 45.57          & 68.36          & 58.02          & 58.48          & 41.90          & 56.42          & 39.33          & 65.67          & 54.04          \\ \hline
		\multirow{4}{*}{TGDN}            & U-Net                         & 66.67          & 52.86          & 53.00          & 39.15          & 46.54          & 32.08          & 66.29          & 52.45          & 49.21          & 40.51          & 38.28          & 24.39          & 55.61          & 41.67          \\
		& PSPNet                        & \textcolor{red}{80.85} & 64.71          & 67.78          & 52.54          & 64.10          & 48.19          & \textcolor{red}{80.37} & 64.20          & 59.01          & 40.60          & \textcolor{red}{56.66} & 38.34          & 70.48          & 55.35          \\
		& FCN-8s                        & 63.26          & 48.38          & 51.74          & 34.86          & 41.87          & 25.89          & 65.20          & 50.05          & 48.18          & 30.87          & 48.22          & 31.55          & 57.91          & 41.39          \\
		& LinkNet                       & 69.85          & 54.25          & 59.65          & 44.54          & 56.46          & 42.70          & 69.69          & 54.09          & \textcolor{red}{64.42} & \textcolor{red}{50.2}  & \textcolor{red}{60.53} & \textcolor{red}{46.69} & 61.50          & 46.13          \\ \hline
		 & U-Net                         & \textcolor{red}{71.82} & \textcolor{red}{63.80} & \textcolor{red}{66.24} & \textcolor{red}{58.98} & \textcolor{red}{64.35} & \textcolor{red}{56.26} & \textcolor{red}{71.93} & \textcolor{red}{64.21} & \textcolor{red}{59.00} & \textcolor{red}{48.13} & \textcolor{red}{59.19} & \textcolor{red}{51.04} & \textcolor{red}{68.98} & \textcolor{red}{61.84} \\
	UAD-RS	& PSPNet                        & 73.33          & \textcolor{red}{65.54} & \textcolor{red}{69.21} & \textcolor{red}{61.84} & \textcolor{red}{66.97} & \textcolor{red}{60.04} & 73.39          & \textcolor{red}{65.65} & 61.51          & \textcolor{red}{51.12} & 53.12          & \textcolor{red}{42.72} & \textcolor{red}{72.44} & \textcolor{red}{64.42} \\
	(Ours)	& FCN-8s                        & \textcolor{red}{71.78} & \textcolor{red}{63.13} & \textcolor{red}{65.22} & \textcolor{red}{56.67} & \textcolor{red}{62.59} & \textcolor{red}{53.51} & \textcolor{red}{71.56} & \textcolor{red}{63.32} & \textcolor{red}{62.08} & \textcolor{red}{50.54} & \textcolor{red}{62.06} & \textcolor{red}{50.26} & \textcolor{red}{71.12} & \textcolor{red}{62.06} \\
		& LinkNet                       & \textcolor{red}{71.09} & \textcolor{red}{63.01} & \textcolor{red}{65.06} & \textcolor{red}{57.73} & \textcolor{red}{63.78} & \textcolor{red}{56.65} & \textcolor{red}{71.90} & \textcolor{red}{63.84} & 59.70          & 48.77          & 56.54          & 44.94          & \textcolor{red}{69.78} & \textcolor{red}{62.02} \\ \hline
	\end{tabular}}
\end{table*}
	\subsubsection{Qualitative Results}
\begin{figure*}[]
		\centering
		\includegraphics[page=2,scale=0.82]{figures/pdf/colors.pdf}
		\begin{minipage}{0.10\linewidth}
			\vspace{3pt}
			\centerline{\includegraphics[width=1.1\textwidth, angle=90]{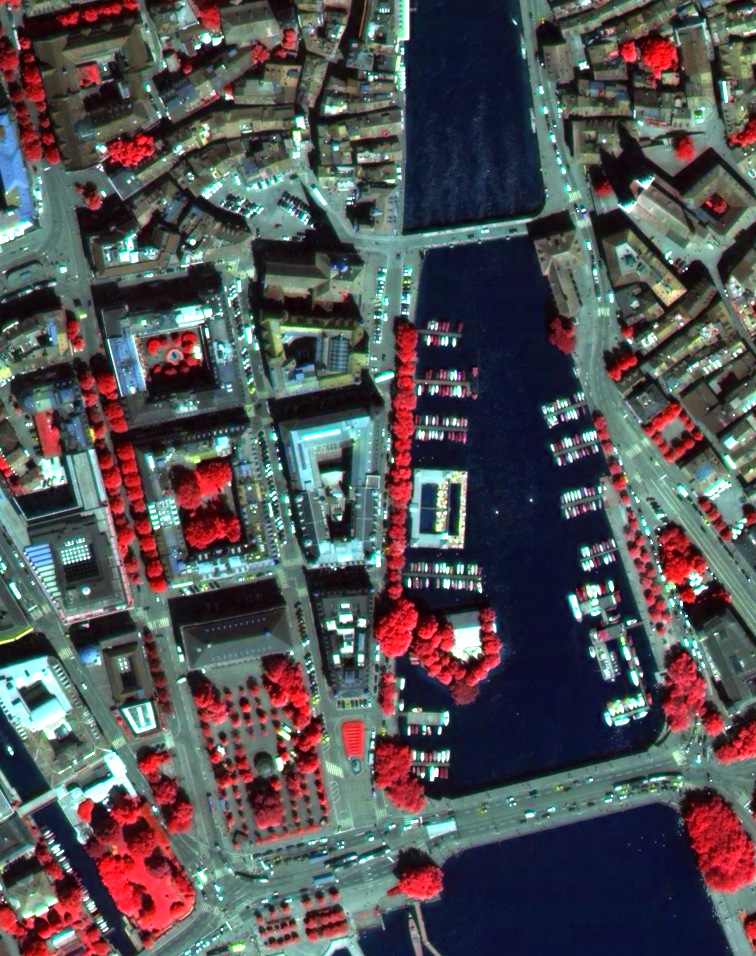}}
			\vspace{3pt}
			\centerline{\includegraphics[width=1.39\textwidth]{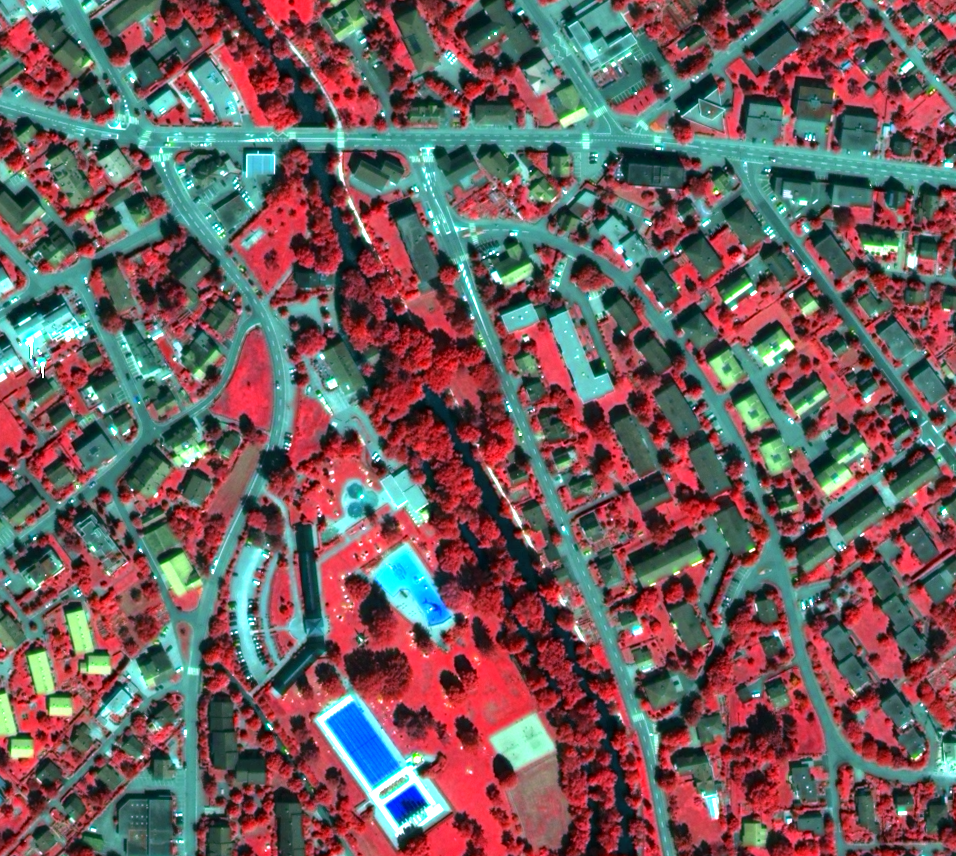}}
			\vspace{3pt}
			\centerline{\includegraphics[width=1.39\textwidth]{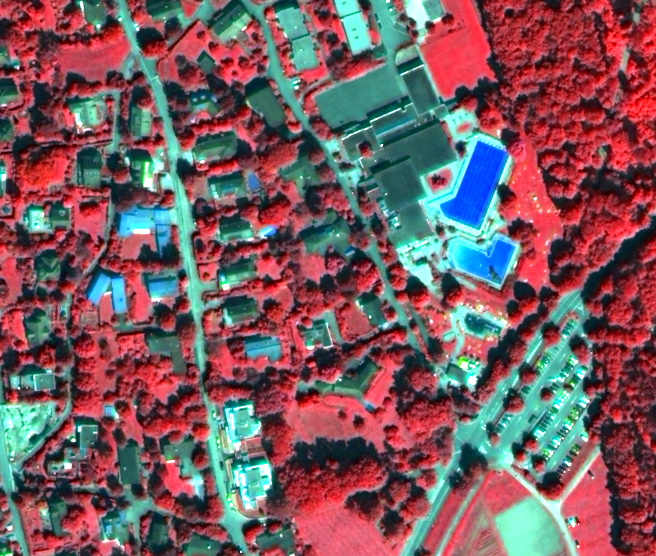}}
			\vspace{3pt}
			\centerline{\includegraphics[width=1.39\textwidth]{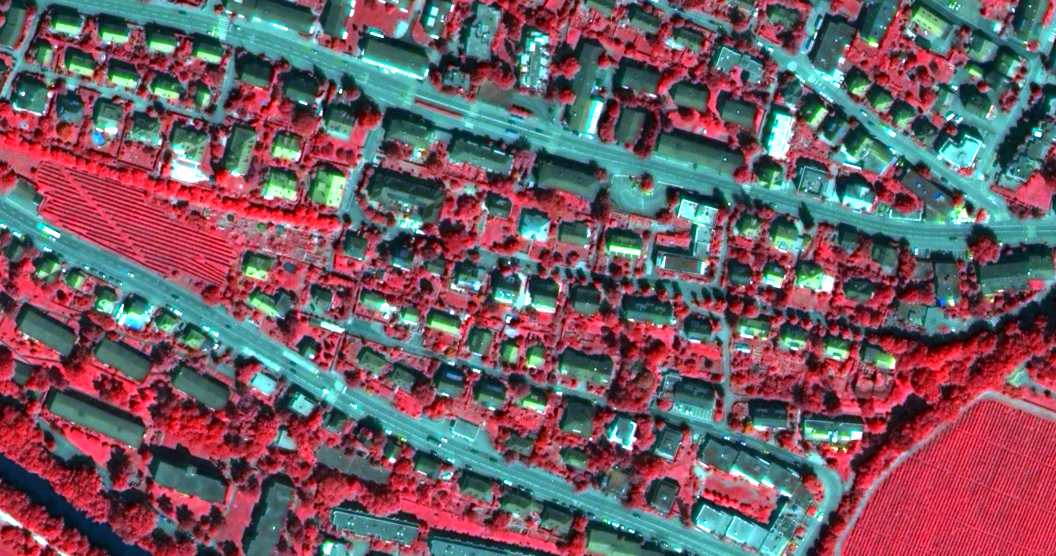}}
			\vspace{3pt}
			\centerline{\includegraphics[width=1.39\textwidth]{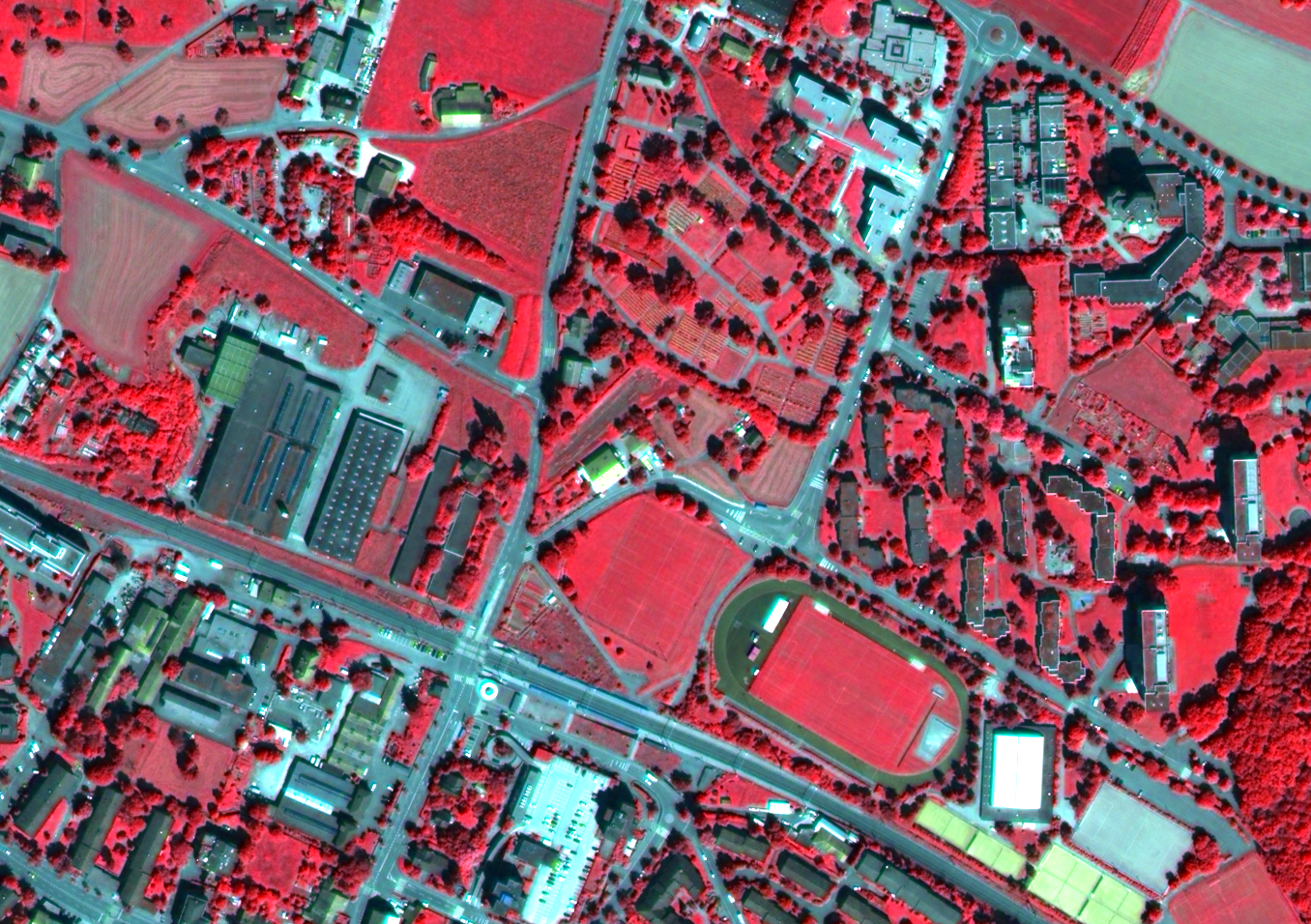}}
			\vspace{3pt}
			\centerline{(a)}
		\end{minipage}
		\hspace{16pt}
		\begin{minipage}{0.10\linewidth}
			\vspace{3pt}
			\centerline{\includegraphics[width=1.1\textwidth, angle=90]{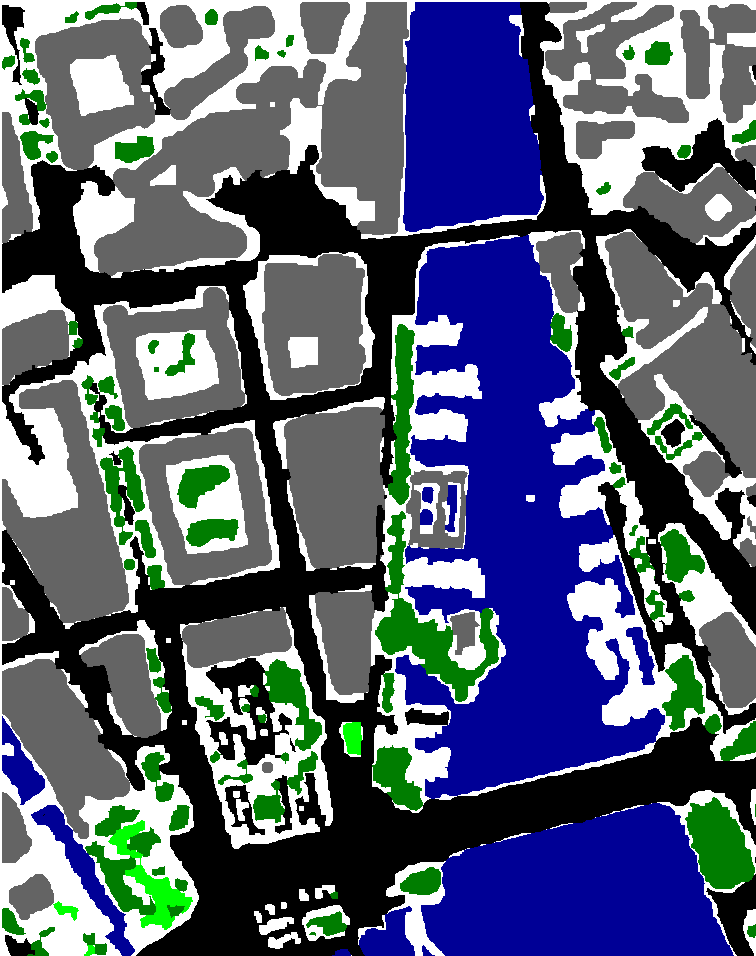}}
			\vspace{3pt}
			\centerline{\includegraphics[width=1.39\textwidth]{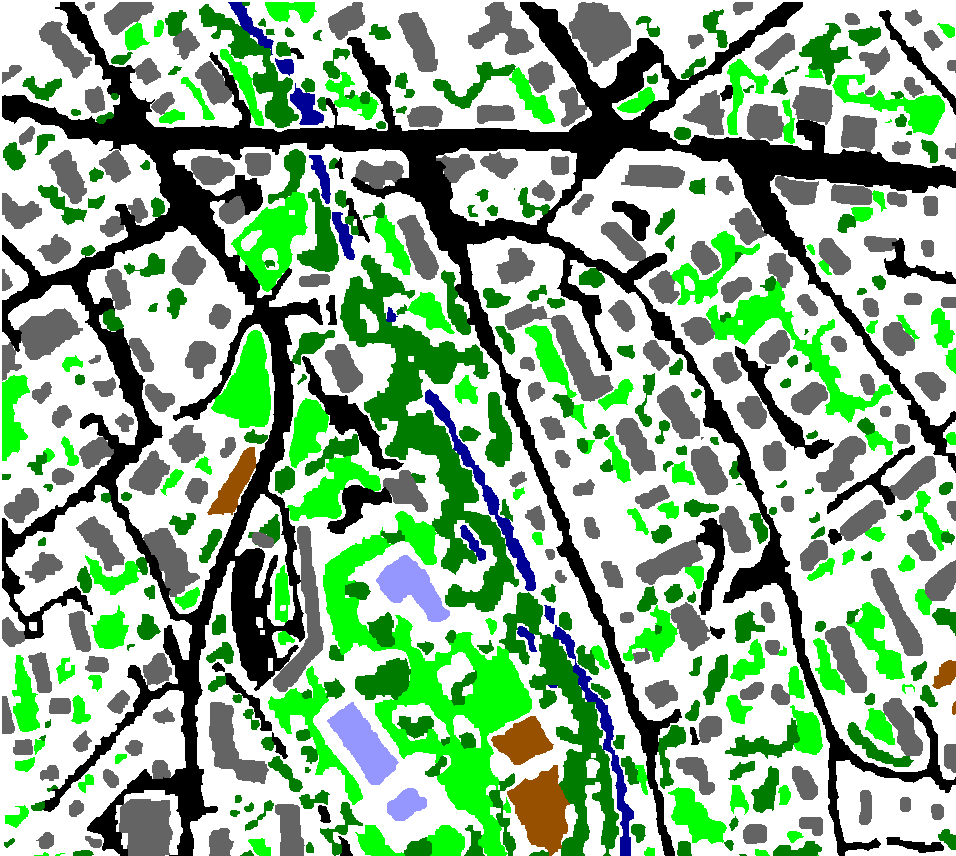}}
			\vspace{3pt}
			\centerline{\includegraphics[width=1.39\textwidth]{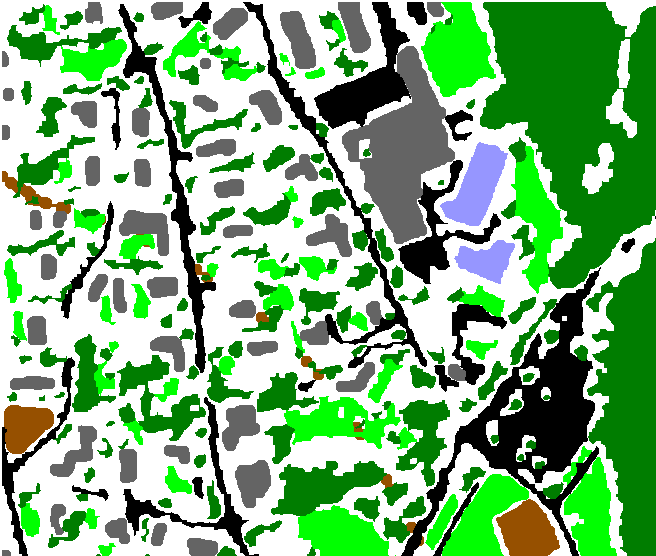}}
			\vspace{3pt}
			\centerline{\includegraphics[width=1.39\textwidth]{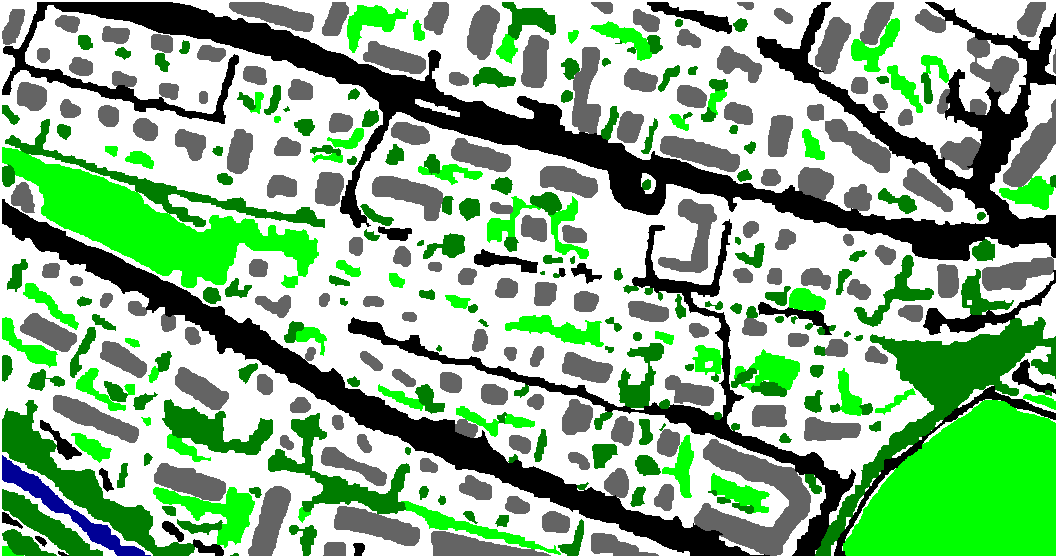}}
			\vspace{3pt}
			\centerline{\includegraphics[width=1.39\textwidth]{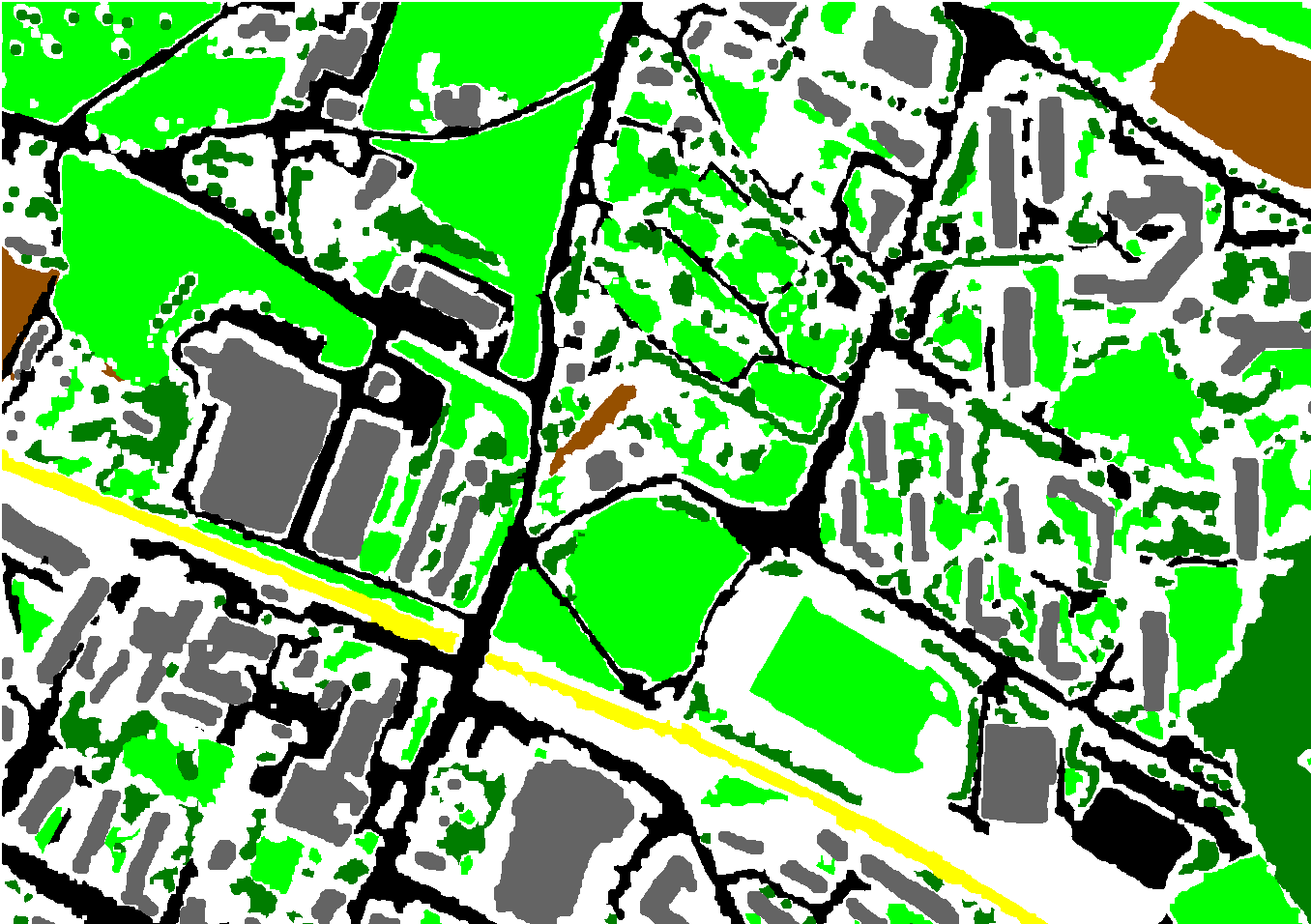}}
			\vspace{3pt}
			\centerline{(b)}
		\end{minipage}
		\hspace{16pt}
		\begin{minipage}{0.10\linewidth}
			\vspace{3pt}
			\centerline{\includegraphics[width=1.1\textwidth, angle=90]{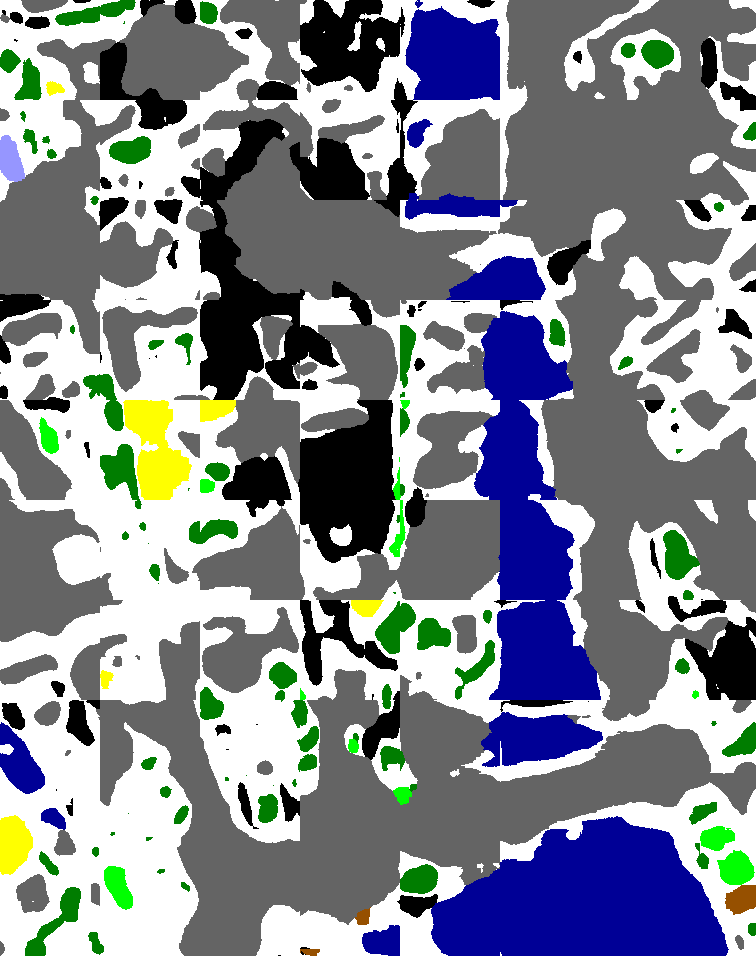}}
			\vspace{3pt}
			\centerline{\includegraphics[width=1.39\textwidth]{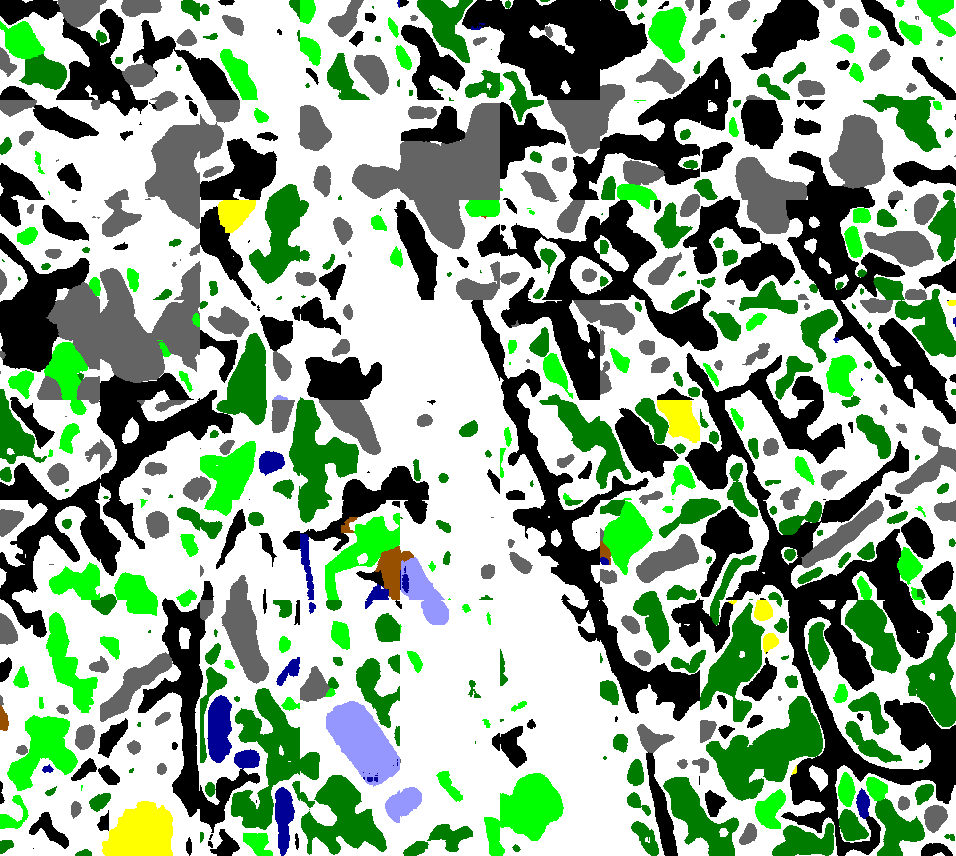}}
			\vspace{3pt}
			\centerline{\includegraphics[width=1.39\textwidth]{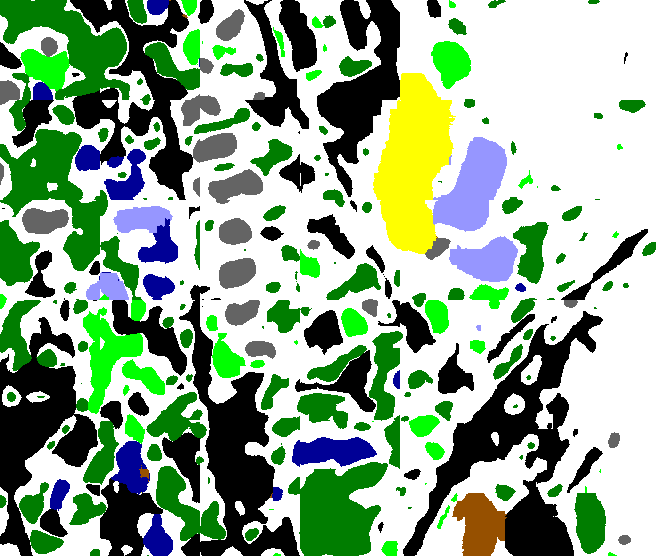}}
			\vspace{3pt}
			\centerline{\includegraphics[width=1.39\textwidth]{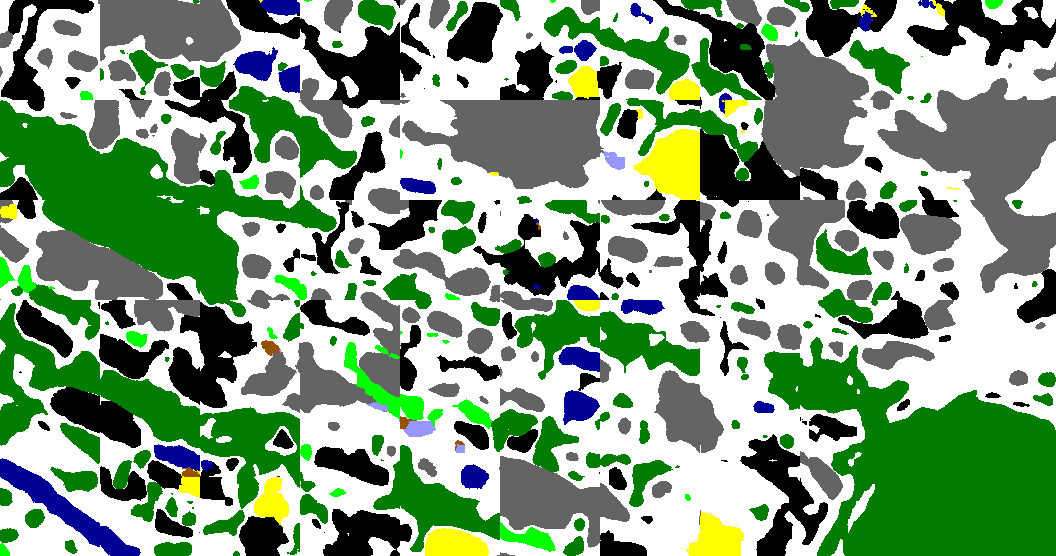}}
			\vspace{3pt}
			\centerline{\includegraphics[width=1.39\textwidth]{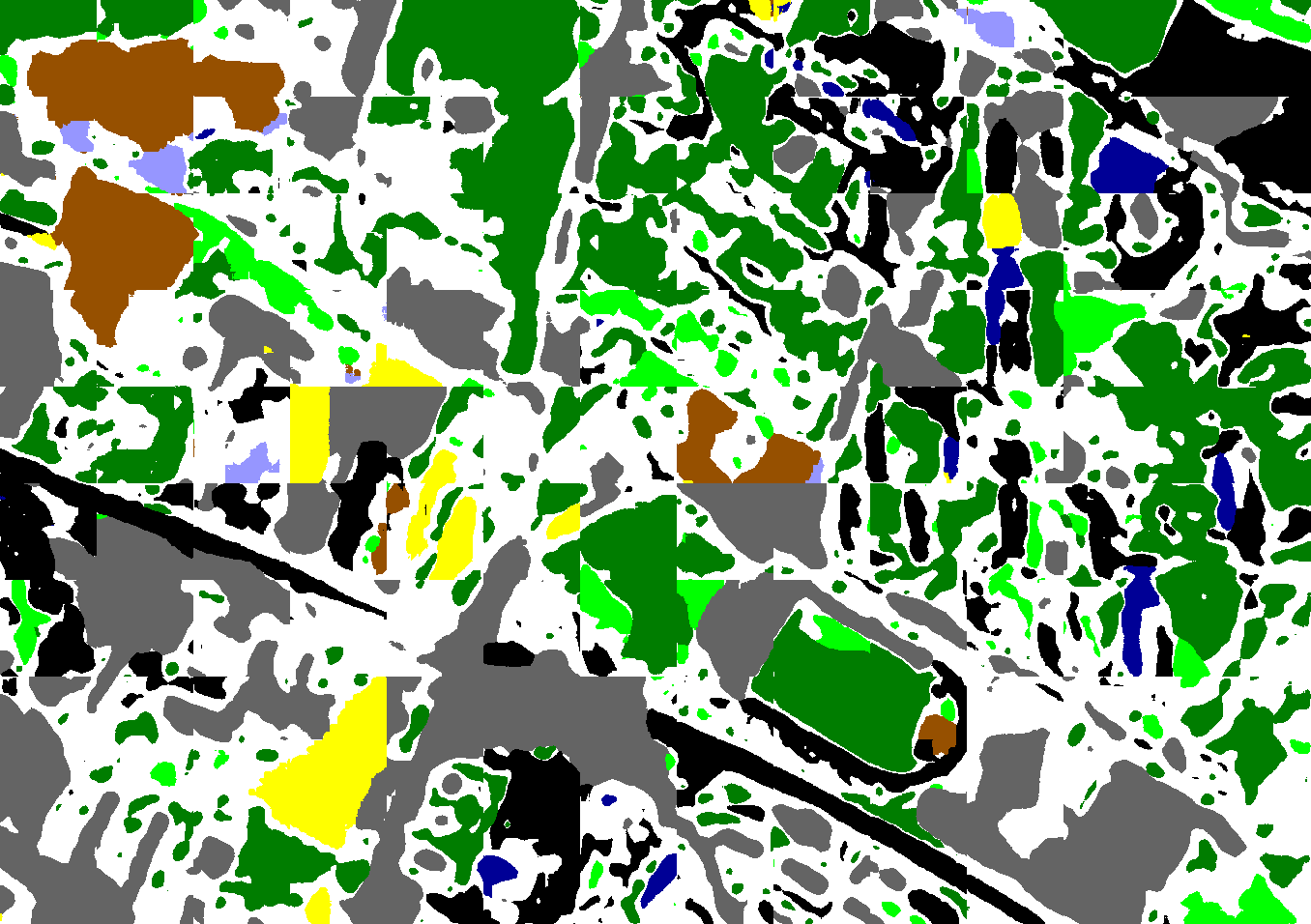}}
			\vspace{3pt}
			\centerline{(c)}
		\end{minipage}
		\hspace{16pt}
		\begin{minipage}{0.10\linewidth}
			\vspace{3pt}
			\centerline{\includegraphics[width=1.1\textwidth, angle=90]{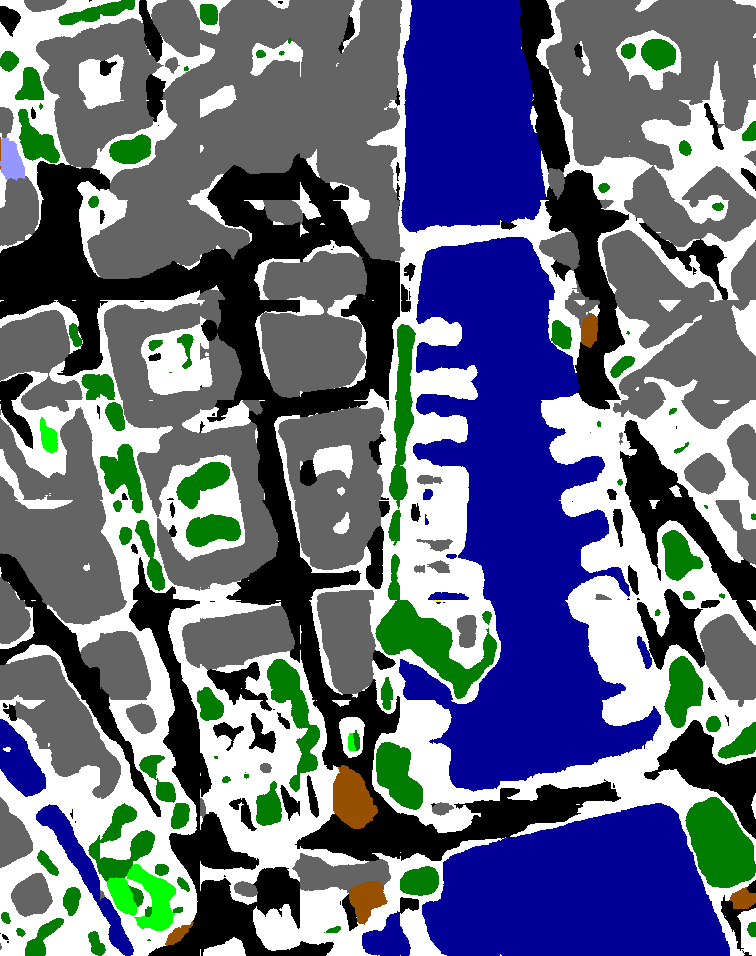}}
			\vspace{3pt}
			\centerline{\includegraphics[width=1.39\textwidth]{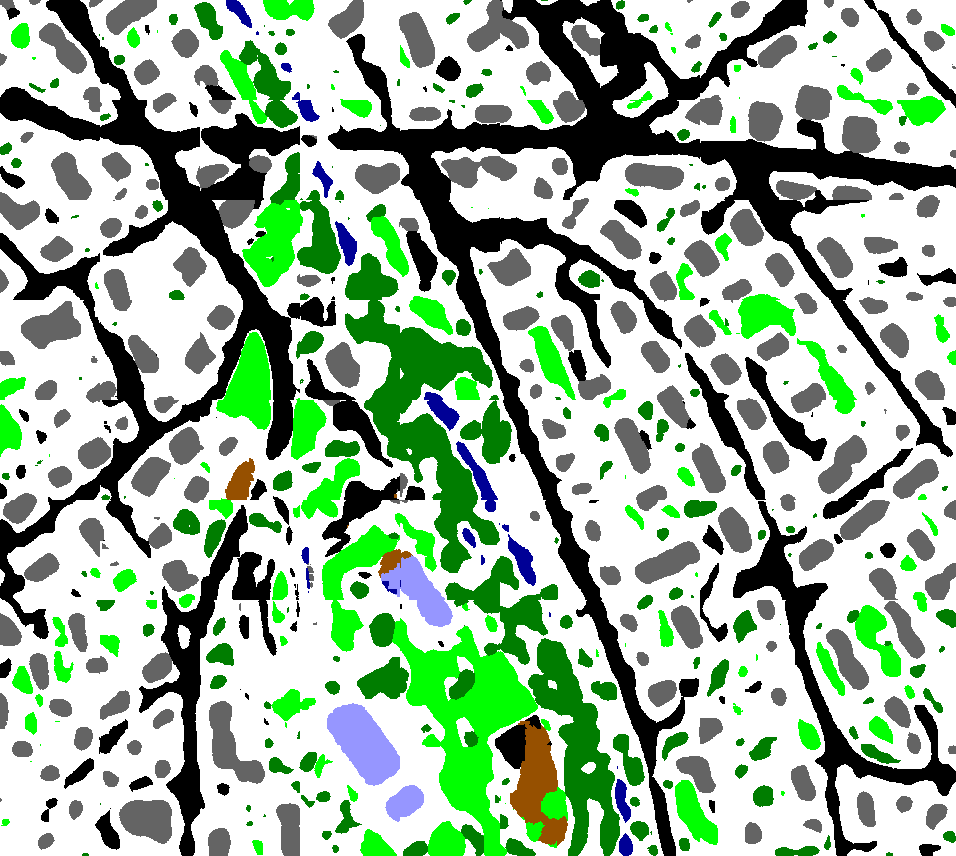}}
			\vspace{3pt}
			\centerline{\includegraphics[width=1.39\textwidth]{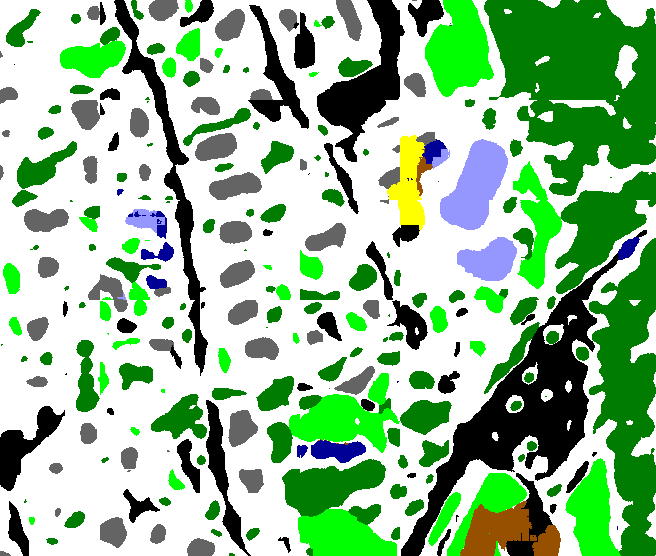}}
			\vspace{3pt}
			\centerline{\includegraphics[width=1.39\textwidth]{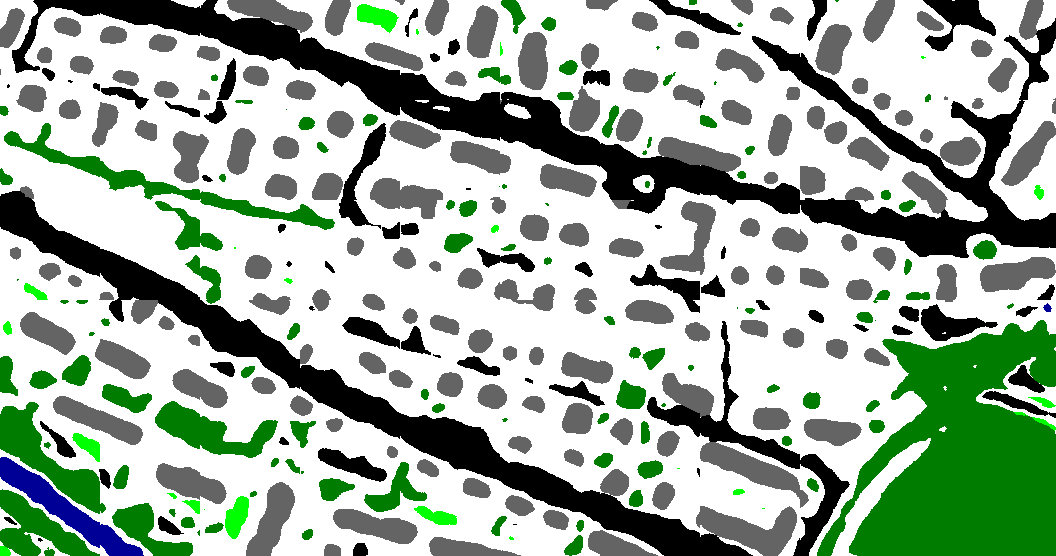}}
			\vspace{3pt}
			\centerline{\includegraphics[width=1.39\textwidth]{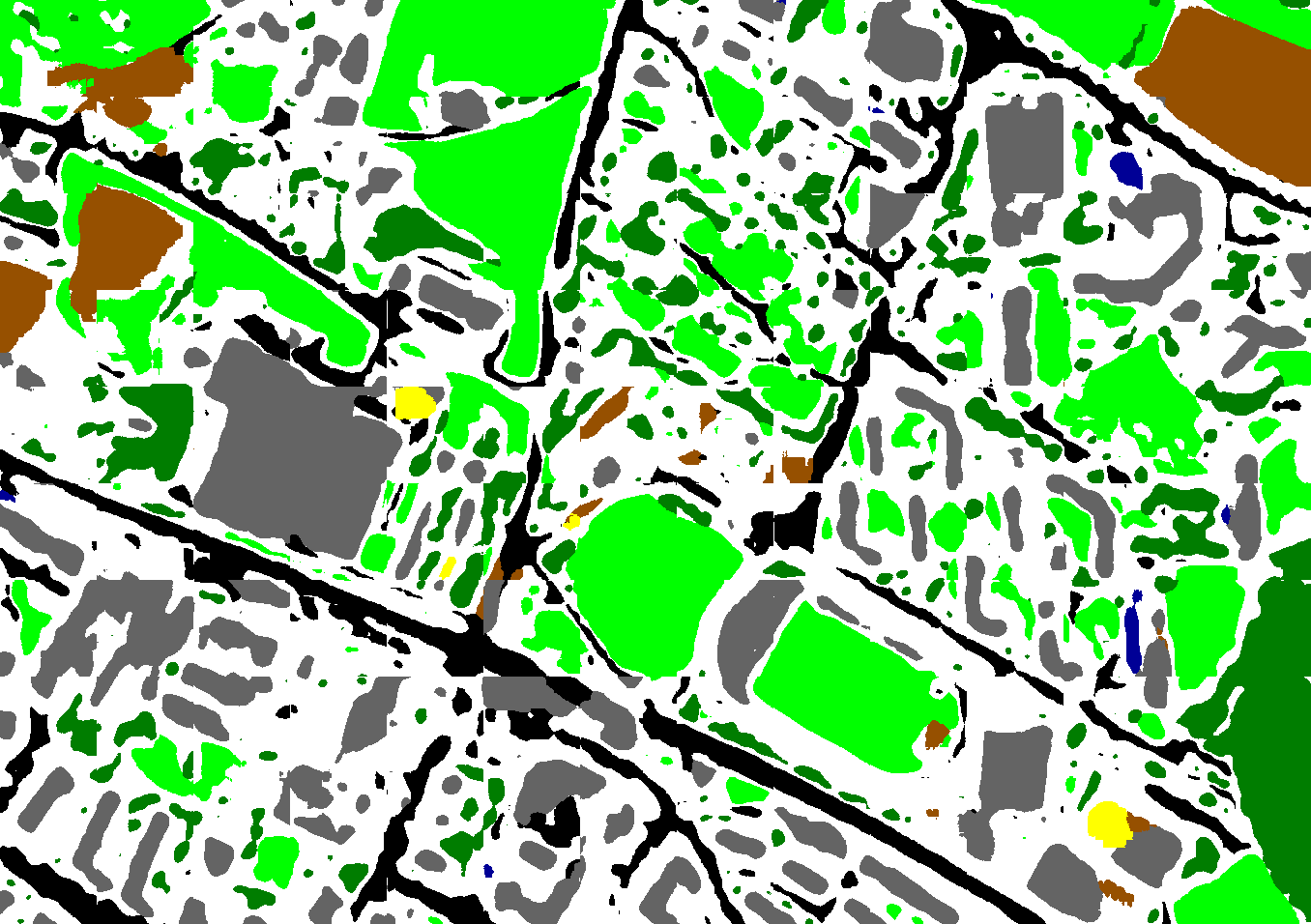}}
			\vspace{3pt}
			\centerline{(d)}
		\end{minipage}
		\hspace{16pt}
		\begin{minipage}{0.10\linewidth}
			\vspace{3pt}
			\centerline{\includegraphics[width=1.1\textwidth, angle=90]{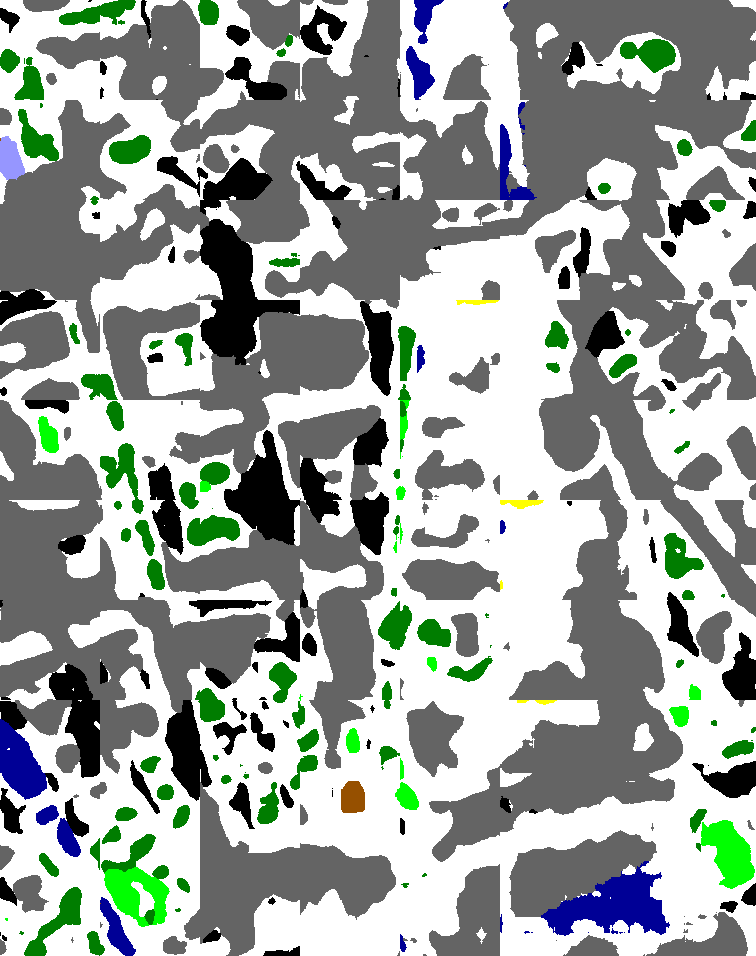}}
			\vspace{3pt}
			\centerline{\includegraphics[width=1.39\textwidth]{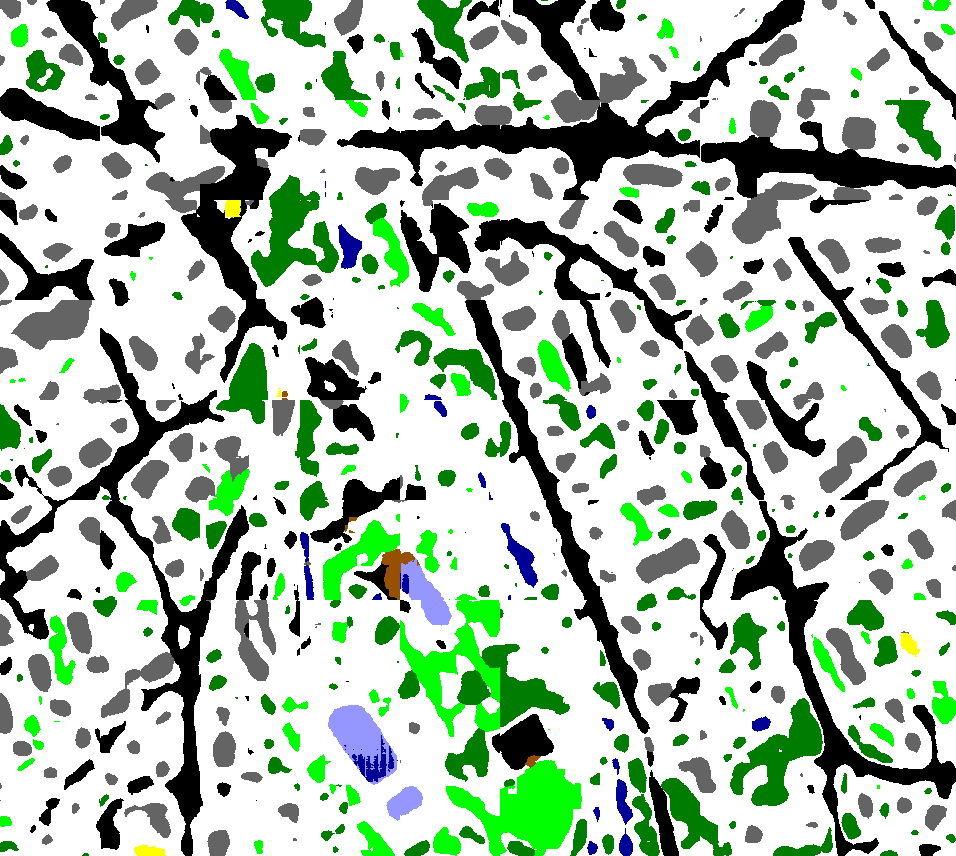}}
			\vspace{3pt}
			\centerline{\includegraphics[width=1.39\textwidth]{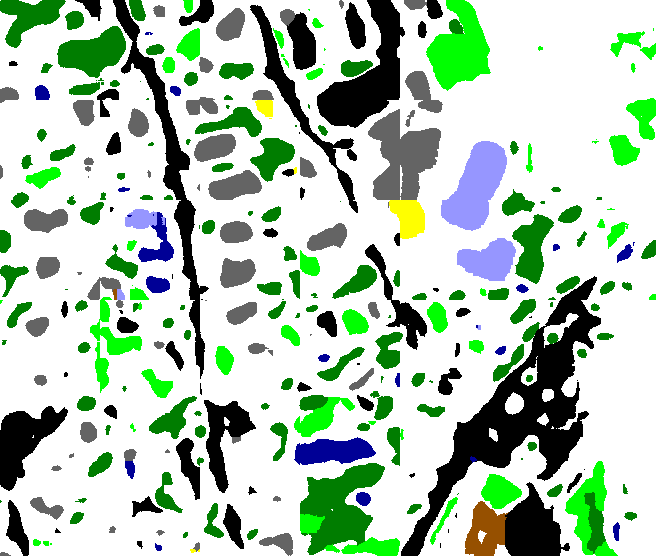}}
			\vspace{3pt}
			\centerline{\includegraphics[width=1.39\textwidth]{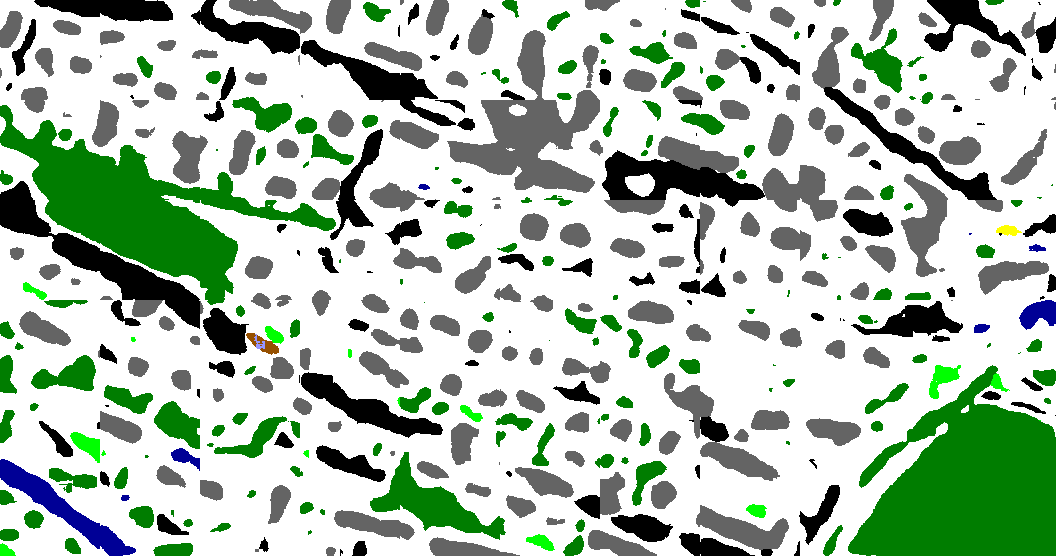}}
			\vspace{3pt}
			\centerline{\includegraphics[width=1.39\textwidth]{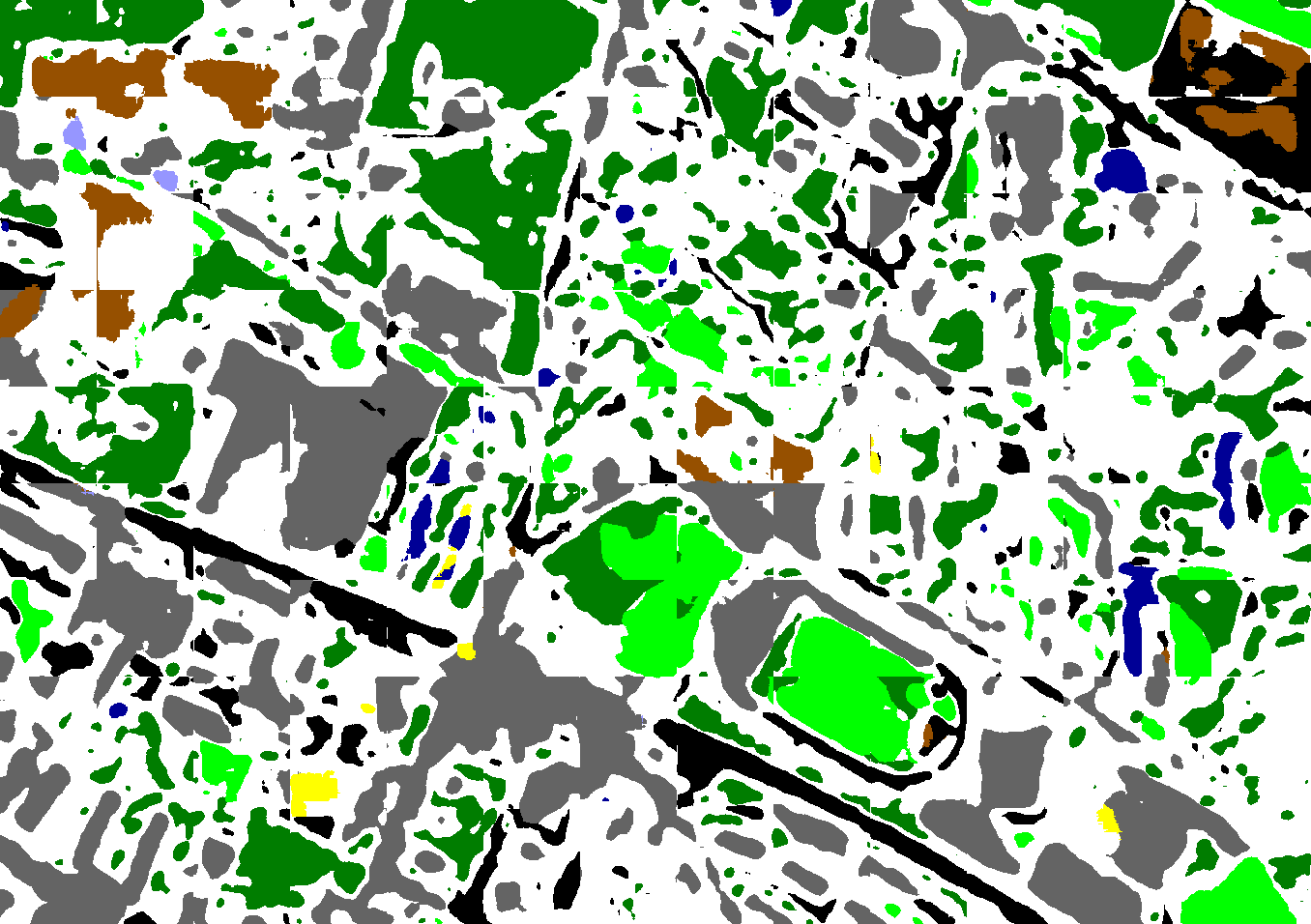}}
			\vspace{3pt}
			\centerline{(e)}
		\end{minipage}
		\hspace{16pt}
		\begin{minipage}{0.10\linewidth}
			\vspace{3pt}
			\centerline{\includegraphics[width=1.1\textwidth, angle=90]{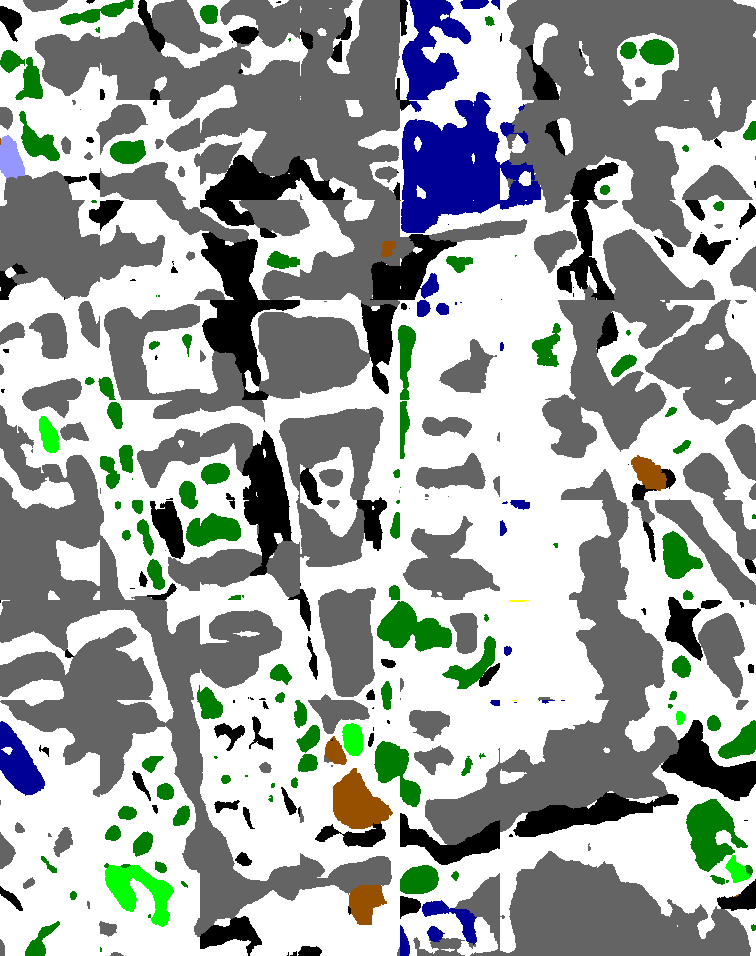}}
			\vspace{3pt}
			\centerline{\includegraphics[width=1.39\textwidth]{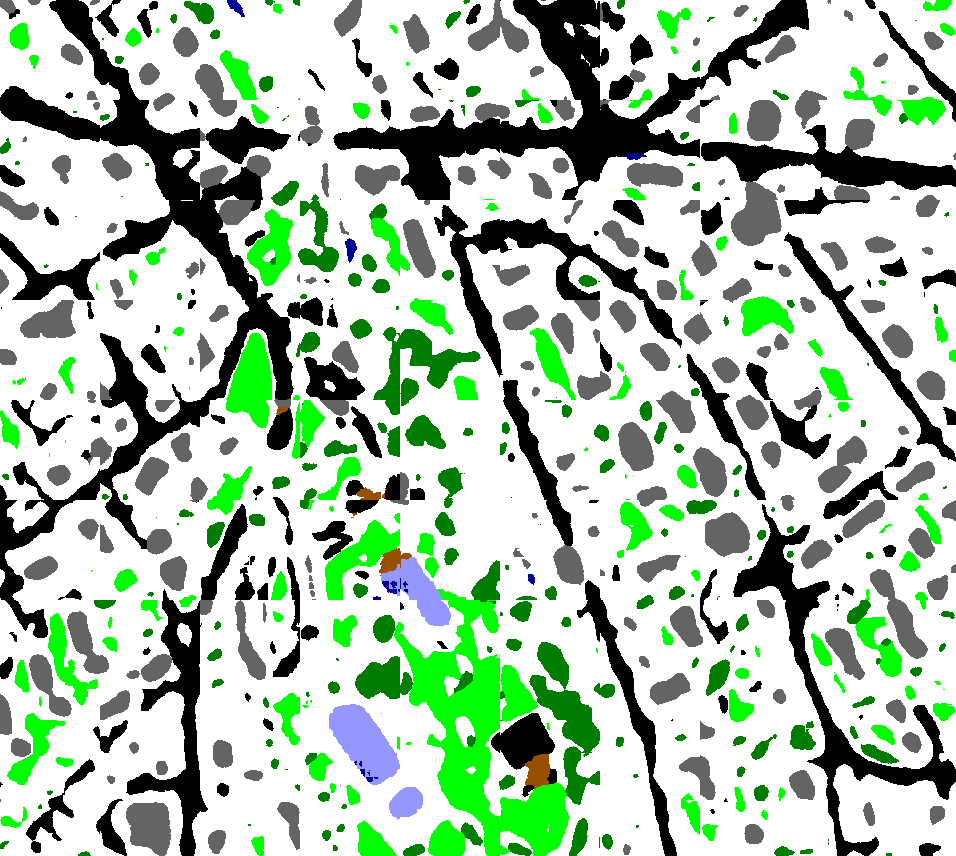}}
			\vspace{3pt}
			\centerline{\includegraphics[width=1.39\textwidth]{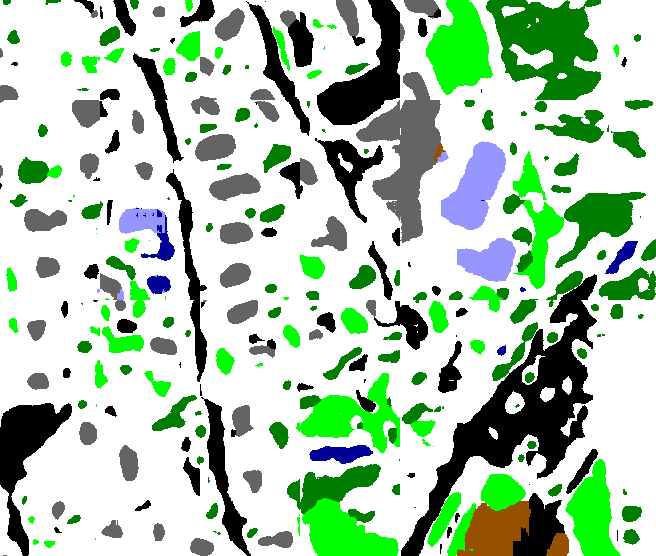}}
			\vspace{3pt}
			\centerline{\includegraphics[width=1.39\textwidth]{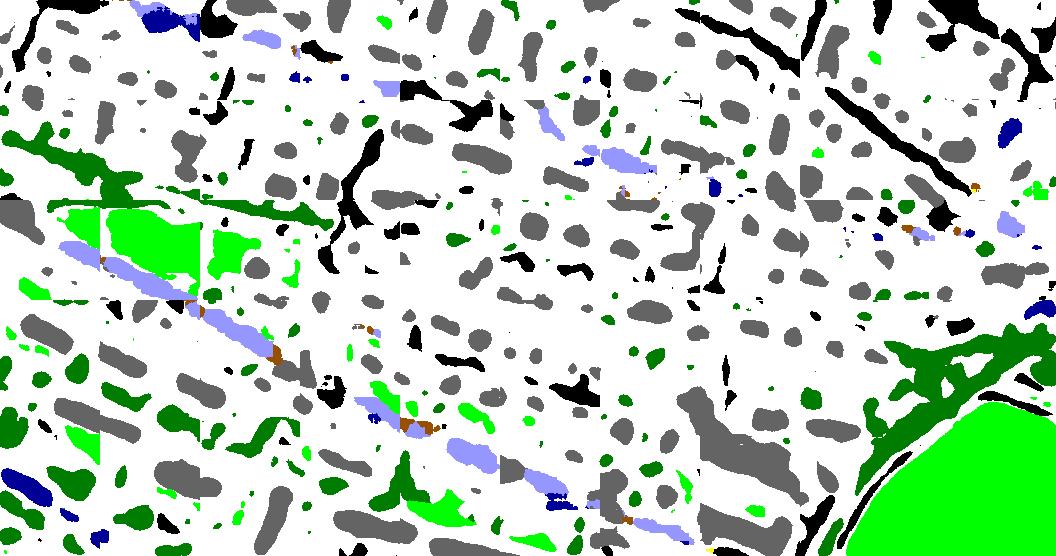}}
			\vspace{3pt}
			\centerline{\includegraphics[width=1.39\textwidth]{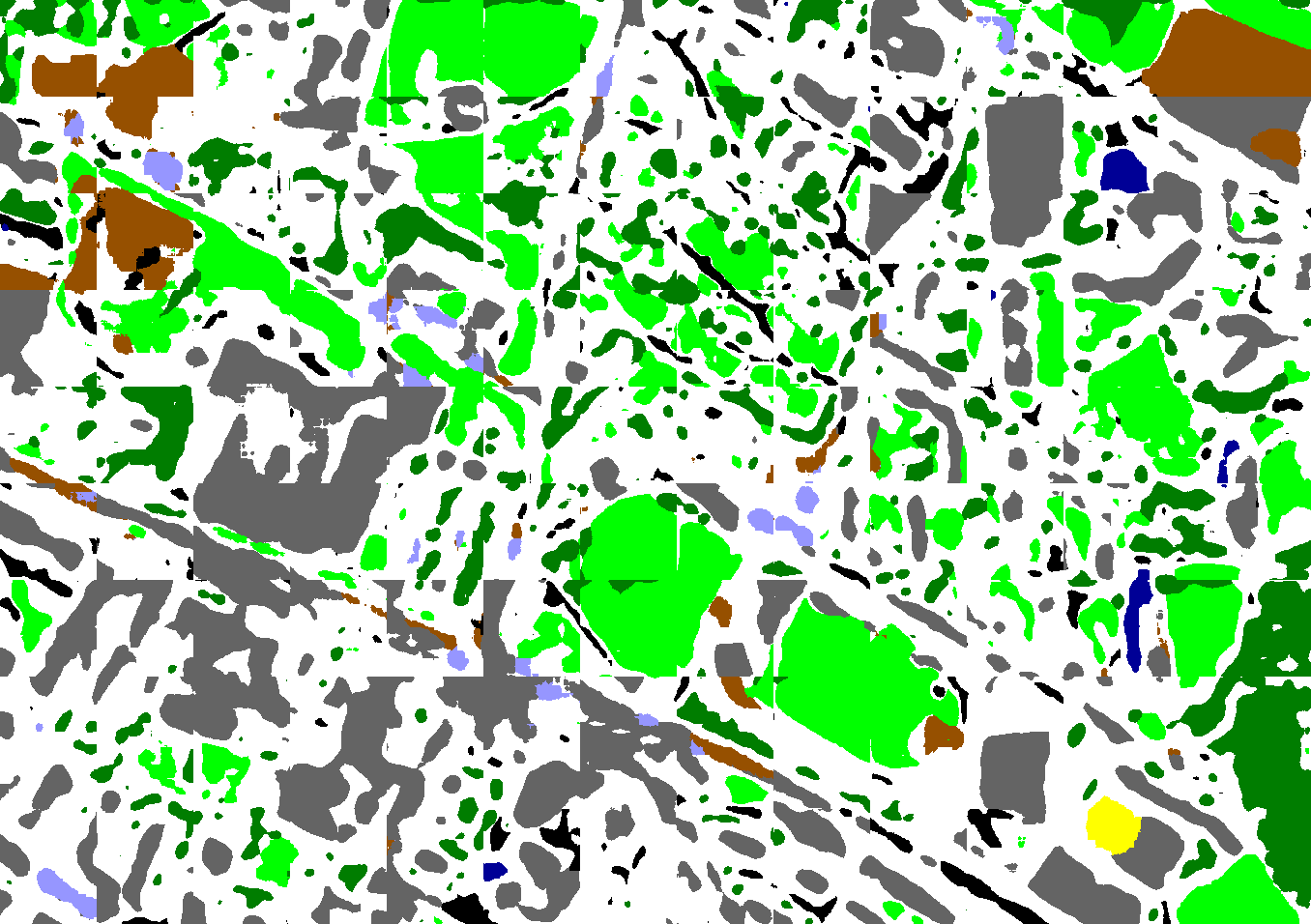}}
			\vspace{3pt}
			\centerline{(f)}
		\end{minipage}
		\hspace{16pt}
		\begin{minipage}{0.10\linewidth}
			\vspace{3pt}
			\centerline{\includegraphics[width=1.1\textwidth, angle=90]{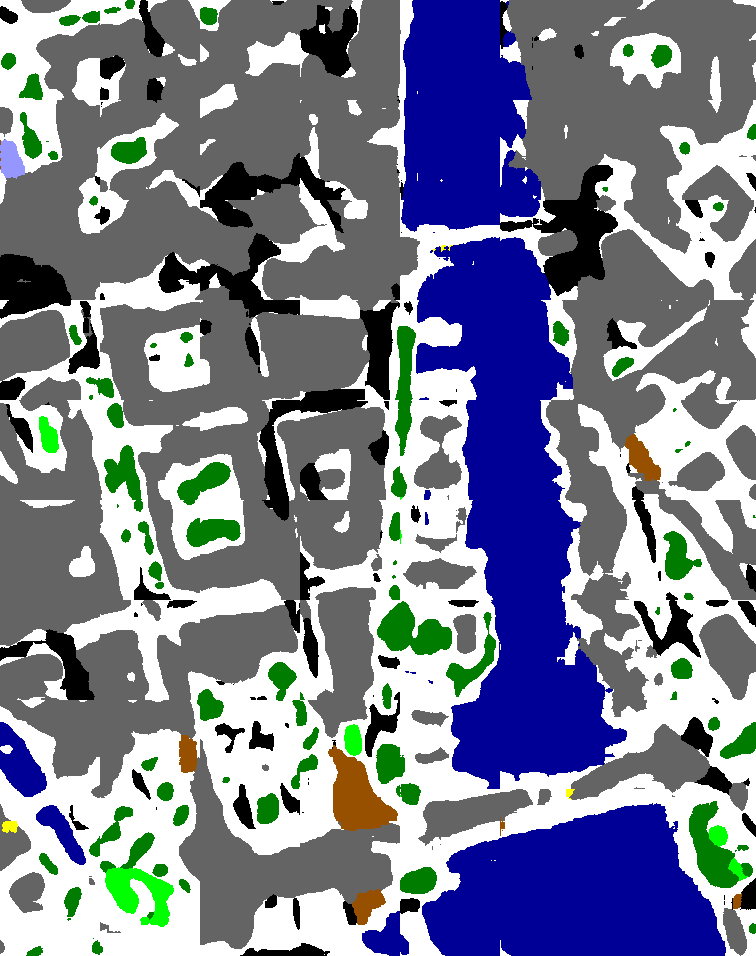}}
			\vspace{3pt}
			\centerline{\includegraphics[width=1.39\textwidth]{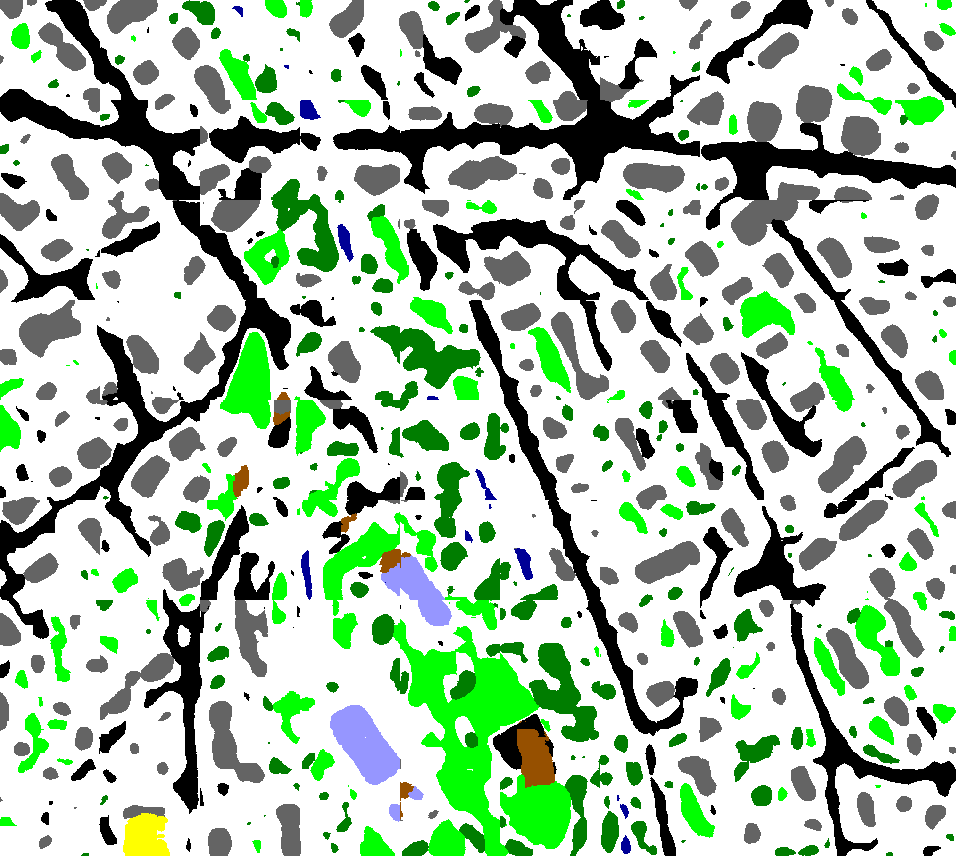}}
			\vspace{3pt}
			\centerline{\includegraphics[width=1.39\textwidth]{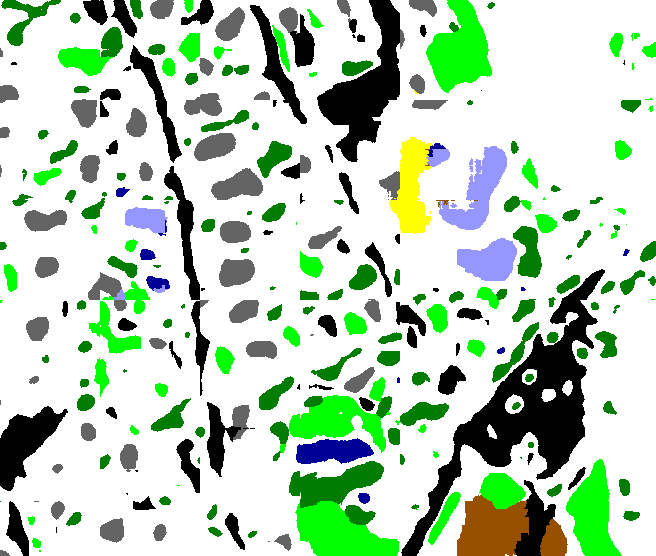}}
			\vspace{3pt}
			\centerline{\includegraphics[width=1.39\textwidth]{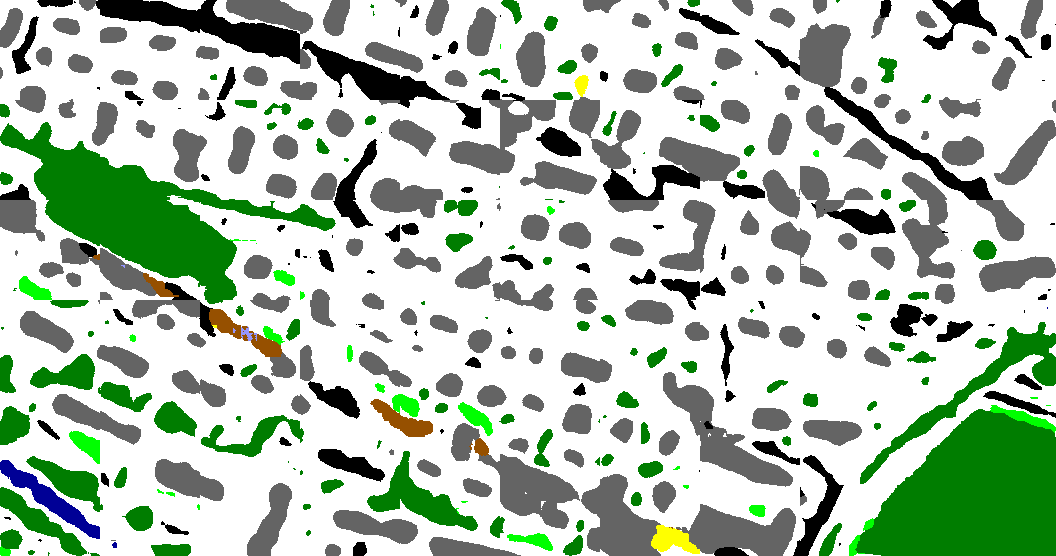}}
			\vspace{3pt}
			\centerline{\includegraphics[width=1.39\textwidth]{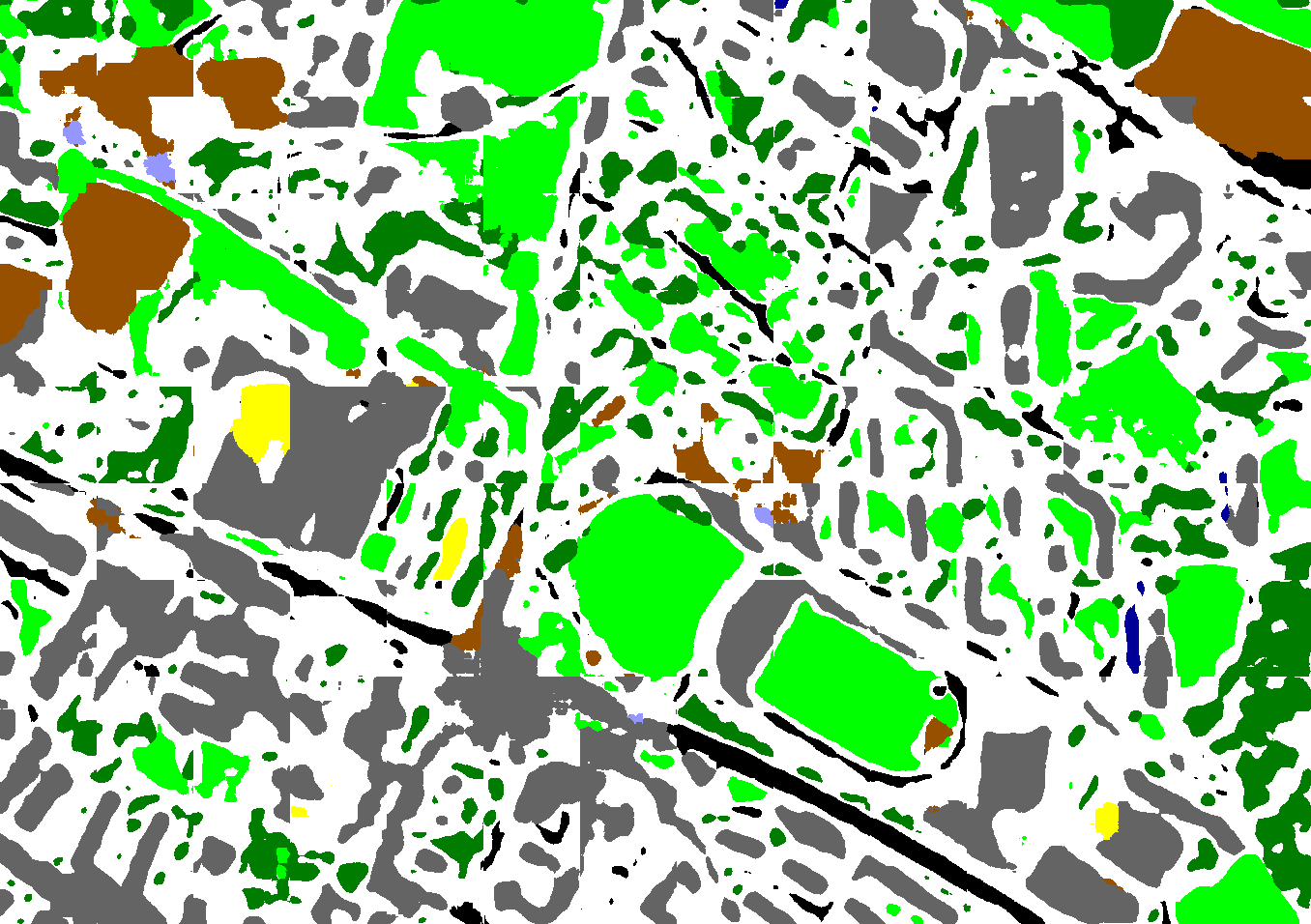}}
			\vspace{3pt}
			\centerline{(g)}
		\end{minipage}
		\caption{Qualitative comparison for adversarial purification on Zurich Summer dataset. (a) Image inputs. (b) Ground truth. (c)-(d) Segmetation maps obtained from (c) Adversarial samples. (d) Clean images. (e)-(g) Segmentation maps of the purified results obtained by (e) Pix2Pix. (f) TGDN. (g) UAD-RS.} 
		\label{ZurQuali}
	\end{figure*}

Figure \ref{ZurQuali} presents qualitative comparisons of segmentation results from all methods on the Zurich Summer dataset under CW attacks. Segmentation maps from different purification methods exhibit omission errors that are not adequately improved from the results of adversarial samples. Pix2Pix and TGDN methods notably miss many pixels, classifying them as background, particularly for water objects. In contrast, UAD-RS recovers most pixels, accurately segmenting high-precision intact targets, closely resembling predictions of clean images.
	\subsection{Ablation Studies}
	\subsubsection{RS Image Synthesis}
\begin{figure}
		\centering
		\begin{minipage}{0.18\linewidth}
			\vspace{3pt}
			\centerline{\includegraphics[width=\textwidth]{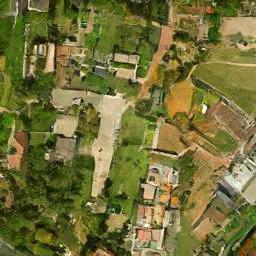}}
			\vspace{3pt}
			\centerline{\includegraphics[width=\textwidth]{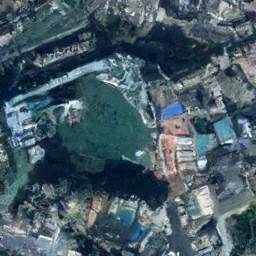}}
			\vspace{3pt}
			\centerline{\includegraphics[width=\textwidth]{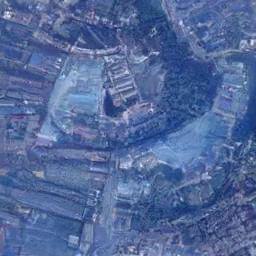}}
		\end{minipage}
		\begin{minipage}{0.18\linewidth}
			\vspace{3pt}
			\centerline{\includegraphics[width=\textwidth]{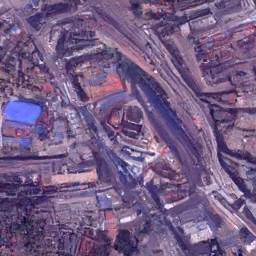}}
			\vspace{3pt}
			\centerline{\includegraphics[width=\textwidth]{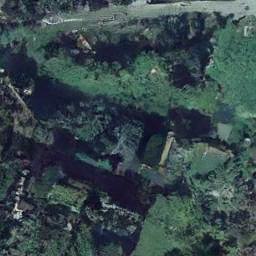}}
			\vspace{3pt}
			\centerline{\includegraphics[width=\textwidth]{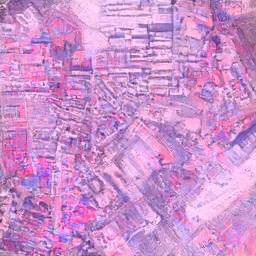}}
		\end{minipage}
		\begin{minipage}{0.18\linewidth}
			\vspace{3pt}
			\centerline{\includegraphics[width=\textwidth]{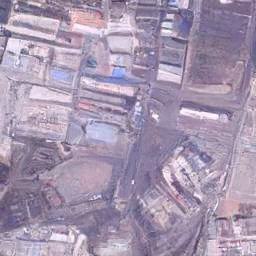}}
			\vspace{3pt}
			\centerline{\includegraphics[width=\textwidth]{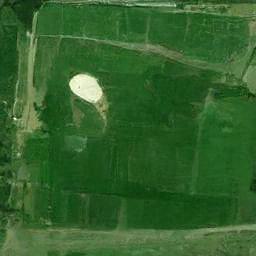}}
			\vspace{3pt}
			\centerline{\includegraphics[width=\textwidth]{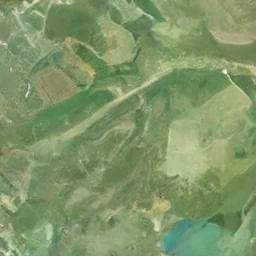}}
		\end{minipage}
		\begin{minipage}{0.18\linewidth}
			\vspace{3pt}
			\centerline{\includegraphics[width=\textwidth]{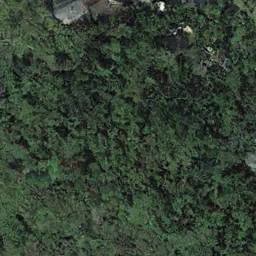}}
			\vspace{3pt}
			\centerline{\includegraphics[width=\textwidth]{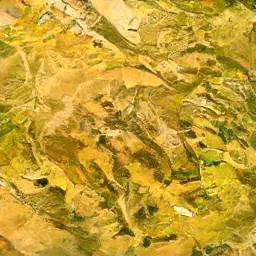}}
			\vspace{3pt}
			\centerline{\includegraphics[width=\textwidth]{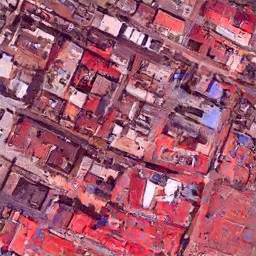}}
		\end{minipage}
		\begin{minipage}{0.18\linewidth}
			\vspace{3pt}
			\centerline{\includegraphics[width=\textwidth]{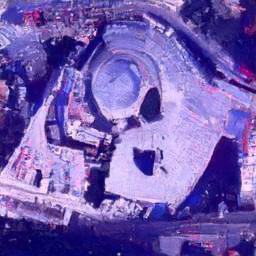}}
			\vspace{3pt}
			\centerline{\includegraphics[width=\textwidth]{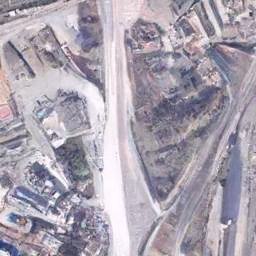}}
			\vspace{3pt}
			\centerline{\includegraphics[width=\textwidth]{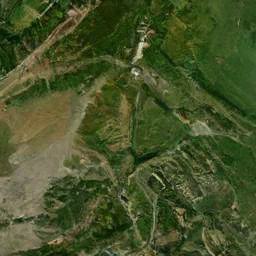}}
		\end{minipage}
		\caption{RS images synthesized by the pre-trained DDPM with random Guassian noises.}
		\label{img_gen}
	\end{figure}

Diffusion models have gained prominence in synthesizing high-resolution images in computer vision studies, often outperforming traditional GAN structures \cite{dhariwal2021diffusion}. We conducted ablation studies to assess the image-generation capability of the DDPM in the realm of remote sensing (RS).
Given that we trained the DDPM unconditionally, the generated images are randomly distributed within the domain of the training dataset. As depicted in Figure \ref{img_gen}, we generated images using a pre-trained DDPM on the AID dataset with an inference step of 1000. These results showcase that the DDPM can produce high-quality RS images with precise textures and accurate object boundaries. Moreover, the diversity of the generated images is promising, encompassing a wide array of scenarios in the context of RS.
In conclusion, the potential of RS image synthesis using diffusion models is promising, and further exploration into related studies, such as conditional image generation, is encouraged.

	\subsubsection{Noise Level Analysis}\label{ablation:noise level}
	\begin{figure*}[]
		\centering
		\begin{minipage}{0.10\linewidth}
			\vspace{3pt}
			\centerline{\includegraphics[width=\textwidth]{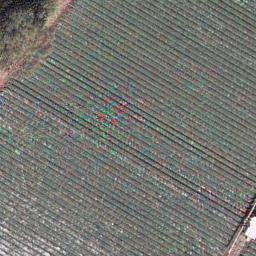}}
			\vspace{3pt}
			\centerline{\includegraphics[width=\textwidth]{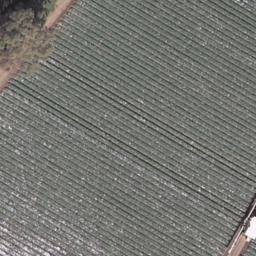}}
			\vspace{3pt}
			\centerline{Original}
		\end{minipage}
		\begin{minipage}{0.10\linewidth}
			\vspace{3pt}
			\centerline{\includegraphics[width=\textwidth]{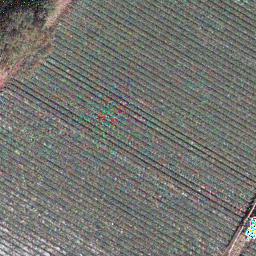}}
			\vspace{3pt}
			\centerline{\includegraphics[width=\textwidth]{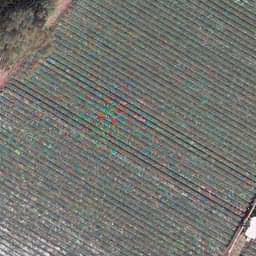}}
			\vspace{3pt}
			\centerline{$T_{m}=10$}
		\end{minipage}
		\begin{minipage}{0.10\linewidth}
			\vspace{3pt}
			\centerline{\includegraphics[width=\textwidth]{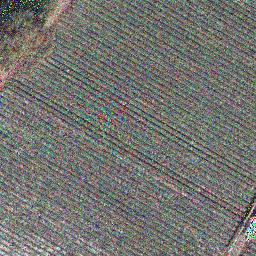}}
			\vspace{3pt}
			\centerline{\includegraphics[width=\textwidth]{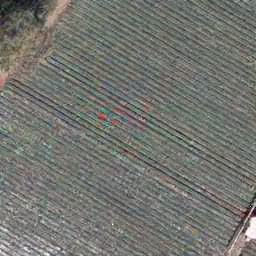}}
			\vspace{3pt}
			\centerline{$T_{m}=30$}
		\end{minipage}
		\begin{minipage}{0.10\linewidth}
			\vspace{3pt}
			\centerline{\includegraphics[width=\textwidth]{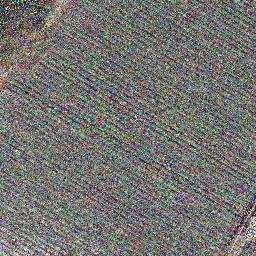}}
			\vspace{3pt}
			\centerline{\includegraphics[width=\textwidth]{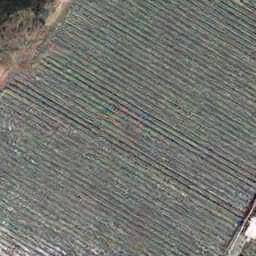}}
			\vspace{3pt}
			\centerline{$T_{m}=50$}
		\end{minipage}
		\begin{minipage}{0.10\linewidth}
			\vspace{3pt}
			\centerline{\includegraphics[width=\textwidth]{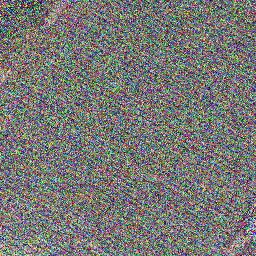}}
			\vspace{3pt}
			\centerline{\includegraphics[width=\textwidth]{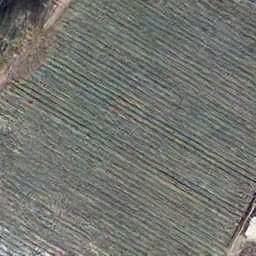}}
			\vspace{3pt}
			\centerline{$T_{m}=70$}
		\end{minipage}
		\begin{minipage}{0.10\linewidth}
			\vspace{3pt}
			\centerline{\includegraphics[width=\textwidth]{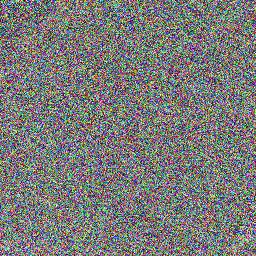}}
			\vspace{3pt}
			\centerline{\includegraphics[width=\textwidth]{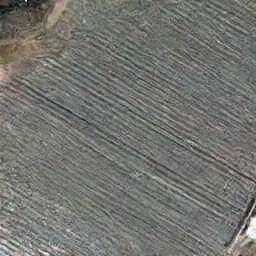}}
			\vspace{3pt}
			\centerline{$T_{m}=90$}
		\end{minipage}
		\begin{minipage}{0.10\linewidth}
			\vspace{3pt}
			\centerline{\includegraphics[width=\textwidth]{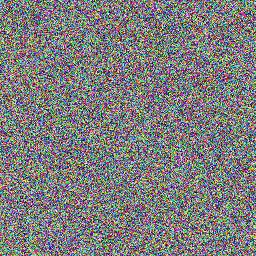}}
			\vspace{3pt}
			\centerline{\includegraphics[width=\textwidth]{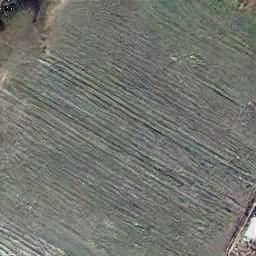}}
			\vspace{3pt}
			\centerline{$T_{m}=120$}
		\end{minipage}
		\begin{minipage}{0.10\linewidth}
			\vspace{3pt}
			\centerline{\includegraphics[width=\textwidth]{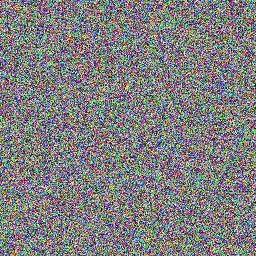}}
			\vspace{3pt}
			\centerline{\includegraphics[width=\textwidth]{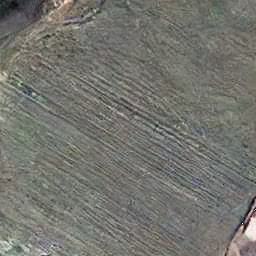}}
			\vspace{3pt}
			\centerline{$T_{m}=150$}
		\end{minipage}
		\begin{minipage}{0.10\linewidth}
			\vspace{3pt}
			\centerline{\includegraphics[width=\textwidth]{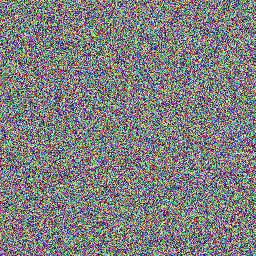}}
			\vspace{3pt}
			\centerline{\includegraphics[width=\textwidth]{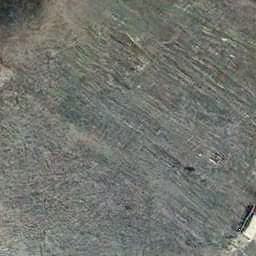}}
			\vspace{3pt}
			\centerline{$T_{m}=200$}
		\end{minipage}
		\caption{Purification results with different noise levels. The first column displays the adversarial sample (top row) and the original image (bottom row). The subsequent columns exhibit the diffused images with noise level $T_{m}$ (top row) and the corresponding purified results (bottom row).}
		\label{ablation_noise_level}
	\end{figure*}
 To visually illustrate the impact of different noise levels selected for purification, we diffuse an adversarial sample with varying noise levels $T_{m}$ and subsequently denoise it, as depicted in Figure \ref{ablation_noise_level}. When the adversarial sample is diffused with a small $T_{m}$, most of the texture and background of the image are retained, albeit with some residual adversarial perturbations in the denoised result. Conversely, selecting a larger noise level results in the diffused image appearing predominantly as noise, seemingly destroying the contents in the denoised images, even though no perturbations are present. Both extremes yield unsatisfactory performance for the classifier. Only when $T_{m}$ is correctly selected can the denoised image be best interpreted by the victim DNNs.
	
	\subsubsection{Cross-Domain Purification}
Given the heterogeneity of RS datasets, DNNs often experience performance degradation when applied to data from diverse domains. As part of an ablation study, we assess the cross-domain adversarial purification capabilities of the UAD-RS model.
Initially, we pre-trained the diffusion model on the UCM dataset and then employed it to purify adversarial samples generated from the AID dataset, using various classifiers and attack algorithms. The adaptive noise level selection mechanism was also integrated into this ablation study.

Table \ref{ablation3} presents the classification results of the cross-domain purified samples and their deviation from intra-domain performance. Notably, although the UAD-RS model was trained on the UCM dataset, it demonstrates a degree of effectiveness in purifying adversarial samples from the AID dataset. Particularly noteworthy are some cross-domain results following FGSM and Jitter attacks, which closely resemble the original intra-domain outcomes. However, a notable performance decline is observed compared to the model pre-trained on the AID dataset, as indicated by the values in parentheses. In conclusion, while the UAD-RS exhibits some capability for cross-domain adversarial purification, its effectiveness is somewhat restricted.

	\begin{table*}[]
		
		\centering
		\caption{Ablation Studies for Cross-domain Adversarial Purification using UAD-RS. The model was first pre-trained on the UCM dataset and then used to purify the adversarial samples generated from the AID dataset.}
    \resizebox{\textwidth}{!}{
		\begin{tabular}{cccccccc}
			\hline
			\multirow{2}{*}{Victim DNNs} & \multicolumn{7}{c}{Attack Methods}                      \\
			& FGSM  & IFGSM & CW    & Jitter & Mixcut & Mixup & TPGD  \\ \hline
			AlexNet                      & 53.72 (-14.96) & 35.76 (-21.36) & 36.12 (-21.28) & 60.26 (-12.30)  & 45.12 (-2.86) & 22.52  (-7.60) & 41.16 (-18.00) \\
			DenseNet-121              & 68.10 (-6.92) & 43.42 (-15.94) & 44.00 (-15.08) & 69.48 (-5.88) & 16.00 (-15.54) & 15.50 (-16.60) & 52.74 (-10.30) \\
			ResNet-18                     & 54.92 (-11.60) & 37.40 (-19.04) & 38.56 (-18.78)  & 61.10 (-8.18) & 23.84 (-11.00) & 23.24 (-11.26) & 50.40 (-11.76)\\
			RegNetX-400MF             & 64.40 (12.48) & 50.68 (-17.04) & 50.02 (-18.18) & 69.66 (-8.84)  & 51.40 (-8.02) & 41.22 (-12.06) & 60.52 (-12.48) \\ \hline
		\end{tabular}}
		\label{ablation3}
	\end{table*}
	\section{Discussion}\label{sec:discussion}
\subsection{Diffusion and GAN Models}
Diffusion and GAN models are both generative models frequently employed in various computer vision tasks and often compared for their efficacy. Both models can predict an image given a noise input and are capable of generalizing representations within the domain of the training dataset, making them suitable for purifying adversarial samples by transferring them to a clean domain.
Specifically, the defense-GAN and diffusion methods both leverage a noise latent to reconstruct an image in the clean domain. However, they differ in their approach. The defense-GAN optimizes a series of random noises to find the optimal one that generates an image similar to the original adversarial sample. In contrast, diffusion-based methods proactively generate a latent noise by progressively diffusing the adversarial sample.

This distinction results in several key differences. The performance of defense-GAN models is often constrained by the randomness of noise optimization and the challenges associated with training GAN models. Additionally, the extensive inference required for hundreds of steps of noise optimization per sample can make defense-GAN approaches less cost-effective.
In contrast, diffusion models tend to outperform GAN models due to their stable conditional noise latent generation. The proactive generation of noise through diffusion results in more reliable purification of adversarial samples. Overall, diffusion models offer a more effective and efficient approach to adversarial sample purification compared to GAN models.

\subsection{Unknown Adversarial Threat}
The unknown adversarial threat poses a significant challenge in adversarial defense studies, as it means the prior information of the adversarial perturbation, including the attack algorithms and intensity, is unknown. Many defense approaches, particularly adversarial training methods, suffer from performance loss due to this lack of prior knowledge. Our proposed UAD-RS overcomes this difficulty by providing a novel universal adversarial defense paradigm that can purify samples based solely on a pre-trained diffusion model without any information from the attacks.

Our proposed universal adversarial defense method aims to purify universal adversarial examples of heterogeneous patterns from a single dataset using only one pre-trained diffusion model. Additionally, unlike traditional methods that require training a new defense model for each classifier and attack combination, our approach is more cost-efficient and adaptable to different settings without unnecessary training efforts.

\subsection{Limitations of the UAD-RS Model}
The main limitations of the UAD-RS model are twofold. First, compared to traditional DNN-based purification models, the UAD-RS model may require more time to purify adversarial examples due to the multiple diffusion and denoising steps involved. Additionally, the large number of trainable parameters in the diffusion models results in higher computational costs during the pre-training process. Second, the diffusion process of the UAD-RS model may be challenged by high-contrast remote sensing (RS) data containing exotic pixels with extreme brightness, such as those found in the Zurich dataset. In these cases, the Gaussian noise added during the diffusion process may be more pronounced in white pixels and less visible in dark ones, leading to ambiguous and uneven denoising results in subsequent steps.
 
\section{Conclusion}\label{sec:conclusion}

In this article, we introduced the UAD-RS model, a universal adversarial defense method designed to counter multiple adversarial attacks on DNNs used for scene classification and semantic segmentation in RS imagery. The UAD-RS leverages generative diffusion models, initially pre-trained on common RS datasets, to capture generalized representations of the data domain. Subsequently, we developed an adversarial purification framework that utilizes these pre-trained models to remove perturbations from adversarial samples. Additionally, an ANLS algorithm was designed to optimize the noise level settings of the diffusion model, achieving purification results that are closest to clean samples.

Experimental results across various tasks, including scene classification and semantic segmentation in complex scenes, demonstrate the effectiveness of UAD-RS in purifying adversarial samples from diverse attacks targeting different DNNs. Compared to state-of-the-art methods, UAD-RS consistently delivers superior performance, yielding accurate purified results suitable for scene categorization and segmentation tasks.

To facilitate further research in RS, we plan to release the pre-trained diffusion models to alleviate the burden of retraining on large datasets. Moving forward, we aim to extend the application of the purification model to defend vulnerable DNNs in additional RS applications susceptible to adversarial attacks. Moreover, exploring the potential application of UAD-RS in tasks like domain adaptation and image enhancement will be a focus of future studies.

	\section{Acknowledgment}
	The Helmholtz Institute Freiberg for Resource Technology is gratefully thanked for supporting this project. We would also like to thank the European Regional Development Fund and the Land of Saxony for funding the GPU resources used in this study.
	
\bibliographystyle{elsarticle-num-names}
\bibliography{reference}

\end{document}